\documentclass[11pt,letterpaper,twoside,reqno,nosumlimits]{amsart}

\usepackage[utf8]{inputenc}
\usepackage{url}
\usepackage[a4paper,margin=2.5cm]{geometry}

\usepackage{comment}
\usepackage{amssymb}
\usepackage{amsmath}
\usepackage{mathrsfs,amsmath} 
\usepackage{amsfonts}
\usepackage{mathtools}
\usepackage{float}
\usepackage{multirow}
\usepackage{dirtytalk}

\mathtoolsset{showonlyrefs}

\usepackage{tikz,pgflibraryshapes}
\usetikzlibrary{shadows}
\usetikzlibrary{arrows}

\newtheorem{thm}{Theorem}[section]

\newtheorem{remark}[thm]{Remark}

\usepackage{algorithm}
\usepackage{algpseudocode}

\newcommand{\R}{{\mathbb R}}

\newcommand{\E}{{\mathbb E}}

\newcommand{\bs}{\boldsymbol}

\newcommand{\norm}[1]{\left\lVert#1\right\rVert}

\title[Learning dynamical systems from data, SDEs]{One-Shot Learning of Stochastic Differential Equations with data adapted kernels
}
\author[M. Darcy]{Matthieu Darcy$^1$}
\address{$^1$ Department of Computing and Mathematical Sciences, Caltech, CA, USA.}
\email{mdarcy@caltech.edu}

\author[B. Hamzi]{Boumediene Hamzi$^2$}
\address{$^4$ Department of Computing and Mathematical Sciences, Caltech, CA, USA.}
\email{boumediene.hamzi@gmail.com}

\author[G. Livieri]{Giulia Livieri$^3$}
\address{$^3$ Scuola Normale Superiore, Pisa, Italy}
\email{giulia.livieri@sns.it}

\author[H. Owhadi]{Houman Owhadi$^4$}
\address{$^5$Department of Computing and Mathematical Sciences, Caltech, CA, USA. }
\email{ owhadi@caltech.edu}

 \author[P. Tavallali]{Peyman Tavallali$^5$}
\address{$^2$ JPL, Caltech, CA, USA.}
\email{peyman.tavallali@jpl.nasa.gov}

\begin{document}
\maketitle

 \begin{abstract}
     We consider the problem of learning Stochastic Differential Equations of the form  $dX_t = f(X_t)dt+\sigma(X_t)dW_t $ from one sample trajectory. This problem is more challenging than learning deterministic dynamical systems because one sample trajectory only provides indirect information on the unknown functions $f$, $\sigma$, and stochastic process $dW_t$ representing the drift, the diffusion, and the stochastic forcing terms, respectively. We propose a method that combines Computational Graph Completion \cite{owhadi2021computational} and data adapted kernels learned via a new variant of cross validation. Our approach can be decomposed as follows: (1) Represent the time-increment map $X_t \rightarrow X_{t+dt}$ as a Computational Graph in which $f$, $\sigma$ and $dW_t$ appear as unknown functions and random variables. (2) Complete the graph (approximate unknown functions and random variables) via Maximum a Posteriori Estimation (given the data) with Gaussian Process (GP) priors on the unknown functions. (3) Learn the covariance functions (kernels) of the GP priors from data with randomized cross-validation. Numerical experiments illustrate the efficacy, robustness, and scope of our method.
 \end{abstract}
 
\section{Introduction}
Forecasting a stochastic or a deterministic time series is a fundamental problem in, e.g., Econometrics or Dynamical Systems, which is commonly solved by learning and/or inferring a stochastic or a deterministic dynamical system model from the observed data, respectively; see, e.g., \cite{kantz97,CASDAGLI1989,survey_kf_ann,Sindy,jaideep1,nielsen2019practical,abarbanel2012analysis, kleinhans_sdes,  pmlr-v1-archambeau07a,8516991, Neumaier98,DBLP:journals/corr/abs-2105-08449, peinke1998,Klimo, FRIEDRICH201187,dietrich2021learning, opper,hyndman2021}, among many others.

\subsection{On the kernel methods to forecasting time series} Among the various learning-based approaches, methods based on kernels hold potential for considerable advantages in terms of theoretical analysis, numerical implementation, regularization, guaranteed convergence, automatization, and interpretability over, e.g., methods based on variants of Artificial Neural Networks (ANNs); see, e.g., \cite{chen2021solving, owhadi2021computational}. In particular, Reproducing Kernel Hilbert Spaces (RKHS) \cite{CuckerandSmale} have provided a strong mathematical foundations for studying dynamical systems \cite{BH10allerton, bhcm11,bhcm1,lyap_bh,bh2020a,hamzi2019kernel, bh2020b,klus2020data,ALEXANDER2020132520,bhks,bh12,bh17,hb17,mmd_kernels_bh} and surrogate modeling (see, e.g., \cite{santinhaasdonk19} for a survey). However, these emulators' accuracy hinges on the kernel's choice; nonetheless, the problem of selecting a good kernel has received less attention so far. Numerical experiments have recently shown that when the time series is regularly \cite{BHPhysicaD} or is irregularly sampled \cite{lee2021learning}, simple kernel methods can successfully reconstruct the dynamics of prototypical chaotic dynamical systems when kernels are also learned from data via Kernels Flows (KF), a variant of cross-validation \cite{Owhadi19}. KF approach has then been applied to complex, large-scale systems, including geophysical data \cite{hamzimaulikowhadi,bhkfjpl,bhkfjpl1}, and to learning non-parametric kernels for dynamical systems \cite{bh_kfs_nonpar}.
\subsection{On the learning of Stochastic Differential Equations (SDEs)} While time series produced by deterministic dynamical systems offer a direct observation of the vector-field (i.e., of the drift) driving those systems, those produced by SDEs only present an indirect observation of the underlying drift, diffusion, and stochastic forcing terms. A popular approach employed to recover the drift and the diffusion of an SDE is the so-called Kramers-Moyal expansion; see, e.g., \cite{risken1989fpe, FRIEDRICH201187}. In this manuscript, we formulate the problem of learning stochastic dynamical systems described by SDEs as that of completing a computation graph \cite{owhadi2021computational}, which represents the functional dependencies between the observed increments of the time-series and the unknown quantities. Our approach to solving this Computational Graph Completion (CGC) problem can be summarized as (1) replacing unknown functions and variables with Gaussian Processes (GPs) and (2) approximating those functions by the Maximum a Posteriori (MAP) estimator of those GPs given available data. The covariance kernels of these GPs are learned from data via a randomized cross-validation procedure.    

\subsection{Outline of the article} Section \ref{sec:section2} describes the problem we focus on in this manuscript and our proposed solution. Section \ref{sec:section3} describes the MAP estimator for the GPs in the one dimensional case. Section \ref{sec:section4} describes the algorithm we propose to learn the kernels and the hyper-parameters. Section \ref{sec: exp} provides numerical results, with an additional discussion of the impact of the choice of kernels in Section \ref{sec: specified kernels} and the impact of time discretization in Section \ref{sec: failure}. We conclude with a brief discussion in Section \ref{sec: discussion}. Section \ref{sec:section7} displays additional plots.

\section{Statement of the problem and proposed solution}\label{sec:section2}
We first describe the type of SDEs used here. We consider SDEs of the form: 
\begin{equation}\label{eq:SDE}
    dX_t =f(X_t)dt + \sigma(X_t)  dW_t
\end{equation}
\noindent with initial condition $X_0 = x_0$. In the previous equation, $(W_t)_{t \in [0, T]}$ denotes a Wiener process. We assume that the process in Equation~\eqref{eq:SDE} is observed at discrete times $t_n$, $n = 1 \ldots N$, such that the time intervals $\Delta t_n:=(t_{n+1}-t_n)$ between observations $X_n:=X_{t_n}$ of the time series are not small enough so that $\sigma(X_t)$ can be efficiently approximated by estimating the quadratic variation of $X_t$ near $t$, but small enough so that the following approximation holds:
\begin{equation}\label{eq:SDE_approx}
    X_{n+1} = X_n + f(X_n) \Delta t_n + \sigma (X_n) \sqrt{\Delta t_n} \xi_n + \varepsilon_n
    \,,
\end{equation}
where the \emph{i.i.d.} random variables $\xi_n \overset{d}{\sim} \mathcal{N}(0, 1)$ represent Brownian Motion increments and the \emph{i.i.d.} random variables $\varepsilon_n \overset{d}{\sim} \mathcal{N}(0, \lambda)$ represent discretization noise/misspecification\footnote{While this assumption is not well justified by theory, the Euler-Maruyama method yields a practical scaling for the variance $\lambda$ of the error. Because the Euler-Maruyama method has a strong order of convergence $\mathcal{O}(\Delta t)$, it is therefore reasonable to choose $\lambda = C \Delta t$. In practice, we find that the constant $C$ should be small and use $C = 0.01$.}; henceforth, the notation $``\overset{d}{\sim}"$ stands for ``distributed as". We seek to recover/approximate the unknown functions $f$ and $\sigma$ from the data 
 $(\boldsymbol{X}, \boldsymbol{Y}) = \{ (X_n, Y_n)\}_{1 \leq n \leq N}$, where 
\begin{equation*}
    Y_{n} := X_{n+1} - X_{n} \,.
\end{equation*}
Therefore, the relation between $X_n$ and $Y_n$ is given by our modeling assumption:
\begin{equation}\label{eq:xi}
    Y_n = f(X_n)\Delta t_n + \sigma (X_n)\sqrt{\Delta t_n}\xi_n + \varepsilon_n\,.
\end{equation}

\subsection{The Computational Graph Completion problem}
In general, a \emph{computational graph} is defined as a graph representing functional dependencies between a finite number of (not necessarily random) variables and functions. We will use nodes to represent variables and arrows to represent functions. We will color known functions in black and unknown functions in red. Random variables are drawn in blue and primary variables as squares. We will distinguish nodes used to aggregate variables by drawing them as circles. Multiple incoming arrows into a square node are interpreted as a sum. Now, let $\bar{f}_n:=f(X_n)$ and $\bar{\sigma}_n:=\sigma(X_n)$ be two intermediate (unobserved) variables. Equation~\eqref{eq:xi} can thus be rewritten as: 
\begin{equation}\label{eq:regression_problem}
    Y_n = \bar{f}_n\Delta t_n + \bar{\sigma}_n\sqrt{\Delta t_n}\xi_n + \varepsilon_n.\,
\end{equation}
In particular, it can be represented as the following computational graph:
\vskip 0.5cm
\centerline{
\begin{tikzpicture}[->,>=stealth',shorten >=1pt,auto,node distance=3cm,
                    thick,main node/.style={rectangle,draw,font=\sffamily\Large\bfseries}]

\node[main node] (1) {$X_n$};
\node[main node] (2) [right of=1] {$\bar{f}_n$};
\node[main node] (3) [left of=1] {$\bar{\sigma}_n$};
\node[main node] (4) [above of=1, node distance=2cm] {$\Delta t_n$};
\node[main node] (5) [left of=3] {$\bar{\sigma}_n \sqrt{\Delta t_n}\xi_n$};
\node[main node] (6) [right of=4,circle] {};
\node[main node] (7) [right of=2] {$\bar{f}_n\Delta t_n$};
\node[main node] (8) [above of=5, node distance=2cm,blue] {$\xi_n$};
\node[main node] (9) [above of=3,node distance=2cm,circle] {};
\node[main node] (10) [below of=1,node distance=2cm] {$Y_n$};
\node[main node] (11) [right of=10,blue] {$\varepsilon_n$};

\path[every node/.style={font=\sffamily\Large\bfseries},red]
    (1) edge node [above ] {$f$} (2);

\path[every node/.style={font=\sffamily\Large\bfseries},red]
    (1) edge node [above ] {$\sigma$} (3);

\path[every node/.style={font=\sffamily\Large\bfseries}]
    (1) edge node [above ] {} (10);
    
\path[every node/.style={font=\sffamily\Large\bfseries}]
    (11) edge node [above ] {} (10);

\path[every node/.style={font=\sffamily\Large\bfseries}]
    (5) edge node [above ] {} (10);
    
\path[every node/.style={font=\sffamily\Large\bfseries}]
    (7) edge node [above ] {} (10);
    
\path[every node/.style={font=\sffamily\Large\bfseries}]
    (4) edge node [above ] {} (6);
\path[every node/.style={font=\sffamily\Large\bfseries}]
    (4) edge node [above ] {} (9);
\path[every node/.style={font=\sffamily\Large\bfseries}]
    (2) edge node [above ] {} (6);
\path[every node/.style={font=\sffamily\Large\bfseries}]
    (3) edge node [above ] {} (9);
\path[every node/.style={font=\sffamily\Large\bfseries}]
    (8) edge node [above ] {} (9);

\path[every node/.style={font=\sffamily\Large\bfseries}]
    (6) edge node [above ] {} (7);
\path[every node/.style={font=\sffamily\Large\bfseries}]
    (9) edge node [above ] {} (5);

\end{tikzpicture}}
\vskip 0.5cm

\noindent We now formulate the learning problem in this manuscript as the problem of completing the just displayed computational graph; see \cite{owhadi2021computational}. Let $X_1, \ldots X_N$, $Y_1, \ldots, Y_N$ and $\Delta t_1, \ldots, \Delta t_N$ be the $N$ observations data: our goal is to approximate the unknown functions $f$ and $\sigma$ from these observations. In order to solve this problem, we will use the GP framework: we replace $\sigma$ and $f$ with GPs and approximate them via MAP estimation given the data. More precisely, we assume that $f$ and $\sigma$ are mutually independent GPs, with centered Gaussian priors $f \overset{d}{\sim} \mathcal{GP}(\boldsymbol{0}, \boldsymbol{K}), \sigma \overset{d}{\sim} \mathcal{GP}(\boldsymbol{0}, \boldsymbol{G})$ defined by the covariance functions/kernels $\boldsymbol{K}$ and $ \boldsymbol{G}$.

\begin{remark}
Although in this case the computational graph serves mostly as a visual aid, its underlying formalism (i.e., draw the graph, replace unknown functions by GPs, and compute their MAP from the data to complete the graph) is a simple pathway to generalize the method to learning more complex systems than SDEs.
\end{remark}
\section{MAP estimator}\label{sec:section3}
Write $\bar{f}$ for the vector with entries $\{\bar{f}_n\}_{1 \leq n \leq N}$ and $\bar{\sigma}$ for the vector with entries 
$\{\bar{\sigma}_n\}_{1 \leq n \leq N}$. Observe that given $\bar{\sigma}$ and $\bar{f}$, the identification of the functions $f$ and $\sigma$ reduces to two separate simple kernel regression problems. We will therefore first focus on the estimation of $\bar{f}$ and $\bar{\sigma}$. Since $f$ and $\sigma$ are independent, $\bar{f}=f(X)$ and $\bar{\sigma}=\sigma(X)$ are conditionally (on $X$) independent. Using the shorthand notations $p(Y|X)$ for $p(A|X)$ where $A$ is the event $\{Y_n = f(X_n)\Delta t_n + \sigma (X_n)\sqrt{\Delta t_n}\xi_n + \varepsilon_n, \text{ for }1 \leq n \leq N\}$, 
we deduce that
\begin{equation*}
    p(\bar{f}, \bar{\sigma} | Y,X ) = p(Y |  \bar{f},\bar{\sigma},X)\frac{p(\bar{f}  |X)p(\bar{\sigma}  |X)}{p(Y |X)}\,,
\end{equation*}

It follows that a MAP estimator of $(\bar{f},\bar{\sigma})$ is a minimizer of the loss $ - \ln \big(p(Y | X, \bar{f}, \bar{\sigma})p(\bar{f}  |X)p(\bar{\sigma}|X)\big)$, which up to a constant ($\log \det K(X,X)+\log \det G(X,X)$) and a multiplicative factor $1/2$ is equal to

\begin{equation}\label{eq: log likelihood sigma0}
\begin{split}
    \mathcal{L}( \bar{f}, \bar{\sigma})& := (Y - \Lambda \bar{f} )^T (\Sigma + \lambda I)^{-1}(Y - \Lambda \bar{f}) 
    + \sum_{n= 1}^{N}\ln(\bar{\sigma}_n^2 \Delta t_n+\lambda)
    \\& +  \bar{f}^TK(X,X)^{-1}\bar{f}+ \bar{\sigma}^T G(X,X)^{-1}\bar{\sigma}\,.
\end{split}
\end{equation}
where  $K(X,X)$ is the $N\times N$ matrix with entries $K(X_i,X_j)$, $G(X,X)$ is the $N\times N$ matrix with entries $G(X_i,X_j)$, $\Sigma$ is the diagonal $N\times N$ matrix with diagonal entries $\bar{\sigma_n}^2\Delta t_n$, and $\Lambda$ is the diagonal $N\times N$ matrix with diagonal entries $\Delta t_n$.

\subsection{ Recovery of $f$.} First, observe that given $\bar{\sigma}$,  \eqref{eq: log likelihood sigma0} is quadratic in $\bar{f}$ and its minimizer in $\bar{f}$ is 
\begin{equation}\label{eq:eq_1}
    \bar{f}^*(\bar{\sigma})=  K(X, X)\Lambda\Big( \Lambda K(X,X) \Lambda  +\Sigma + \lambda I \Big)^{-1}Y\,.
\end{equation}
Therefore $f\overset{d}{\sim}\mathcal{GP}(\boldsymbol{O}, \boldsymbol{K})$ conditioned on  $(X,f(X)=\bar{f})$
is normally distributed with (conditional) mean
\begin{equation}\label{eq:eq_2}
    f^*(x) :=   K(x, X)\Lambda\Big(\Lambda K(X,X)\Lambda +  \Sigma + \lambda I\Big)^{-1}Y \,,
\end{equation}
and (conditional) covariance
\begin{equation}\label{eq:eq_3}
K(x, x) - K(x, X)\Lambda\Big(\Lambda K(X,X)\Lambda +  \Sigma + \lambda I\Big)^{-1}\Lambda K(X, x)\,.
\end{equation}
We therefore estimate $f$ with $f^*=$\eqref{eq:eq_2}. Note that \eqref{eq:eq_2} and \eqref{eq:eq_3} can also be recovered by observing that given $\bar{\sigma}$, Equation \eqref{eq:regression_problem} corresponds to a noisy regression problem, with noise coming from two independent Gaussian variables. This is proved in Appendix \ref{app:gaussian_regression} and it is an easy modification of the proof presented in \cite[Page 306]{bishop2006pattern}

\subsection{Recovery of $\sigma$.} 
Taking $\bar{f}=\bar{f}^*=$\eqref{eq:eq_2} in  \eqref{eq: log likelihood sigma0}, the estimation of $\bar{\sigma}$ reduces to the minimization of the loss

\begin{equation}\label{eq: log likelihood sigma}
\begin{split}
    \mathcal{L}(  \bar{f}^*(\bar{\sigma}), \bar{\sigma})
    &=(Y - \Lambda \bar{f}^*(\bar{\sigma}))^T (\Sigma + \lambda I)^{-1}(Y - \Lambda \bar{f}^*(\bar{\sigma})) 
    + \sum_{n= 1}^{N}\ln(\bar{\sigma}_n^2 \Delta t_n+\lambda)
    \\&\quad +  \bar{f}^*(\bar{\sigma})^TK(X,X)^{-1}\bar{f}^*(\bar{\sigma})+ \bar{\sigma}^T G(X,X)^{-1}\bar{\sigma}\,.
\end{split}
\end{equation}
Write $\bar{\sigma}^\dagger$ for a minimizer of \eqref{eq: log likelihood sigma} obtained through a numerical optimization algorithm (e.g.,  gradient descent). Because the numerical approximation of $\bar{\sigma}^\dagger$ is noisy, we then further estimate $\bar{\sigma}$ as the mean of  the Gaussian vector $\sigma(X)$ conditioned on  $\sigma(X)=\bar{\sigma}^\dagger+Z$ where the entries $Z_i$ of the (noise) vector $Z$ are \emph{i.i.d.} Gaussian with variance $\gamma$. We therefore approximate $\bar{\sigma}$ with 
\begin{equation}\label{eq: mean sigma}
    \bar{\sigma}^*= \E[\sigma (X)|\sigma(X)+Z=\bar{\sigma}^\dagger] = G(X, X)(G(X, X) + \gamma I )^{-1}\sigma^{\dagger}.
\end{equation}
and $\sigma$ with 
\begin{equation*}
 \sigma^*(x)=G(x, X)(G(X, X) + \gamma I )^{-1}\bar{\sigma}^{\dagger}\,.
\end{equation*}

\subsection{Minimization of $\mathcal{L}$}
  The natural approach to identification of $\bar{\sigma}^\dagger$ is to minimize (with respect to $\bar{\sigma}$) \eqref{eq: log likelihood sigma} with  $\bar{f}^*$  defined as the function \eqref{eq:eq_1} of $\bar{\sigma}$. The minimization of $ \eqref{eq: log likelihood sigma}$ is difficult given that this is a non-convex, non-linear problem. We, therefore use a gradient descent algorithm with a step size chosen such that the perturbation is no more than $p\%$ in norm. We set $p = 1$ initially and increment to $p \leftarrow 0.9p$ if the resulting perturbation does not reduce the loss. We optimize until $p < 10^{-20}$ or a maximum $N = 10^5$ iterations. We also use Newton method with step-size chosen using Armijo's condition \cite{NoceWrig06} until $\norm{\nabla \mathcal{L}} < 10^{-8}$ for a maximum of $N =1000$ iterations. The gradient descent algorithm is slow compared to Newton's method (10-20 seconds compared to 10-15 minutes in wall clock time) but consistently produces better results and is less sensitive to changes in the initial condition. In our gradient descent algorithm, we ensure that the loss is decreased at every step and we therefore converge to a local minimum. However, there is no guarantee that we converge to the global minima since the loss is non-convex.
  
 \subsection{Initial condition.} The initialization of the optimization problem requires the specification of an initial condition for $\bar{\sigma}_{\text{init}}$. We use an estimate of quadratic variation of the process given the data:
 \begin{equation}
     (\bar{\sigma}_{\text{init}})_i = \frac{(X_i - X_{i-1})^2}{\Delta t_i}.
 \end{equation}
 We smooth out this noisy estimate using the mean of the Gaussian process $\sigma$ conditioned on these values:
 \begin{equation}
    \bar{\sigma}_{\text{init}}= \E[\sigma (X)|\sigma(X)+Z=\bar{\sigma}_{\text{init}}] = G(X, X)(G(X, X) + \gamma I )^{-1}\bar{\sigma}_{\text{init}}.
\end{equation}

\subsection{Extension to the multivariate case} The extension to the multivariate case is relatively straightforward and is presented in appendix \ref{sec:multivariate}.

\section{Learning the kernels and the hyperparameters}\label{sec:section4}

\subsection{Motivation}
The computational graph completion approach relies on a choice of prior and hence on a choice of kernel. There are many possible such choices that encode varying priors on the functions $f$ and $\sigma$. For example isotropic kernels such as the Gaussian and Matérn kernel are used to model the covariance functions of stationary processes \cite{GPforML}. Dot product-based kernels, such as the polynomial kernel, on the other hand are non-stationary \cite{GPforML}. Moreover, each family of kernels is parameterized by one or several parameters, which can have a large impact on the recovery of the function. This motivates the development of a method to select the optimal kernel. In the next section, we present a cross-validation-based method to select the best kernel based on the data. This method is able to not only select the best parameter but also enables ill-specified kernels to perform comparably to well-specified kernels. 

To illustrate this point, consider the Geometric Brownian Motion  
\begin{equation}\label{eqhgeududyu}
    dX_t = \mu X_t dt + \sigma X_tdW_t 
\end{equation}
which has both non-stationary drift and diffusion functions. In section \ref{sec: specified kernels}, we will show (see
figure \ref{fig: pred, GBM linear vs matern})  how our hyper-parameter learning algorithm enables a non-perfectly-adapted kernel (e.g., a Mat\'{e}rn kernel for approximating the drift and volatility of \eqref{eqhgeududyu}) to perform comparably to a perfectly adapted kernel (e.g., a linear kernel for approximating the drift and volatility of \eqref{eqhgeududyu}).
We also note that learning the kernel improves out-of-sample predictions.

\begin{figure}[h] 
    \centering
        \includegraphics[width = 0.3\linewidth, height = 3cm]{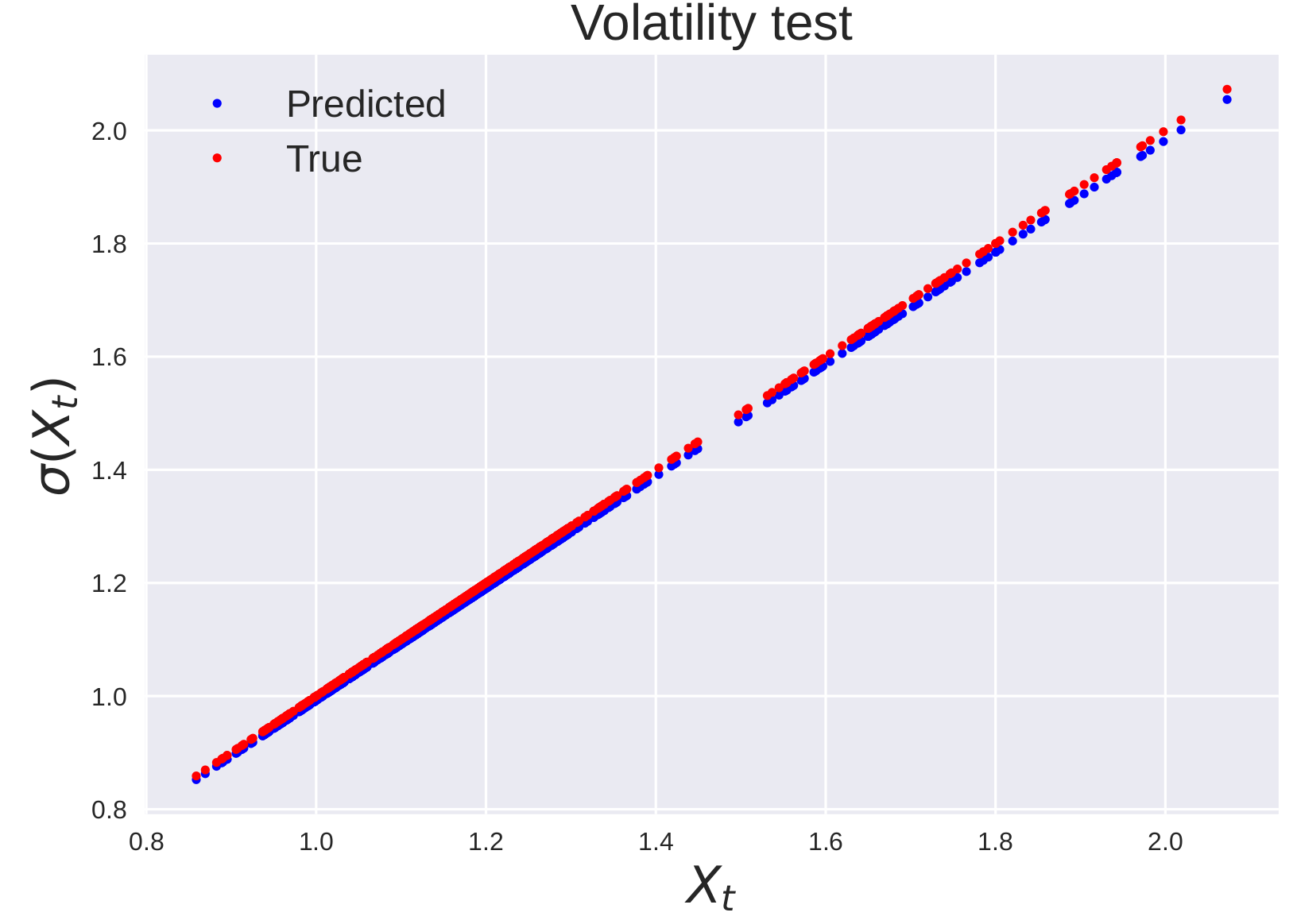}
    \includegraphics[width = 0.3\linewidth, height = 3cm]{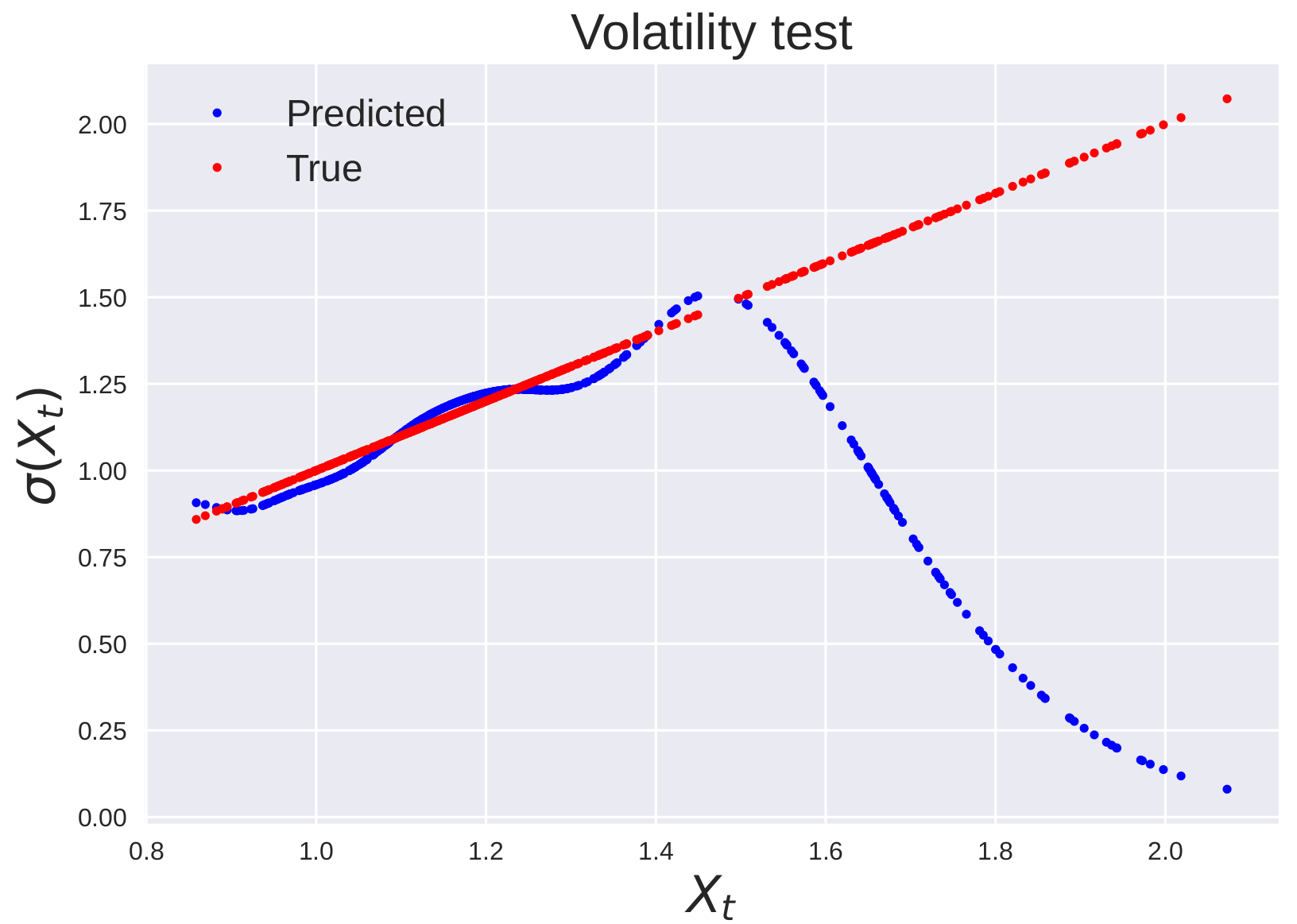}
    \includegraphics[width = 0.3\linewidth, height = 3cm]{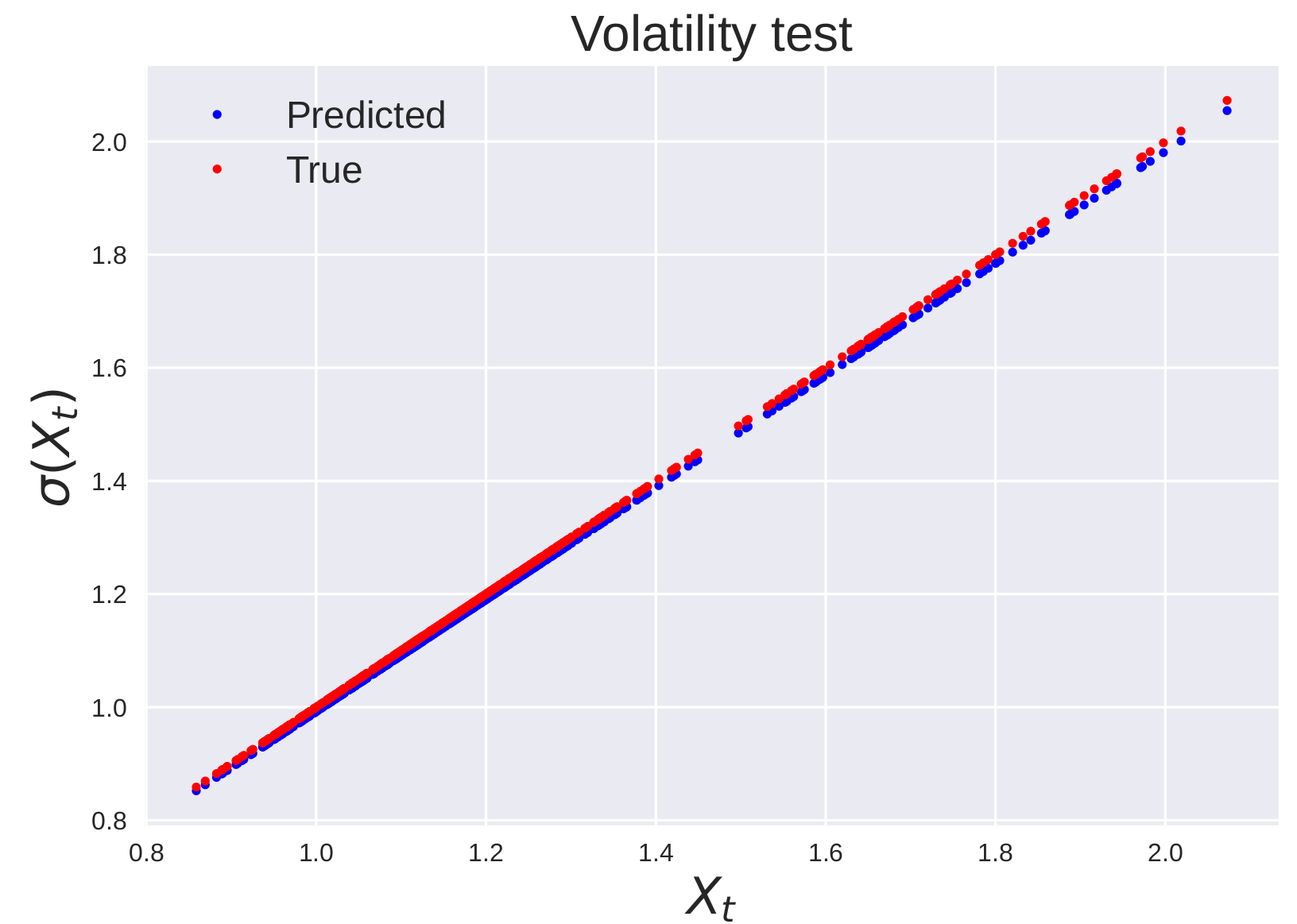}
    \caption{Predicted volatility: well-specified kernel (left), ill specified kernel (middle), and well-specified kernel with learned parameters(right).}
    \label{fig: pred, GBM linear vs matern}
\end{figure}
\subsection{Methodology}
We now describe how to select the kernels $K,G$ in a family of kernels parameterized by  $\theta_k, \theta_g$, which we learn from data using a cross-validation approach. Writing
\begin{equation}
    \boldsymbol{\theta} := (\theta_k, \theta_g)\,,
\end{equation}
for the vector formed by all the hyperparameters of our approach, we learn $\theta$ through a robust-learning cross-validation approach which we will now describe. Consider the set of all sets of possible partitions of the training data $\mathcal{D}_{\mathcal{T}}$ with indices $\mathcal{I}$ into two mutually disjoint subsets of equal size.

\begin{equation}
    \mathcal{A} = \{ (\Pi, \Pi^c) |  \Pi \cup \Pi^c = \mathcal{I}, \Pi \cap \Pi^c = \emptyset, |\Pi| = \frac{|\mathcal{I}|}{2} \}.
\end{equation}
Write $\mathcal{D}_{\Pi} = (x_j, y_j)_{j \in \Pi}$ for the set of points belonging to the first partition and $\mathcal{D}_{\Pi^c} = (x_j, y_j)_{j \in \Pi^c}$ for the set of points belonging to the second set. 
Write $\mathbb{E}_{ (\Pi, \Pi^c) }$ for the uniform distribution over $\mathcal{A}$.
For $(\Pi, \Pi^c)\in \mathcal{A}$ write $\mathcal{L} (\mathcal{D}_{\Pi^c}, \bar{f}, \bar{\sigma})$ for  the MAP loss \eqref{eq: log likelihood sigma0} calculated with dataset $\mathcal{D}_{\Pi^c}$. 
Write 
\begin{equation}
\begin{split}\label{equyyyuygygbh}
    \mathcal{L}_{\text{CV}}(\boldsymbol{\theta}; \bar{f}^*, \bar{\sigma}^*, \mathcal{D}_\Pi) &= - \ln p(Y_{\Pi}| \bar{f}^*, \bar{\sigma}^*, X_\Pi) \\
     &= \sum_{i } \frac{(Y_i -  \bar{f}^*_i\Delta t_i)^2}{2(\bar{\sigma}_i^{*2} \Delta t_i + \lambda )} + \frac{1}{2}\ln (\bar{\sigma}_i^{*2}\Delta t_i + \lambda),
\end{split}
\end{equation}
for the negative log-likelihood of the validation data $\mathcal{D}_\Pi$  given $\bar{f}^*, \bar{\sigma}^*$.
Our proposed cross-validation approach is then to select $\boldsymbol{\theta}^*$ as 
\begin{equation}
\begin{array}{ll}
& \boldsymbol{\theta}^{*}  = \underset{\boldsymbol{\theta}}{\arg\min}\,\,\mathbb{E}_{ (\Pi, \Pi^c) }\{ \mathcal{L}_{\text{CV}}(\boldsymbol{\theta}; \bar{f}^*, \bar{\sigma}^*, \mathcal{D}_\Pi)\} \\
\mathrm{subject\,to} & \bar{f}^*, \bar{\sigma}^* =\underset{\bar{f}, \bar{\sigma}}{\arg\min}\,\,\mathcal{L} (\mathcal{D}_{\Pi^c}, \bar{f}, \bar{\sigma})
\end{array}\label{eq: Method}
\end{equation}
Note that $\bar{f}^*, \bar{\sigma}^*$ are selected as  described in  Section~\ref{sec:section3}. 
In practice, the computation of the exact value of $\mathbb{E}_{ (\Pi,\Pi^c)}\{\mathcal{L}_{\text{CV}}(\boldsymbol{\theta}; \bar{f}^*, \bar{\sigma}^*, \mathcal{D}_\Pi\}$ is intractable. Therefore, we instead approximate $\mathbb{E}_{ (\Pi,\Pi^c)}\{ \mathcal{L}_{\text{CV}}\}$ with the empirical average
\begin{equation}\label{eqkjhgyugui}
    \frac{1}{M} \sum_{i=1}^{M} \mathcal{L}_{\text{CV}}(\boldsymbol{\theta}; \bar{f}_i^*, \bar{\sigma}_i^*, \mathcal{D}_{\Pi_i})
\end{equation}
where the $(\Pi_i,\Pi_i^c) $ are i.i.d. 
samples from $\mathbb{P}_{ (\Pi,\Pi^c)}$ and $\bar{f}_i^*, \bar{\sigma}_i^* =\underset{\bar{f}, \bar{\sigma}}{\arg\min}\,\,\mathcal{L}(\mathcal{D}_{\Pi_i^c}, \bar{f}, \bar{\sigma})$.  We use a gradient free  optimization algorithm to minimize \eqref{eqkjhgyugui} (see Subsection~\ref{secalalg}).
This algorithm only uses noisy observations \eqref{eqkjhgyugui} of the true loss.

The proposed cross-validation algorithm can be summarized as follows:

\begin{enumerate}
    \item Select a gradient-free optimization algorithm.
    \item At each iteration, given the hyperparameters $\boldsymbol{\theta}^k$, select $M$ divisions of the data $\mathcal{D}_\mathcal{T}$ into a training set $\mathcal{D}_{\Pi_i^c}$ and a validation set  $\mathcal{D}_{\Pi_i}$. 
    \item For each $1 \leq i \leq M$, recover $\bar{f}^*_i, \bar{\sigma}^*_i$ using training data $\mathcal{D}_{\Pi_i^c}$ using hyper-paramters $\boldsymbol{\theta}^k$.
    \item  For each $1 \leq i \leq M$, compute the loss $\mathcal{L}(\boldsymbol{\theta}^k; \bar{f}_i^*, \bar{\sigma}_i^*, \mathcal{D}_{\Pi_i})$ and  the empirical average  \eqref{eqkjhgyugui}.
    \item Minimize \eqref{eqkjhgyugui} with the gradient free optimization algorithm to select $\boldsymbol{\theta}^{k+1}$.
\end{enumerate}

\subsection{The active learning algorithm}\label{secalalg}

We choose the Bayesian optimization algorithm \cite{snoek2012practical}, where the loss function is modeled using a Gaussian Process with Matern kernel \cite{GPforML}, implemented in the scikit-optimize library in Python \cite{gp_min}. In our case, we set $M = 1$ when using gradient descent minimization and $M = 10$ when using Newton's method for minimization. The maximum number of iterations for the Bayesian optimization algorithm is set to $K = 75$ when using gradient descent minimization and $K =150$ when using Newton's method for minimization. While we can use a greater number of samples and a greater number of iterations with Newton's method because of its greater speed, in practice, the gradient descent method offered better performance in our tests.

\section{Experimental results}\label{sec: exp}

We first illustrate the effectiveness of our methods by considering two systems with non-linear drift or volatility. The two processes we consider are:
\begin{align}\label{eq: basis processes}
    dX_t &= \mu X_t dt + b\exp(-X_t^2) dW_t   &&\text{Exponential decay volatility.} \\
    dX_t &= \sin(2k\pi X_t) dt + b\cos(2k\pi X_t) dW_t   &&\text{Trigonometric process.}
\end{align}

In both cases, we generate trajectories using a Euler-Maruyama disretization. We use 500 points for training and 500 points for testing. 
\subsection{Metrics.}
To measure the performance of each method, we use three metrics. The first is the likelihood of the model given the data of the test set defined as 
\begin{equation*}
    \mathcal{L}(\bar{f}^*, \bar{\sigma}^*| X,Y) = -\log(p(Y | \bar{f}^*, \bar{\sigma}^*, X)) \propto \sum_{i=1}^M \frac{(Y_i -  \bar{f}^*\Delta t_i)^2}{2(\bar{\sigma}_i^{*2} \Delta t_i + \lambda )} + \frac{1}{2}\ln (\bar{\sigma}_i^{*2}\Delta t_i + \lambda).
\end{equation*}
\noindent The other two metrics are the relative error of the test drift and volatility at the test points:
\begin{align*}
    \delta_{f} = \frac{|| f -\bar{f} ||}{||f||} \\
    \delta_{\sigma} = \frac{||\sigma - \bar{\sigma} ||}{||\sigma||}
\end{align*}
\noindent where $f$ is the vector of drift values at the test points $(f)_i = f(X_i)$ and $\bar{f}$ is the vector of prediction $(\bar{f})_i = \bar{f}(X_i)$ (likewise for $\sigma, \bar{\sigma}$). Note that in practice, only $ \mathcal{L}(\bar{f}^*, \bar{\sigma}^*| (X,Y))$ may be computed without access to the true drift $f$ and true volatility $\sigma$. We still compute $\delta_{f}, \delta_{\sigma}$ to illustrate how a lower loss on the recovery of the drift and volatility yields a lower loss on the likelihood.

\subsection{Choice of kernels}
In all experiments, we use the Mat\'{e}rn Kernel \cite{GPforML} with smoothness parameter $\nu = \frac{5}{2}$  defined as 
\begin{equation}\label{eq: matern kernel}
    K_{\text{Mat\'{e}rn}}(x,y) = \sigma^2\Big(1 + \frac{\sqrt{5}||x -y ||}{l} + \frac{5||x -y ||^2}{3l^2}\Big)\exp\Big(- \frac{\sqrt{5}||x -y ||}{l}\Big).
\end{equation}
We learn the parameters $(\sigma, l)$ of the kernel (the smoothness parameter $\nu$ is not learned). Note that the Mat\'ern kernel is not well specified for many of the processes we will consider as the processes it defines are only 3 times differentiable (in the mean square sense \cite{GPforML}) and hence it is overly general. A better specified of kernel can in many cases significantly improve the performance. We choose this kernel because it is very general and allows to illustrate how one can obtain good results with few assumptions and little domain knowledge.  

\subsection{Benchmarks.}\label{sec: benchmarks details}
We compare our method and optimized parameters with two baselines. A first baseline that uses our method and unoptimized parameters. For kernels using a lengthscale (such as the  Mat\'{e}rn kernel \eqref{eq: matern kernel}), it is set to be the average $\ell^2$ distance between data points:
\begin{equation*}
    l_{\text{unopt}} = \frac{1}{N(N-1 )}\sum_{i \neq j} \norm{X_i - X_j}.
\end{equation*}
All other paramters are set to $1.0$. This method is labeled as \say{non-learned kernel}. 

The second baseline does not use our method to recover the drift and diffusion separately. Instead we assume that $Y$ is the sum of two Gaussian processes:
\begin{equation}
    y(x)= \xi(x) + W(x)
\end{equation}
where $\xi$ is a smooth Gaussian process (such as a Gaussian process with Mat\'{e}rn covariance function) and $W$ is a white noise Gaussian process with covariance matrix
\begin{equation}
    K_{\text{wn}}(x_i, x_j) = c\delta(x_i -x_j).
\end{equation}
 The posterior distribution $y$ conditioned on the data provides a prediction for the conditional mean (the drift of the SDE) and the conditional variance (the volatility of the SDE)\footnote{Note that this is the usual Gaussian process regression method}.The parameters of the kernel are optimized through the minimization of the negative log marginal likelihood. This method is labeled as \say{benchmark} and uses the implementation present in \cite{scikit-learn} and the details are presented in Appendices \ref{app:gaussian_regression} and \ref{app:logmarginal}. Note that a major drawback of this method is the assumption that the noise $\sigma(X_t)$ is identically distributed Gaussian noise, modeled through the white noise kernel $\delta(x_i -x_j)$. This assumption is valid for only a restricted class of SDEs. 

\subsection{Exponential decay volatility.}
The discretization of the exponential decay volatility is given by
\begin{equation}\label{eq: exp decay discrete}
    X_{n+1} - X_n =  \mu X_n\Delta t + b\exp(-X_n^2)\sqrt{\Delta t}\xi_n, \quad X_0 = x_0,
\end{equation}
with timestep $\Delta t = 0.01$. Observe that this process is pushed towards the origin by its drift value, where the volatility is maximized. Moreover, the volatility decreases exponentially fast away from the origin. We generate two trajectories with the same drift parameter $\mu = 5$ and different volatility parameter $b = 0.5, 1.0$. The results for the learned and unlearned kernel are reported in table \ref{table: exp vol results}. The training and testing data are illustrated in figure \ref{fig: data, exp vol 2} and the prediction of the drift and volatility are presented in figure \ref{fig: pred, exp vol 2} (see also figures \ref{fig: data, exp vol 1}, \ref{fig: pred, exp vol 1} in the appendix).

\begin{table}[h]
\begin{center}
\begin{tabular}{|l|c|c|c|c|c|c|} 
 \hline
 & \multicolumn{3}{|c|}{Trajectory 1} & \multicolumn{3}{|c|}{Trajectory 2} 
  \\ 
 \hline
  &$\mathcal{L}(\bar{f}^*, \bar{\sigma}^* |X, Y)$ & $\delta_f$ & $\delta_\sigma$  &$\mathcal{L}(\bar{f}^*, \bar{\sigma}^* |X, Y)$ &$\delta_f$& $\delta_\sigma$\\ 
     \hline
Benchmark   & \textbf{-2.547} & 0.394 & \textbf{0.033}&  -1.887 & 0.322& 0.085 \\
   \hline
Non-learned kernel   & -2.532 & 0.506 & 0.081 &  -1.887 & 0.472& 0.062 \\
 \hline
Learned kernel   &  -2.546  & \textbf{0.388} & 0.048& \textbf{-1.900} & \textbf{0.249}& \textbf{0.035} \\
\hline
\end{tabular}
\caption{Results for the exponential volatility process.}
\label{table: exp vol results}
\end{center}
\end{table}
\begin{figure}[H]
    \centering
    \includegraphics[width = 0.23\linewidth, height = 3cm]{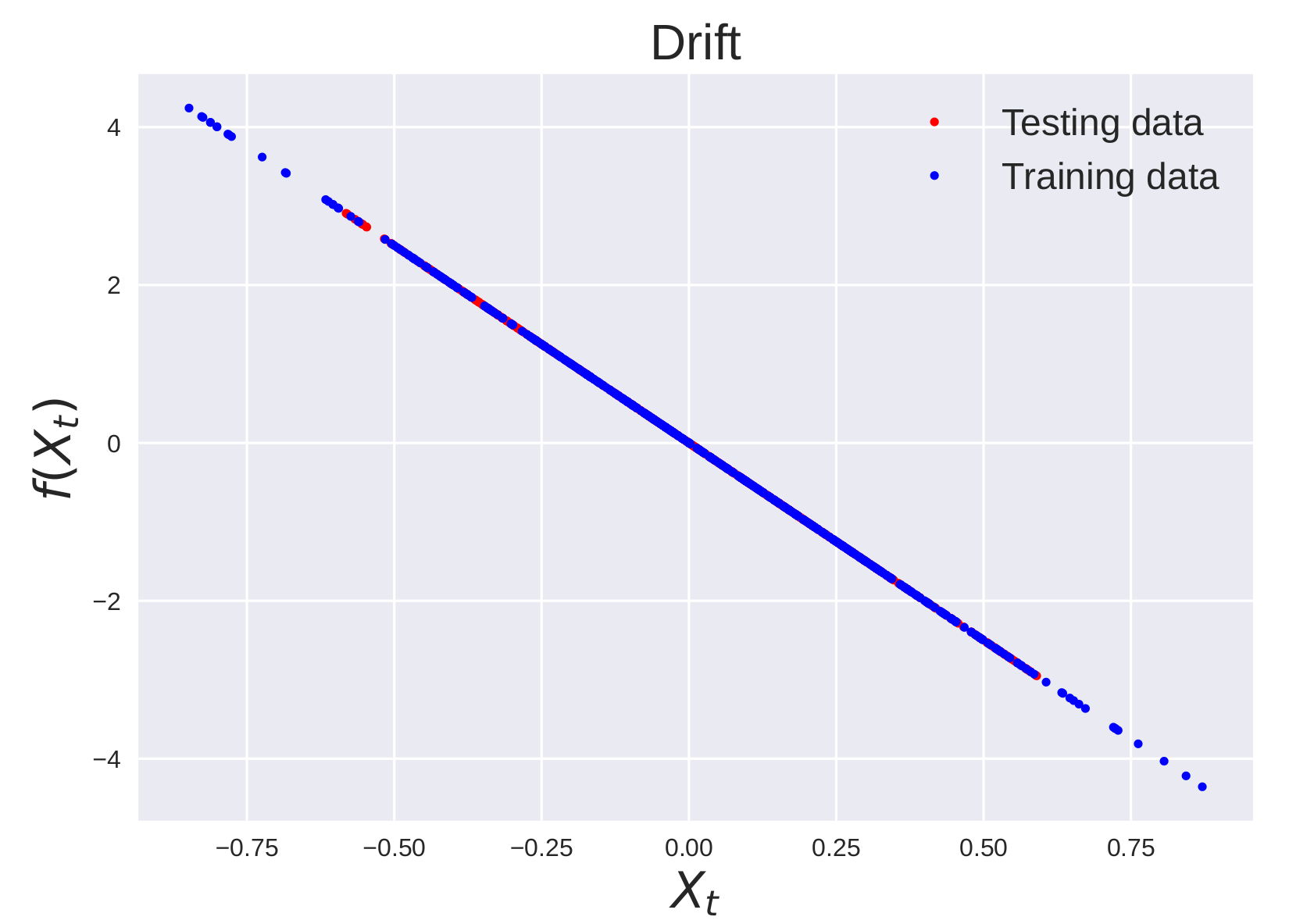}
    \includegraphics[width = 0.23\linewidth, height = 3cm]{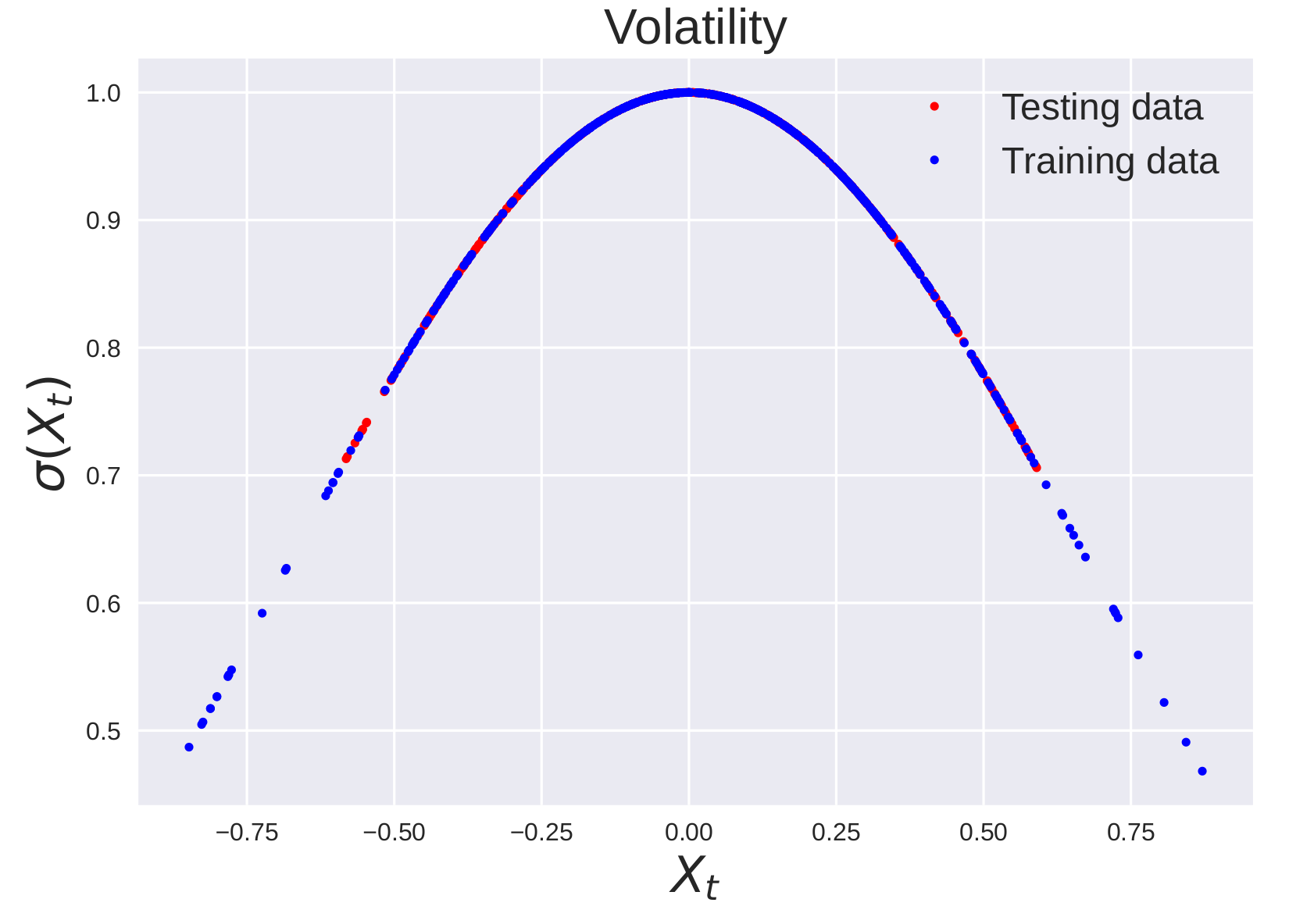}
    \includegraphics[width = 0.23\linewidth, height = 3cm]{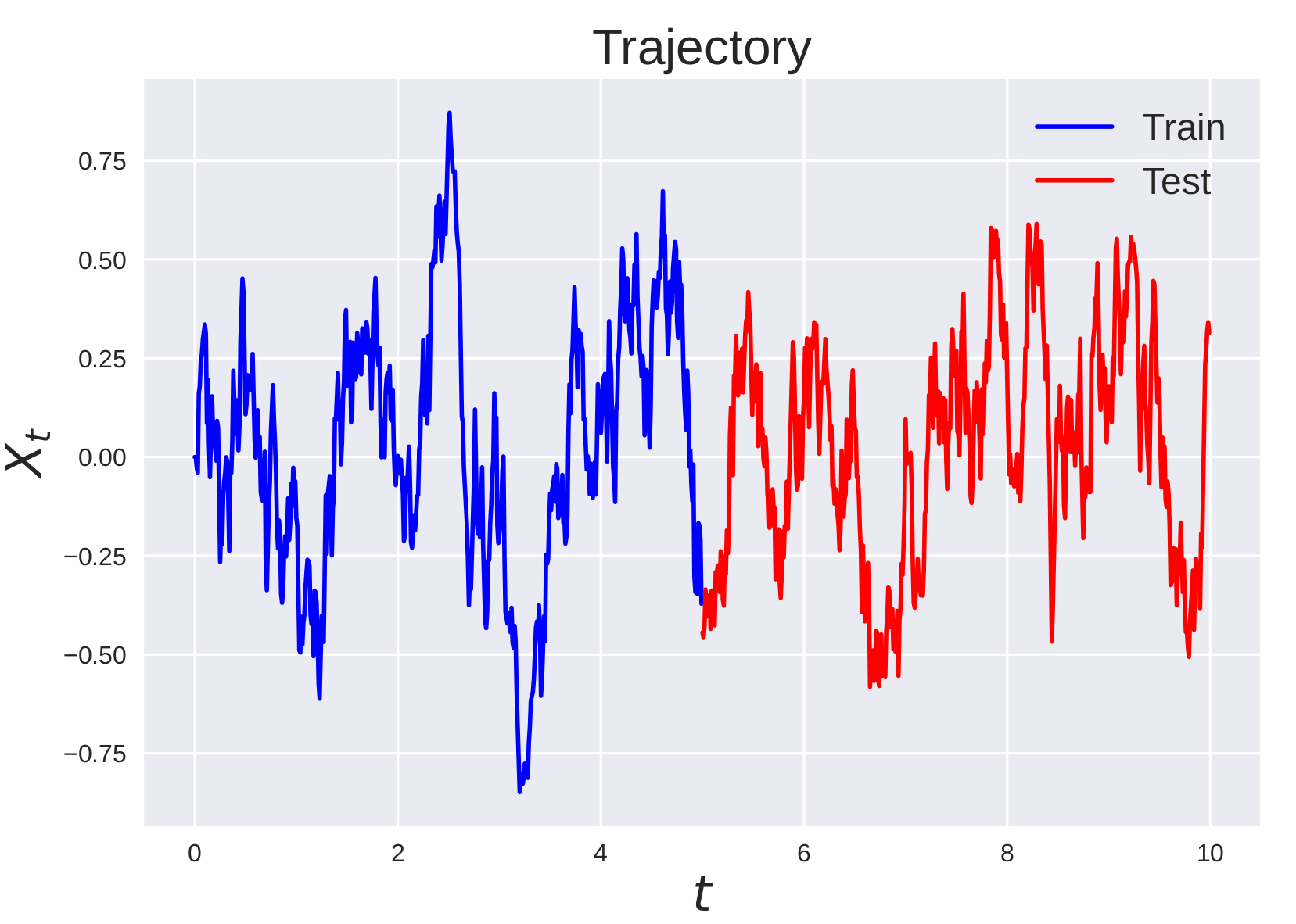}
    \includegraphics[width = 0.23\linewidth, height = 3cm]{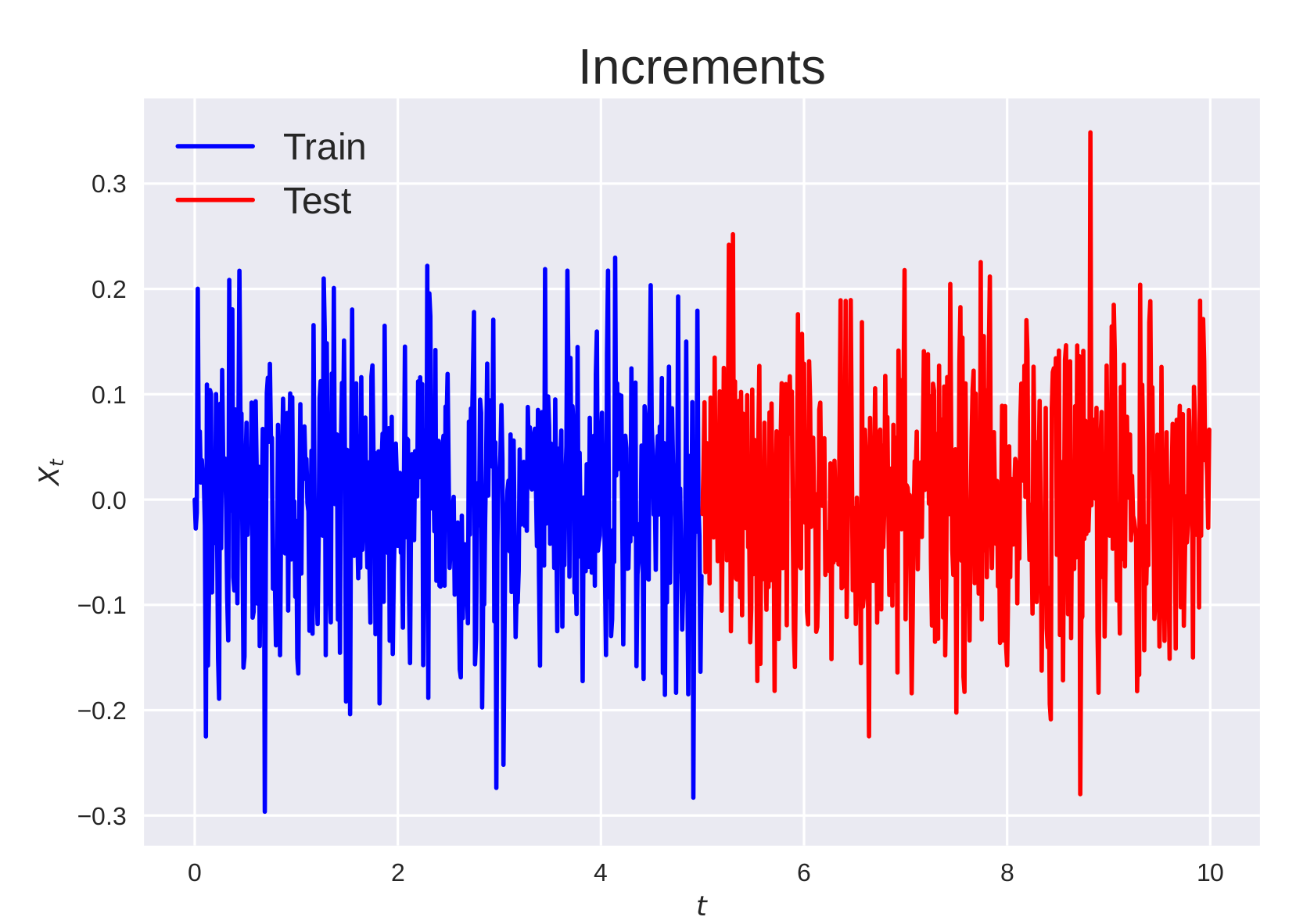}
    \caption{From left to right: drift function, volatility function, sample trajectory and sample increments of the exponential volatility process.}
    \label{fig: data, exp vol 2}
\end{figure}
\begin{figure}[H]
    \centering
    \includegraphics[width = 0.23\linewidth, height = 3cm]{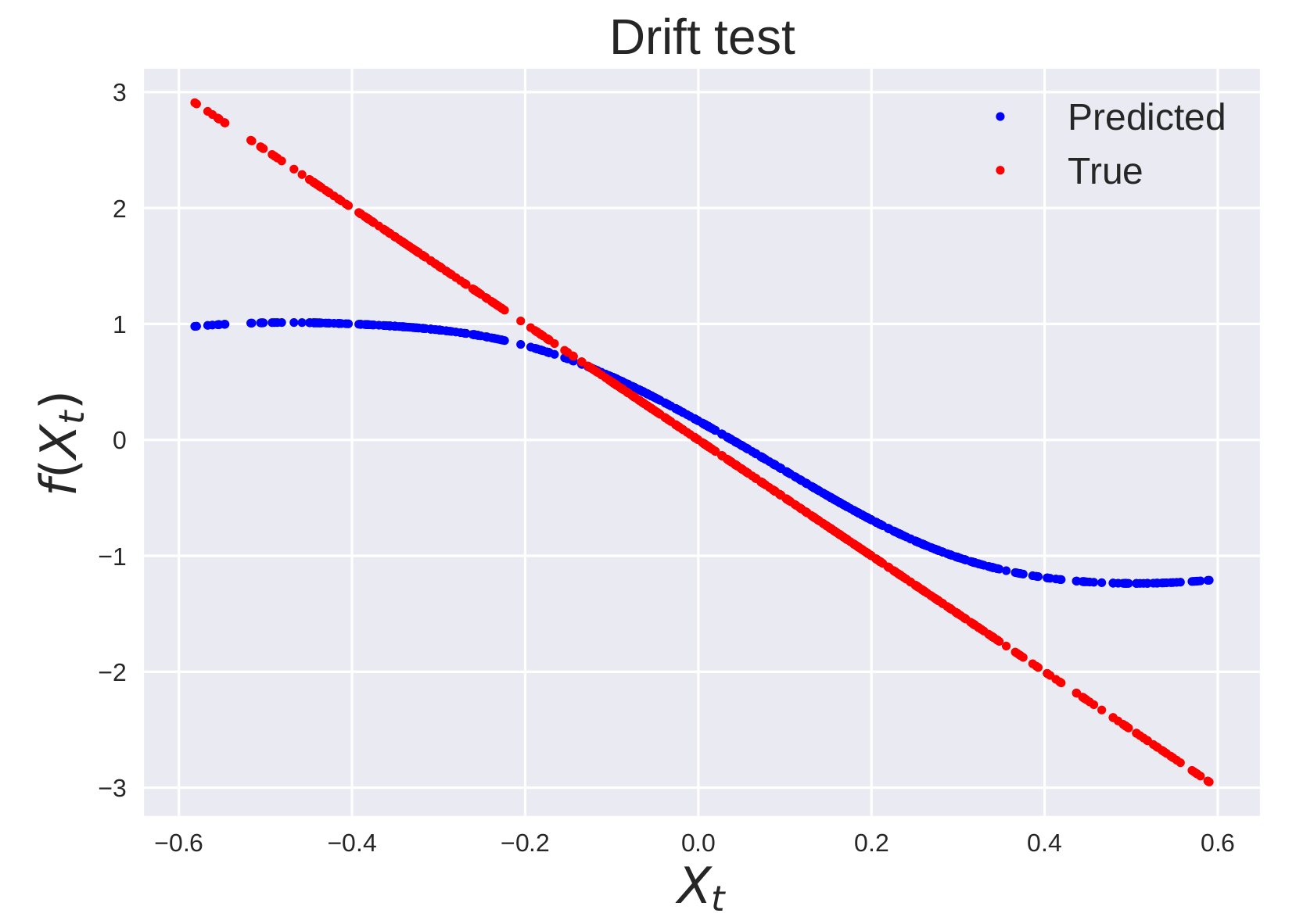}
    \includegraphics[width = 0.23\linewidth, height = 3cm]{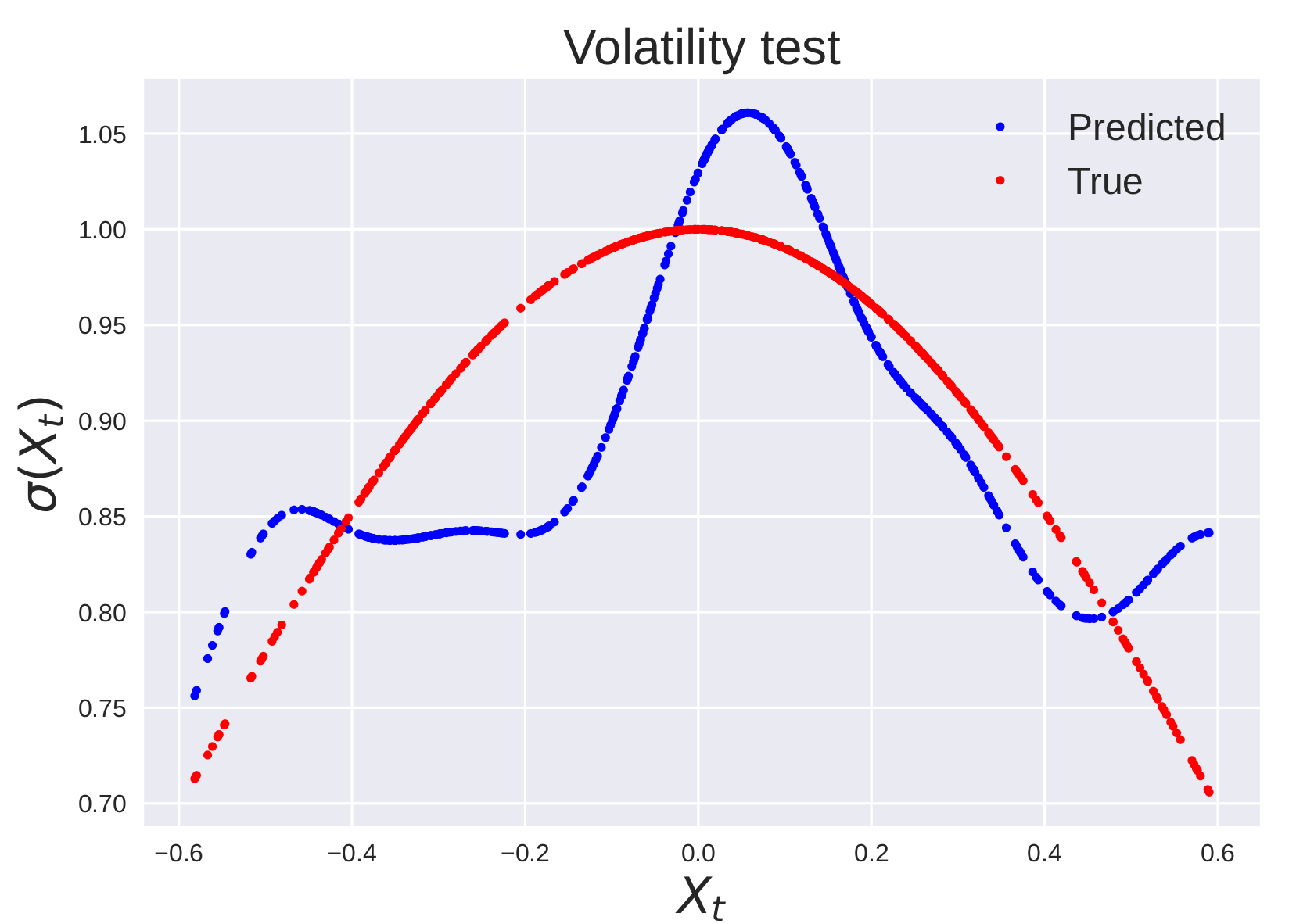}
    \includegraphics[width = 0.23\linewidth, height = 3cm]{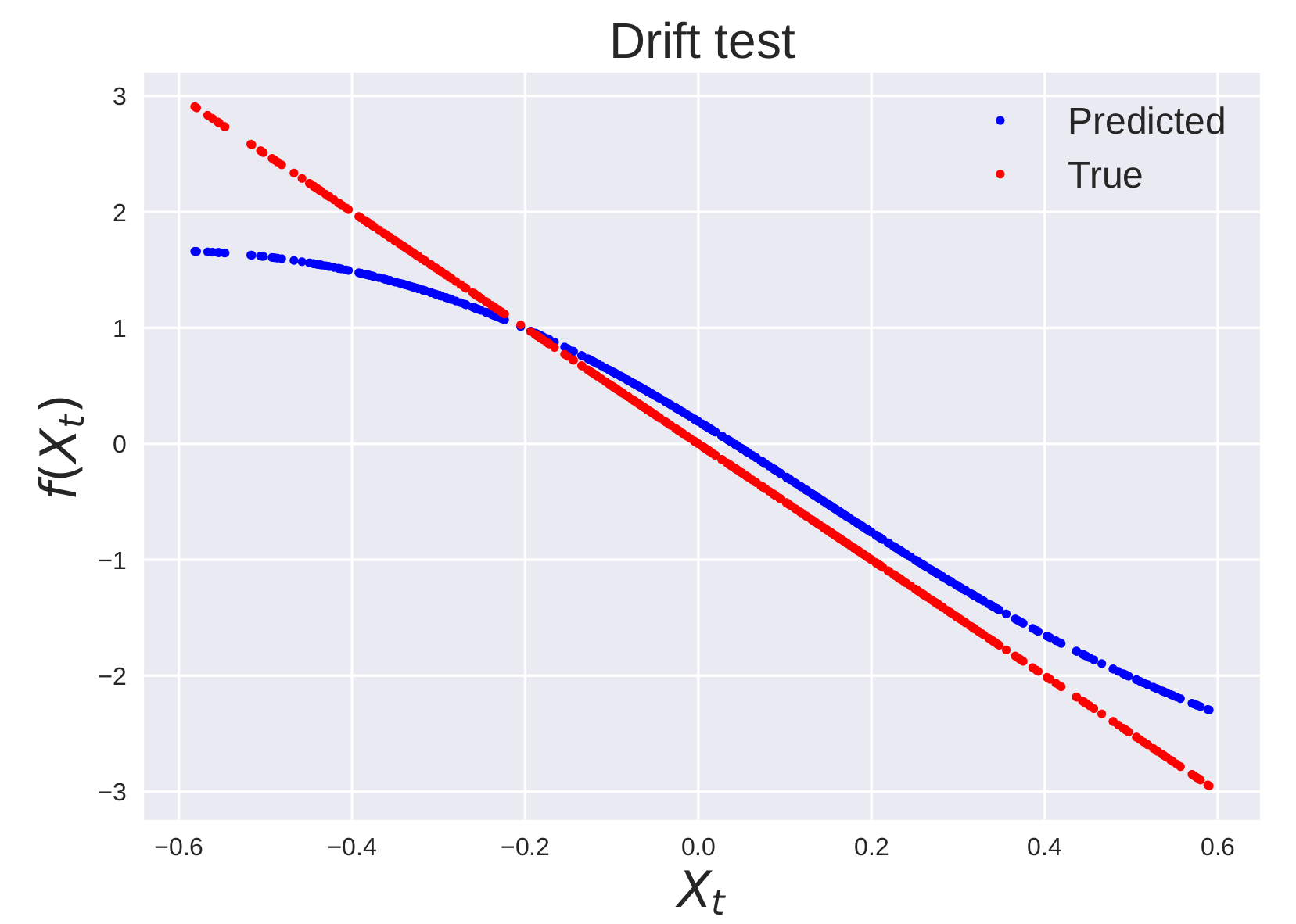}
    \includegraphics[width = 0.23\linewidth, height = 3cm]{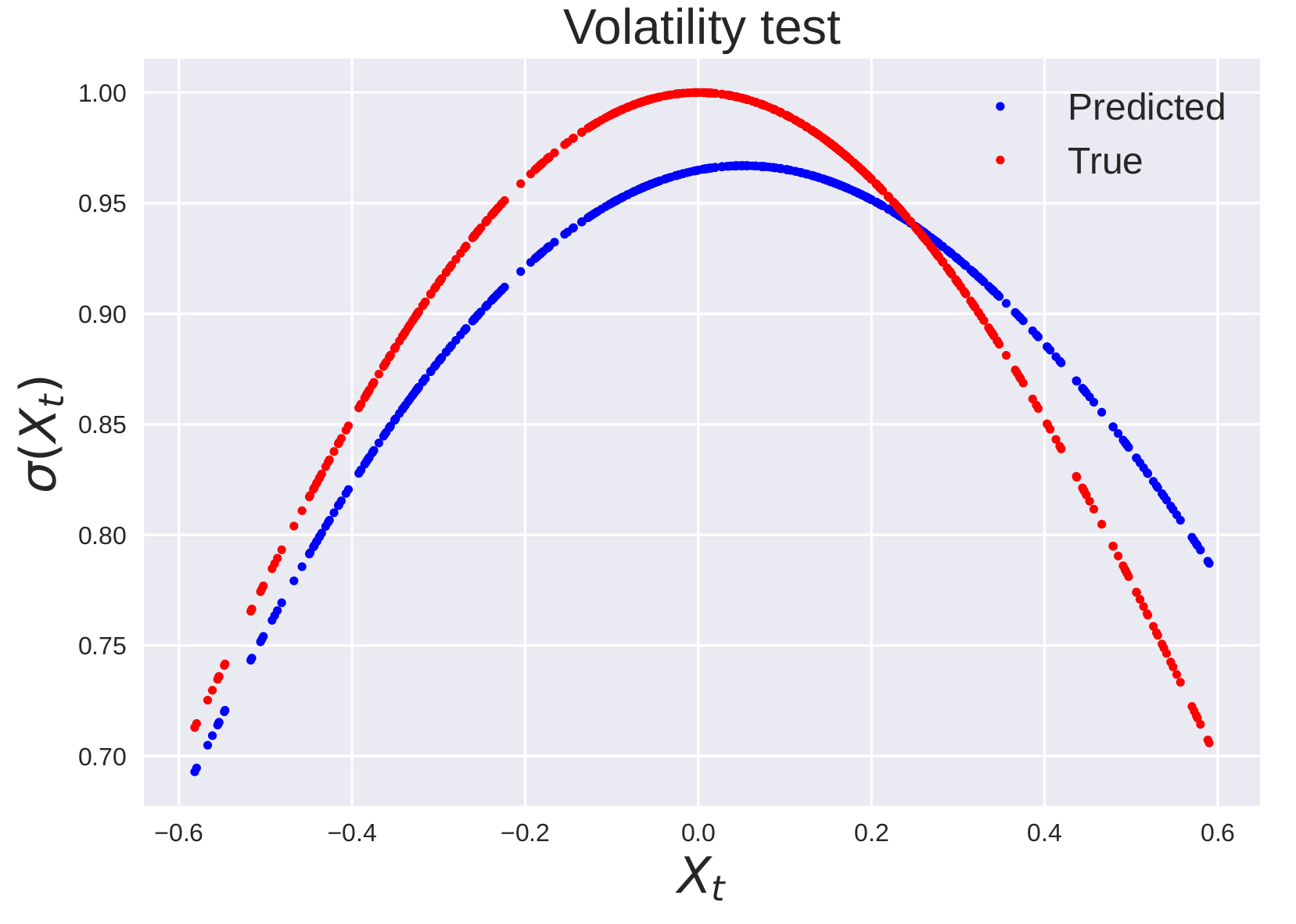}
    \caption{Predicted drift and volatility on the testing set for trajectory 2 of the exponential volatility process. From left to right: drift (non-learned kernel), volatility (non-learned kernel), drift (learned kernel), volatility (learned kernel) .}
    \label{fig: pred, exp vol 2}
\end{figure}

\subsection{Trigonometric process}
The discretization of the trigonometric process is given by
\begin{equation}\label{eq: trig process discrete}
    X_{n+1} - X_n =  \sin(2k\pi X_t)\Delta t + b\cos(2k\pi X_t)\sqrt{\Delta t}\xi_n, \quad X_0 = x_0.
\end{equation}
In this case, both the drift and volatility functions are non-linear functions of $X_t$. We generate two trajectories with volatility parameter $b = 1.0, 0.5$. For $b = 0.5$, we set $\Delta t = 0.01$. For the second trajectory, $b = 0.5$ and the timestep is set $\Delta t = 0.001$. In this case, the testing data contains points outside the training distribution. Hence the problem of prediction is more challenging than the problem of recovery at the training points. The results for the learned and unlearned kernel are reported in table \ref{table: trig results}. The training and testing data are illustrated in figure \ref{fig: data, trig  1}. For trajectory 2, the optimization of the hyper-parameters of the kernel yield a better recovery and a better prediction of the volatility outside the training distribution \ref{fig: data, trig 2}, \ref{fig: pred, trig 2} (also see figures \ref{fig: pred, trig 2} in the appendix). 
\begin{table}[H]
\begin{center}
\begin{tabular}{|l|c|c|c|c|c|c|} 
 \hline
 & \multicolumn{3}{|c|}{Trajectory 1} & \multicolumn{3}{|c|}{Trajectory 2} 
  \\ 
 \hline
  &$\mathcal{L}(\bar{f}^*, \bar{\sigma}^* |X, Y)$ & $\delta_f$ & $\delta_\sigma$  &$\mathcal{L}(\bar{f}^*, \bar{\sigma}^* |X, Y)$ &$\delta_f$& $\delta_\sigma$\\
  \hline
  Benchmark  & -3.411 &  0.327 & 0.442 &\textbf{ -3.808} & 3.454&\textbf{ 0.198}\\
   \hline
Non-learned kernel   & -3.675 &  \textbf{0.157} & 0.093 & -2.405  & \textbf{0.832}& 0.665 \\
 \hline
Learned kernel   &  \textbf{-3.678}  & 0.269 & \textbf{0.088}& -3.642 & 1.481& 0.242 \\
\hline
\end{tabular}
\caption{Results for the trigonometric process.}
\label{table: trig results}
\end{center}
\end{table}

\begin{figure}[H]
    \centering
    \includegraphics[width = 0.23\linewidth, height = 3cm]{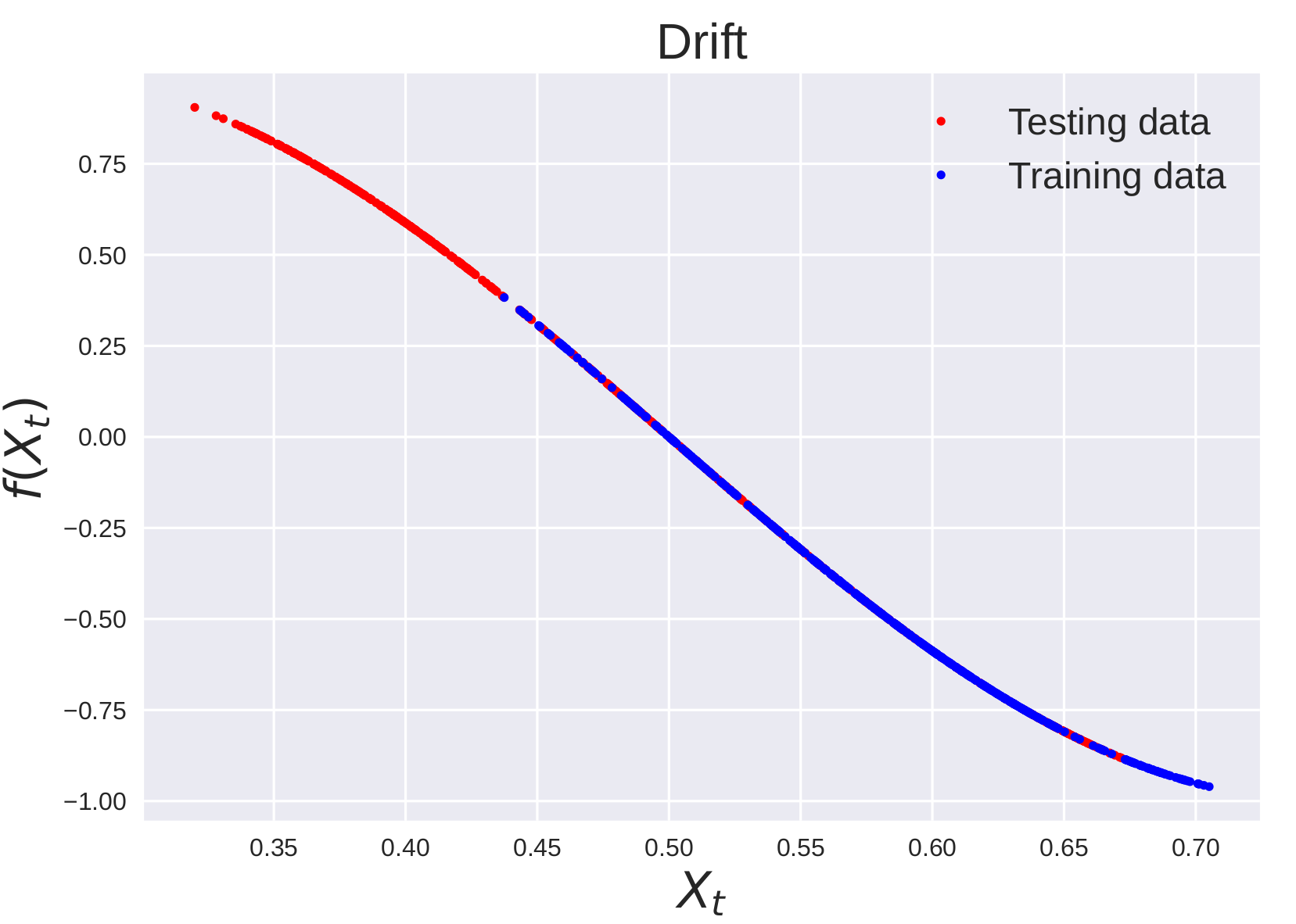}
    \includegraphics[width = 0.23\linewidth, height = 3cm]{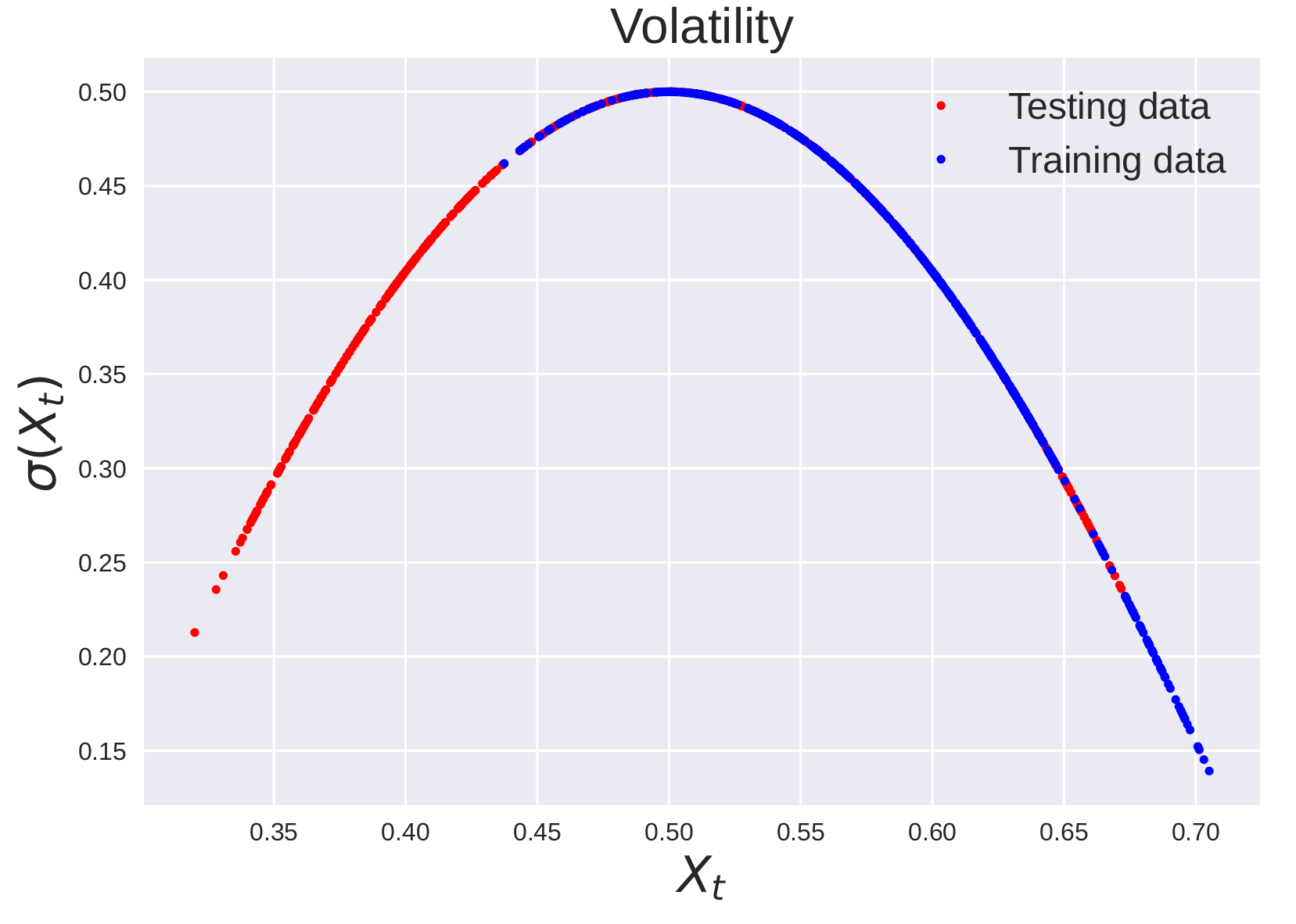}
    \includegraphics[width = 0.23\linewidth, height = 3cm]{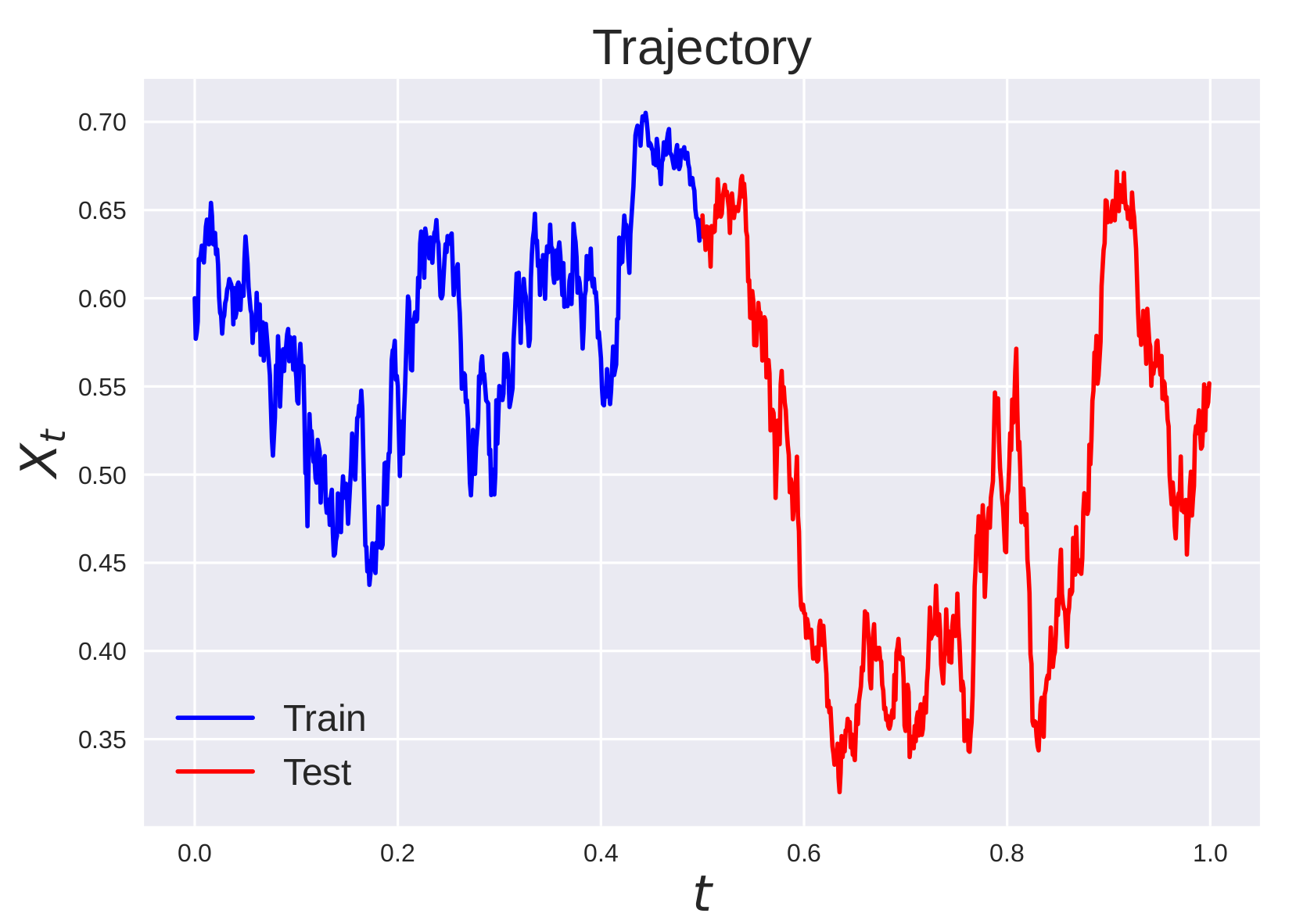}
    \includegraphics[width = 0.23\linewidth, height = 3cm]{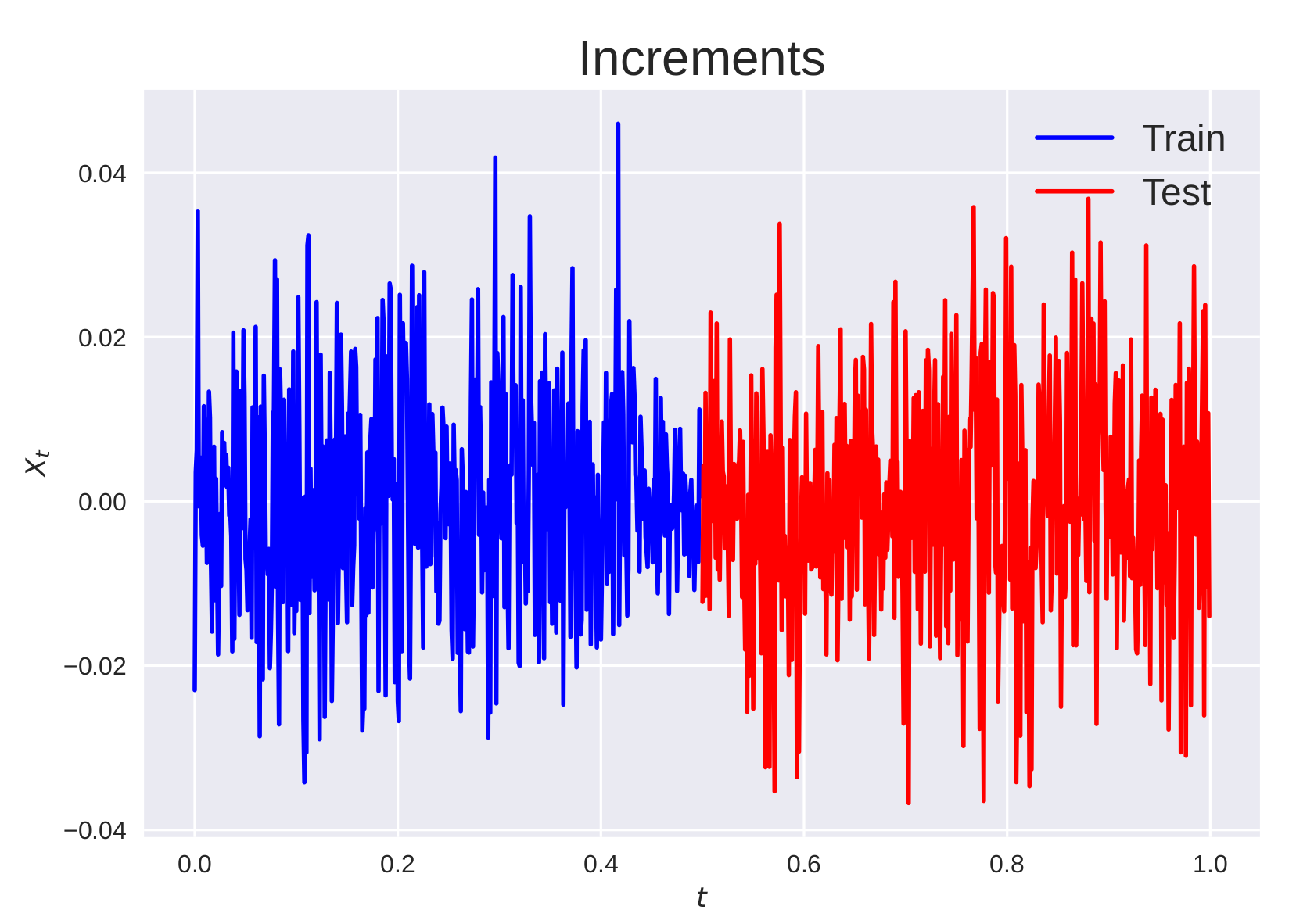}
    \caption{From left to right: drift function, volatility function, sample trajectory and sample increments of the trigonometric process (trajectory 2).}
    \label{fig: data, trig 2}
\end{figure}

\begin{figure}[H]
    \centering
    \includegraphics[width = 0.23\linewidth, height = 3cm]{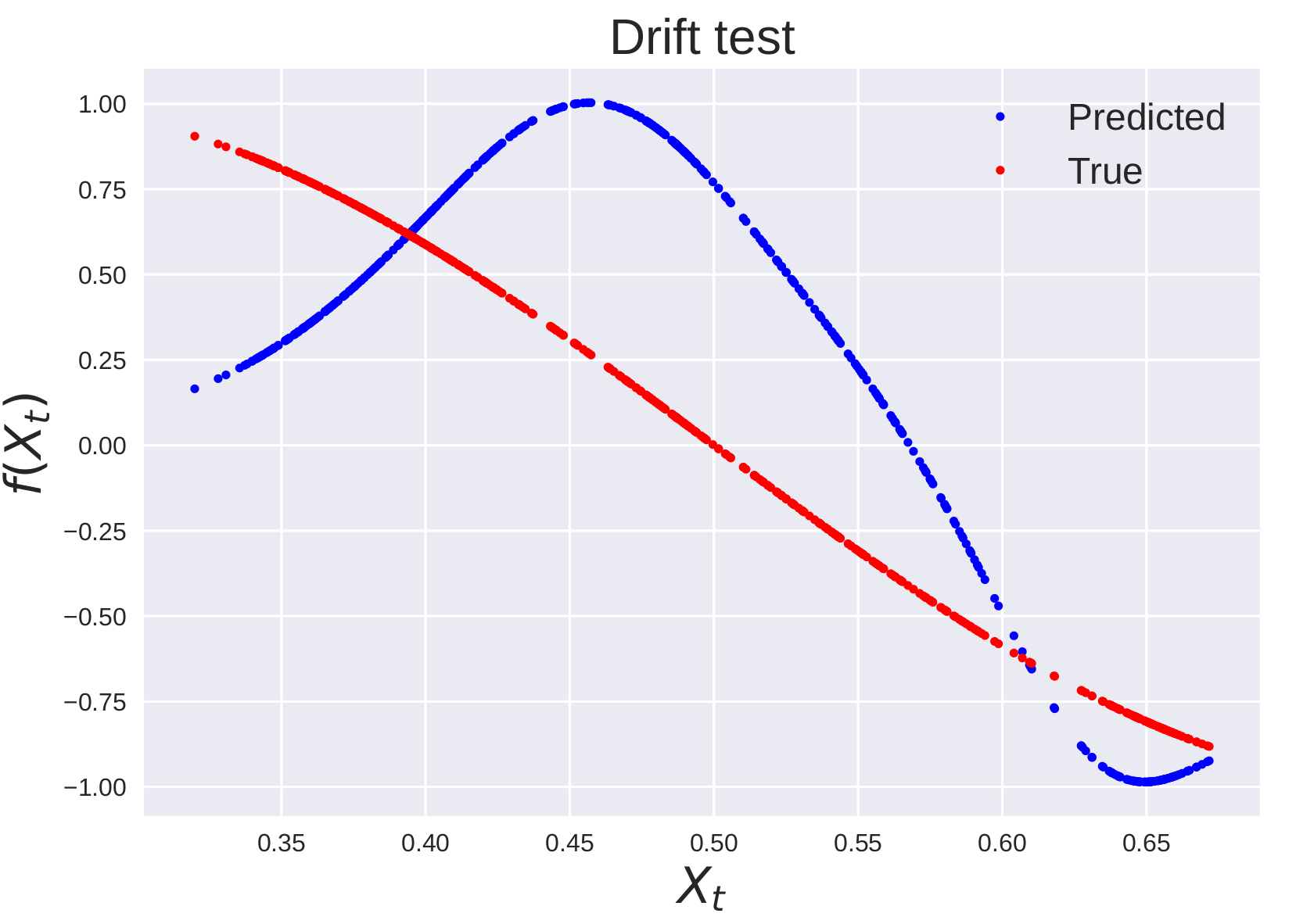}
    \includegraphics[width = 0.23\linewidth, height = 3cm]{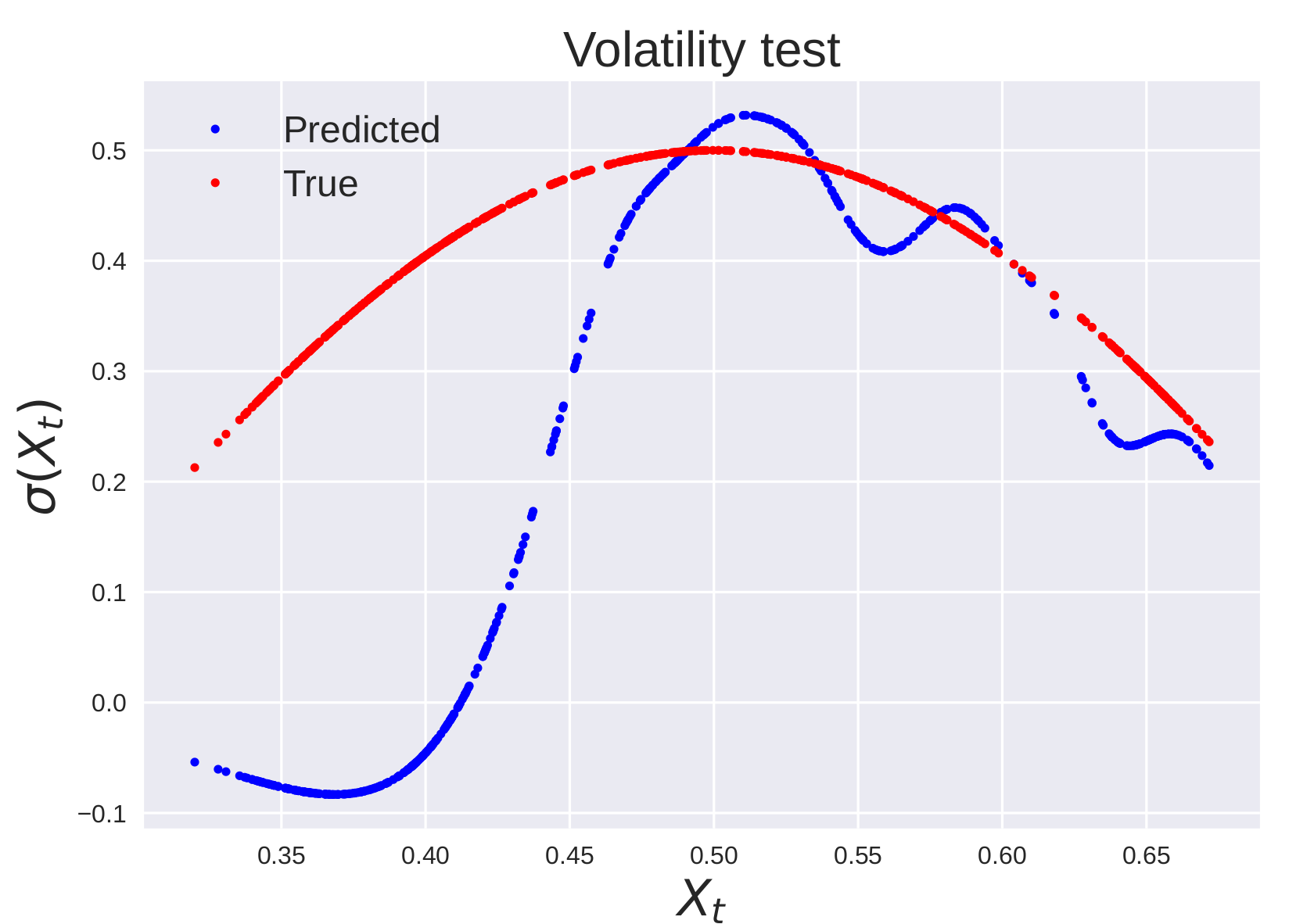}
    \includegraphics[width = 0.23\linewidth, height = 3cm]{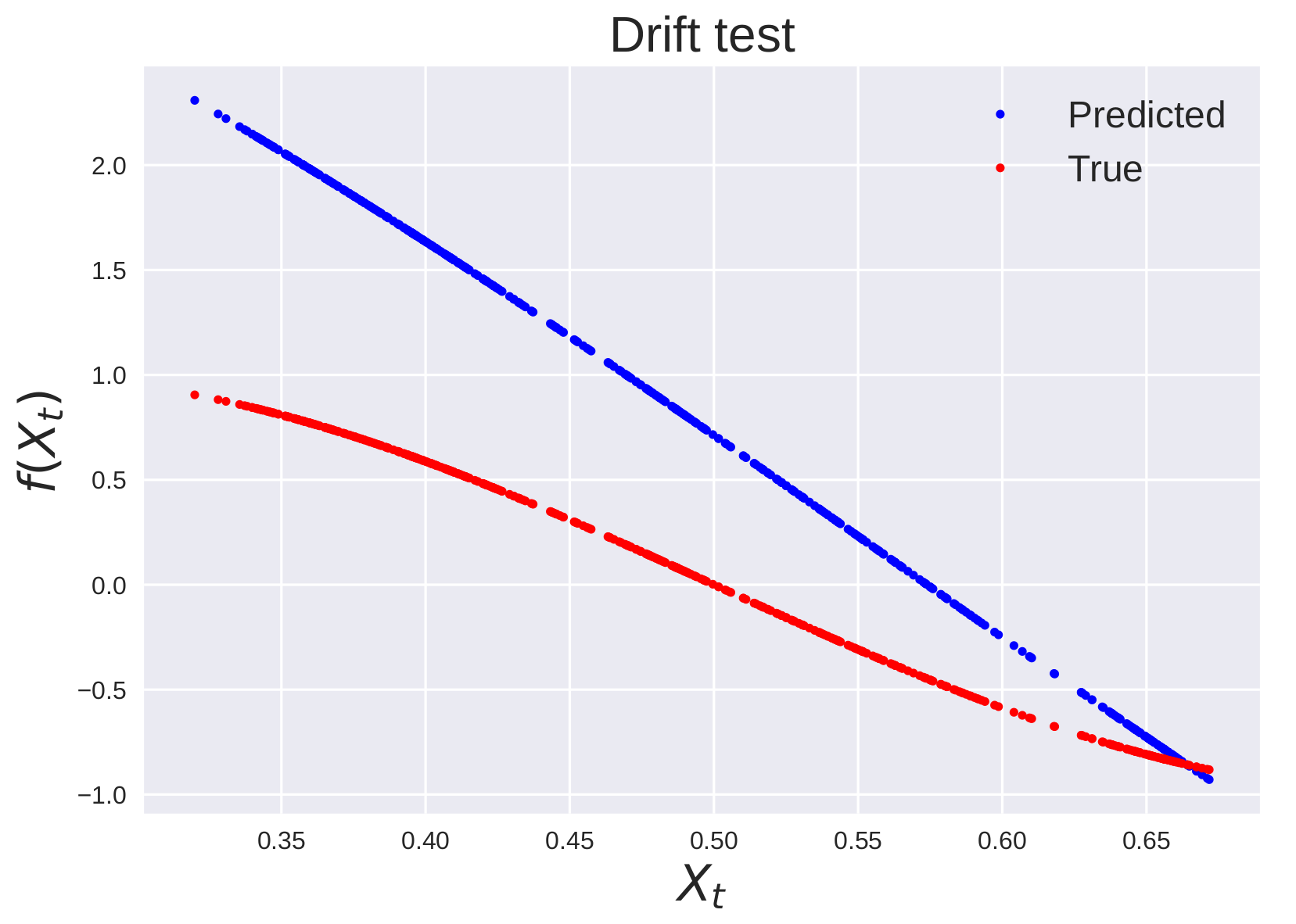}
    \includegraphics[width = 0.23\linewidth, height = 3cm]{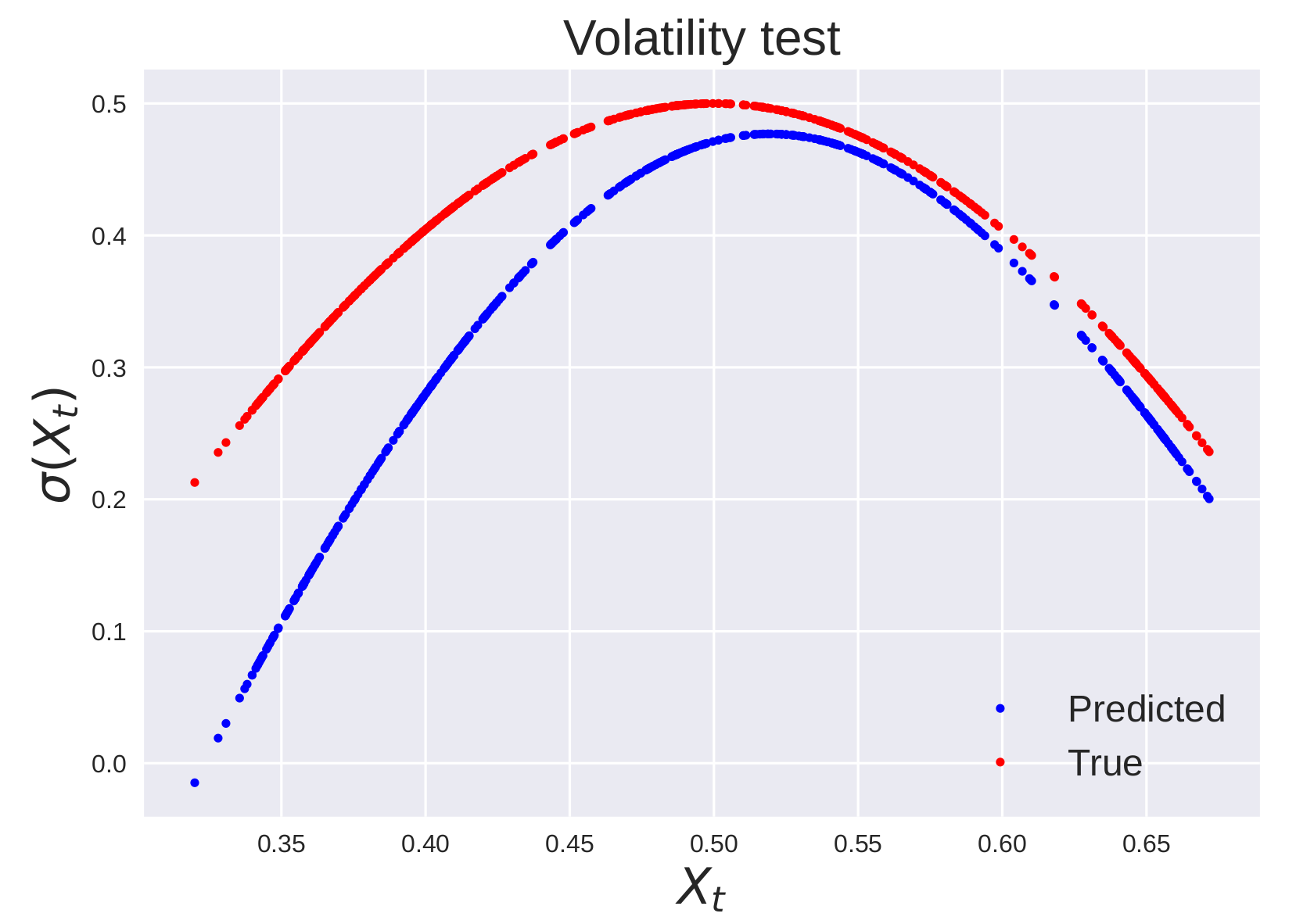}
    \caption{Predicted drift and volatility on the testing set for trajectory 2 of the trigonometric process. From left to right: drift (non-learned kernel), volatility (non-learned kernel), drift (learned kernel), volatility (learned kernel) .}
    \label{fig: pred, trig 2}
\end{figure}

\subsubsection{Benchmark comparison}

Generally our method with optimized parameters outperforms our selected benchmark both in recovery of the drift and the volatility as measured by our metric. Because the standard Gaussian process regressor uses a white noise kernel, it generally does not capture well a non-constant volatility. Our method on the other hand is able to capture non-constant volatility models. Compare for example figures \ref{fig: pred, trig 2} and \ref{fig: benchmark, trig}.
\begin{figure}[h]
    \centering
    \includegraphics[width = 0.23\linewidth, height = 3cm]{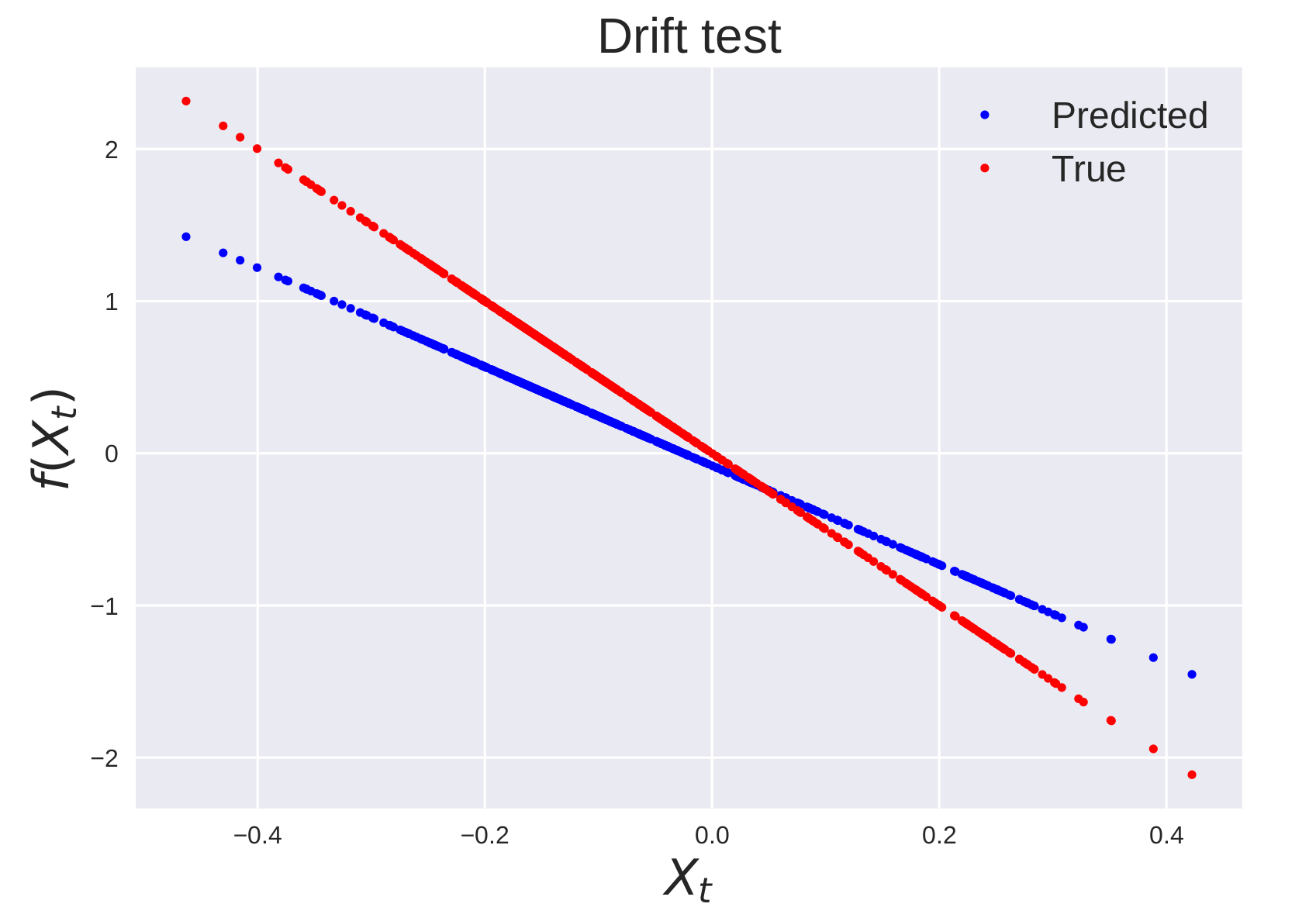}
    \includegraphics[width = 0.23\linewidth, height = 3cm]{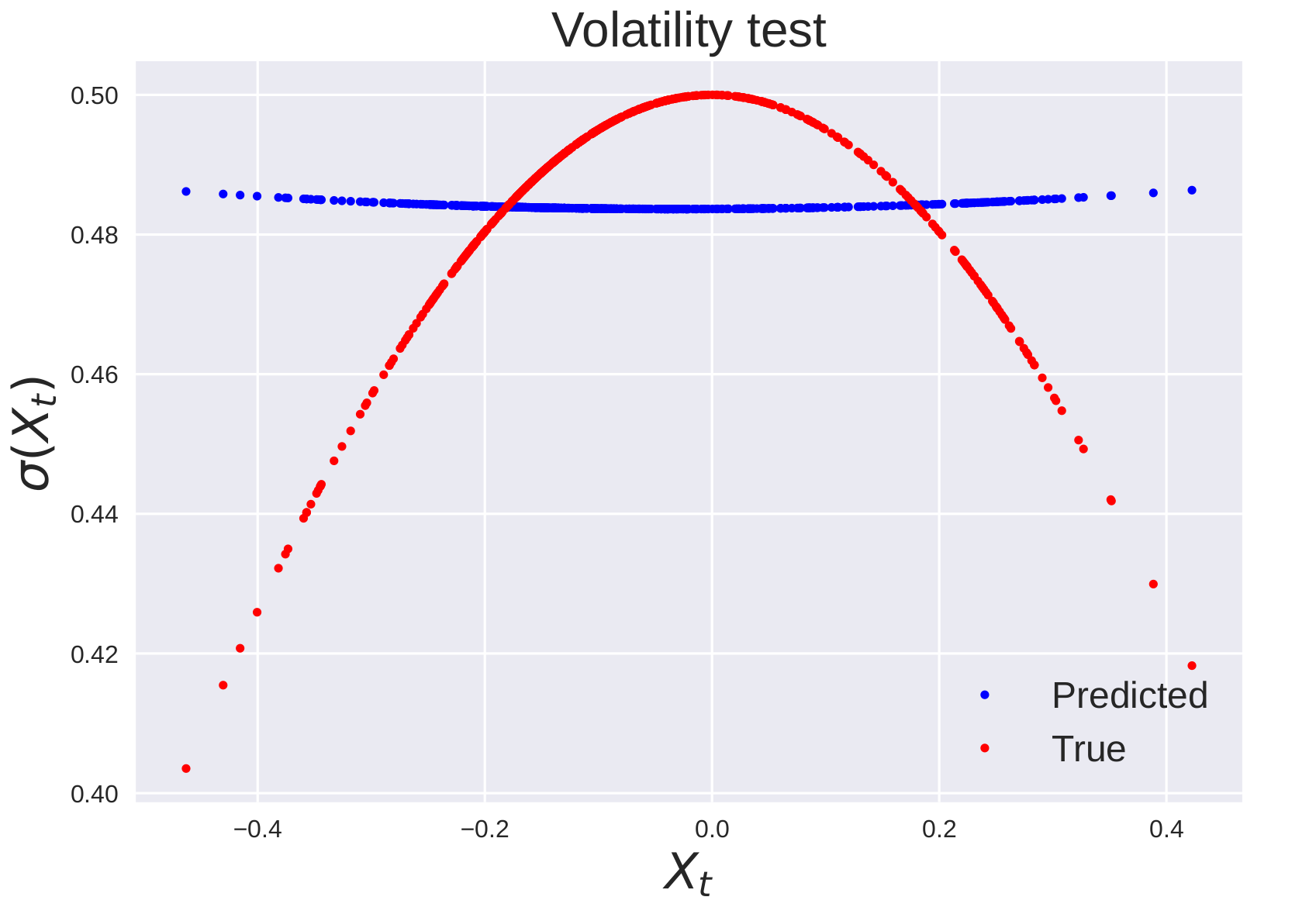}
    \includegraphics[width = 0.23\linewidth, height = 3cm]{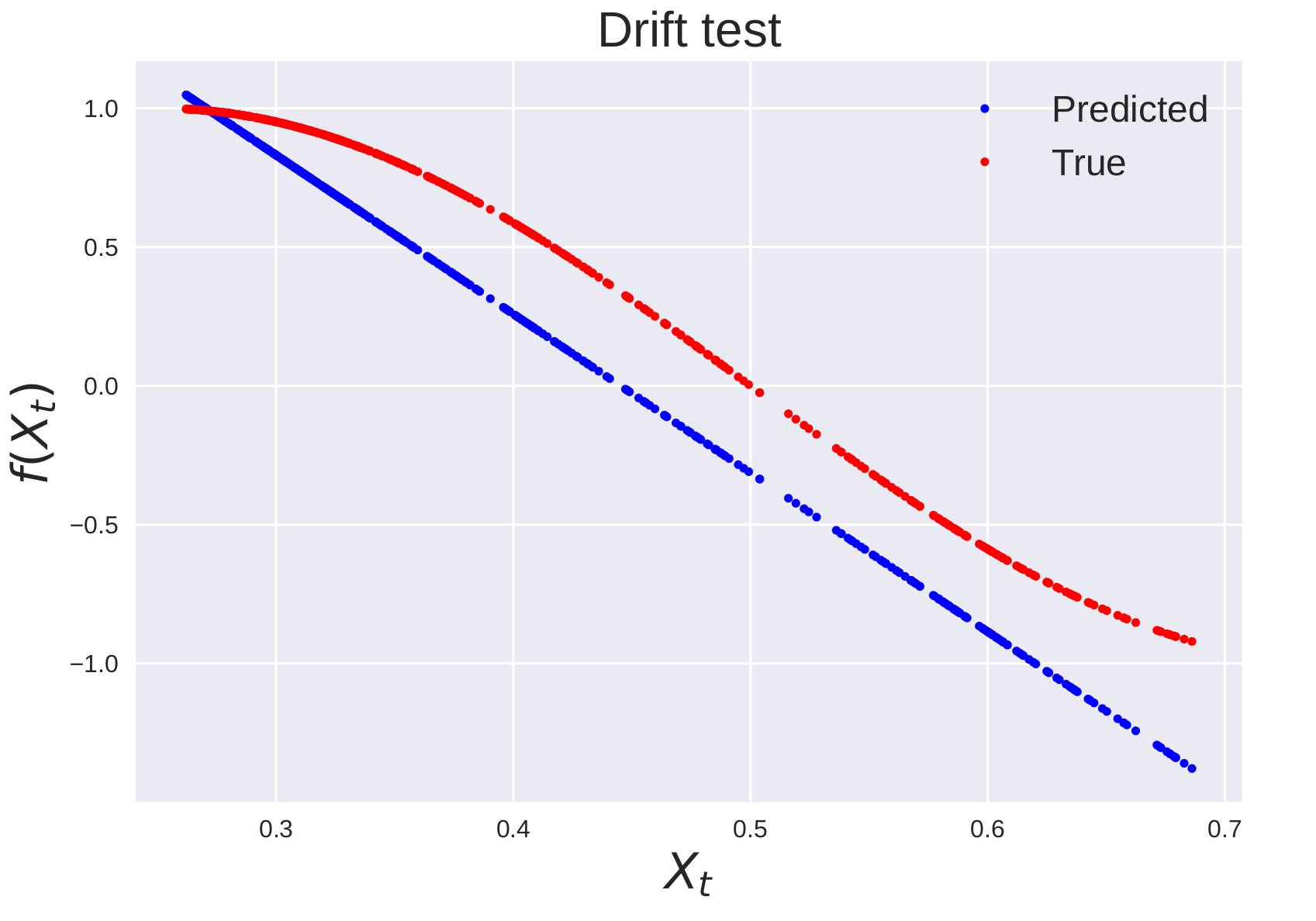}
    \includegraphics[width = 0.23\linewidth, height = 3cm]{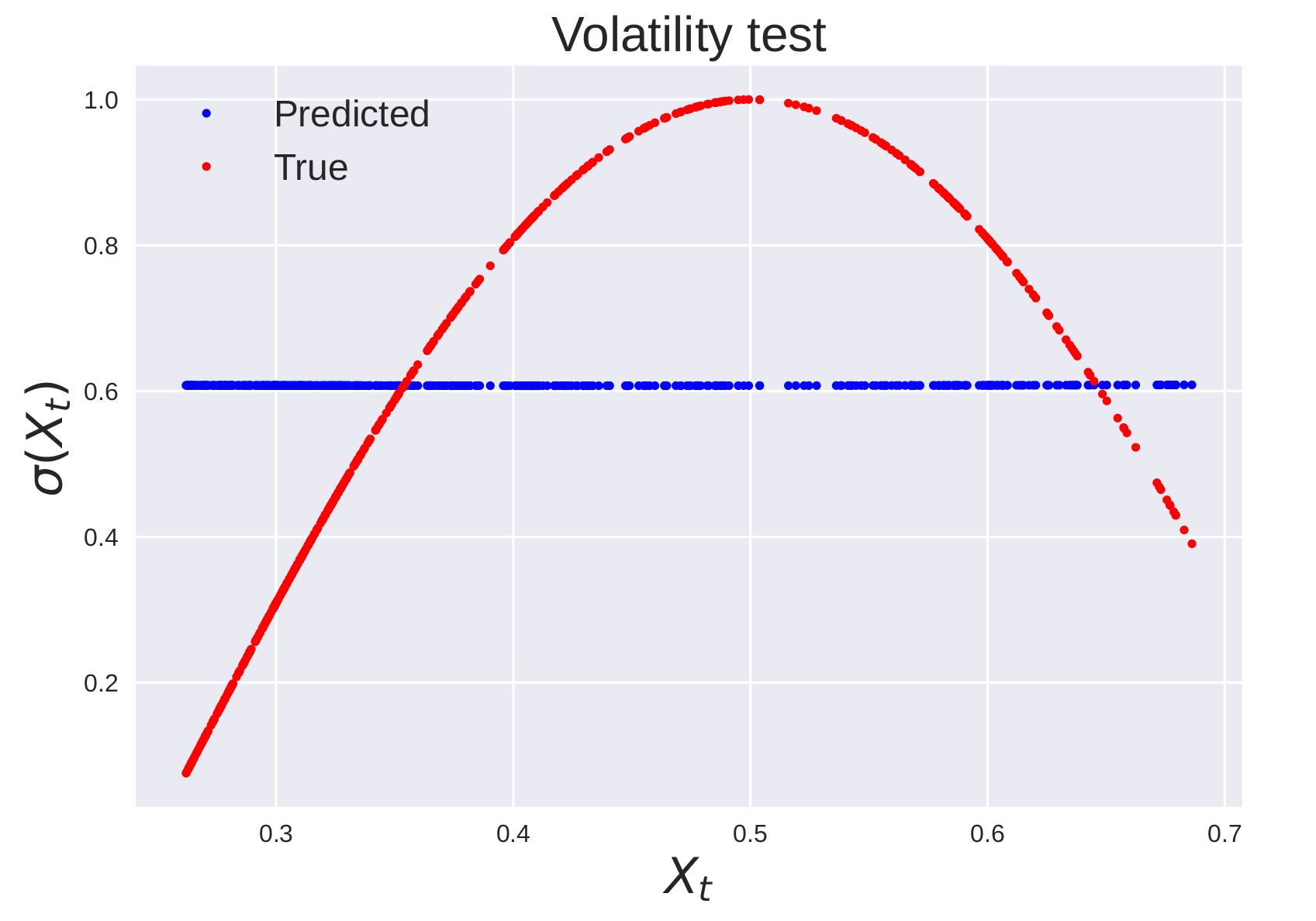}
    \caption{Exponential volatility process benchmark prediction. From left to right: drift (trajectory 1), volatility (trajectory 1), drift (trajectory 2), volatility (trajectory 2).}
    \label{fig: benchmark, exp vol}
\end{figure}

\begin{figure}[h]
    \centering
    \includegraphics[width = 0.23\linewidth, height = 3cm]{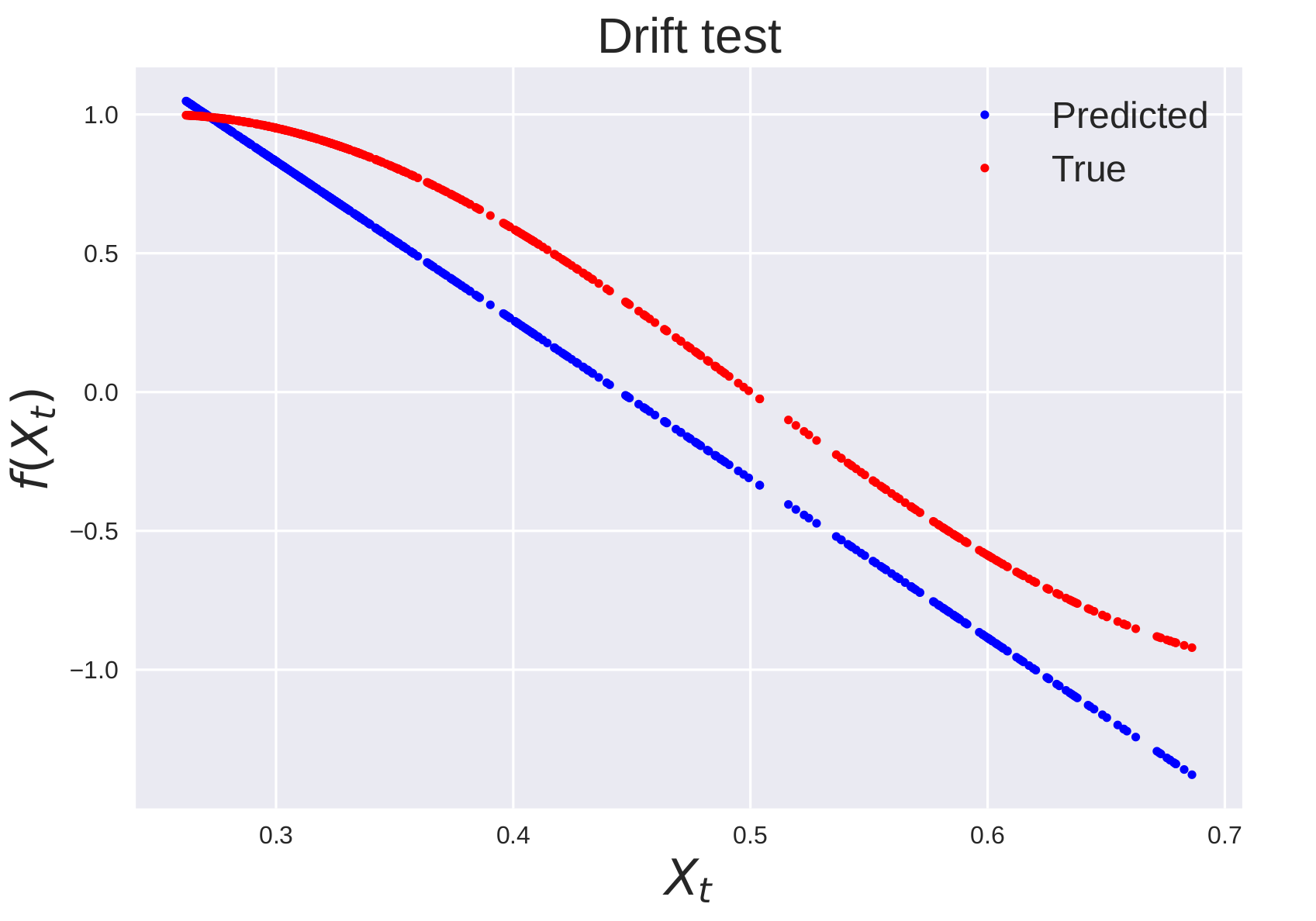}
    \includegraphics[width = 0.23\linewidth, height = 3cm]{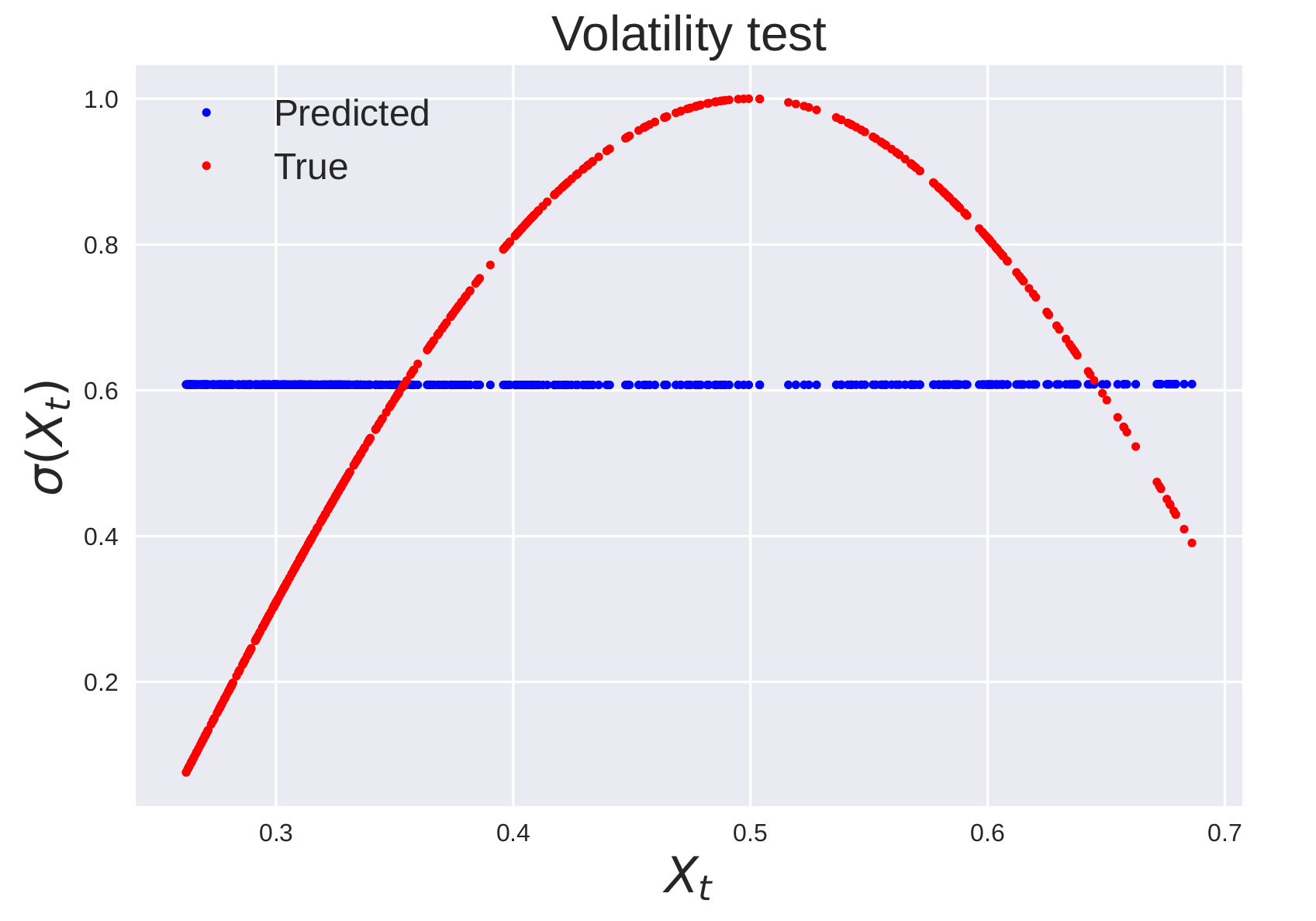}
    \includegraphics[width = 0.23\linewidth, height = 3cm]{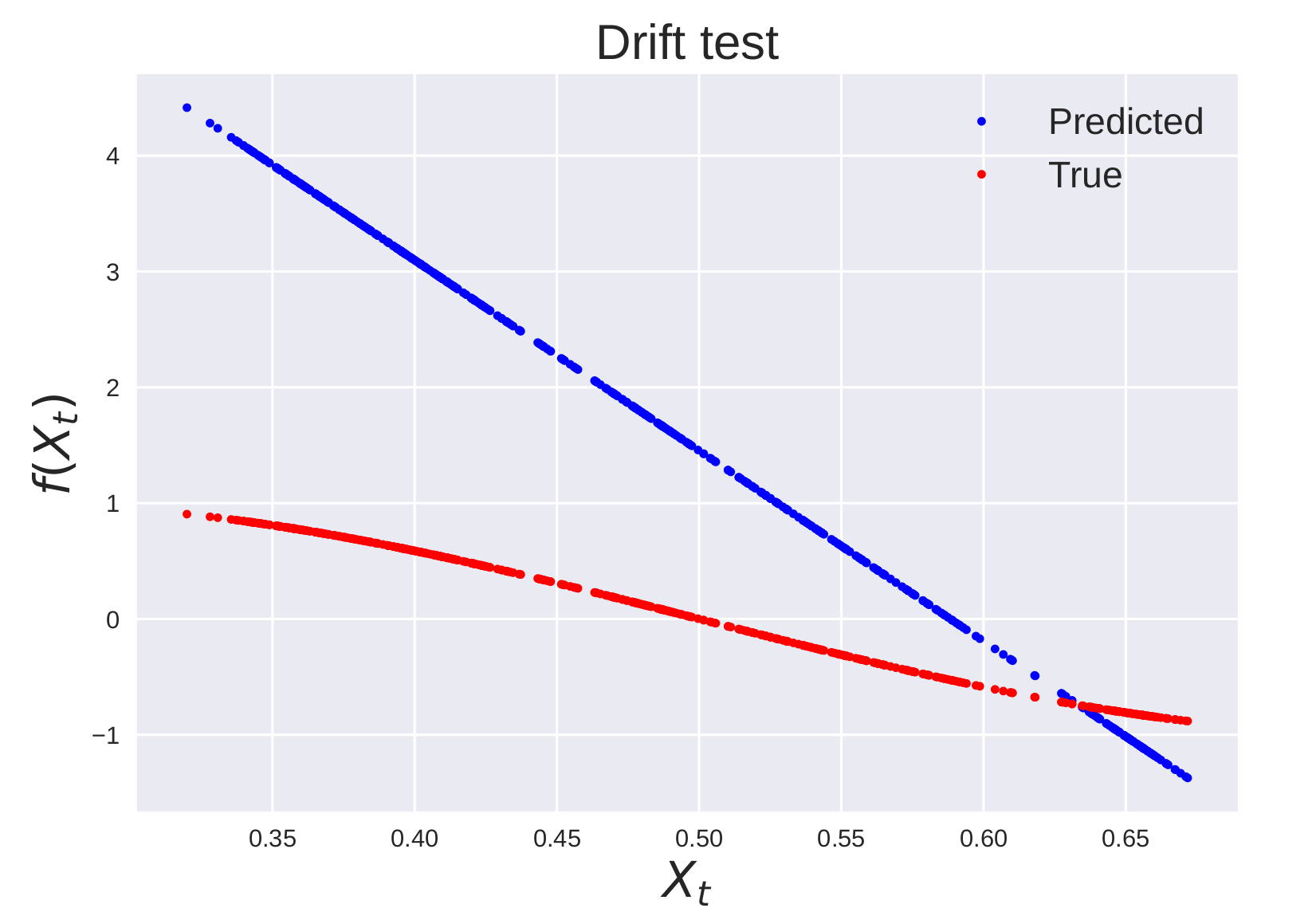}
    \includegraphics[width = 0.23\linewidth, height = 3cm]{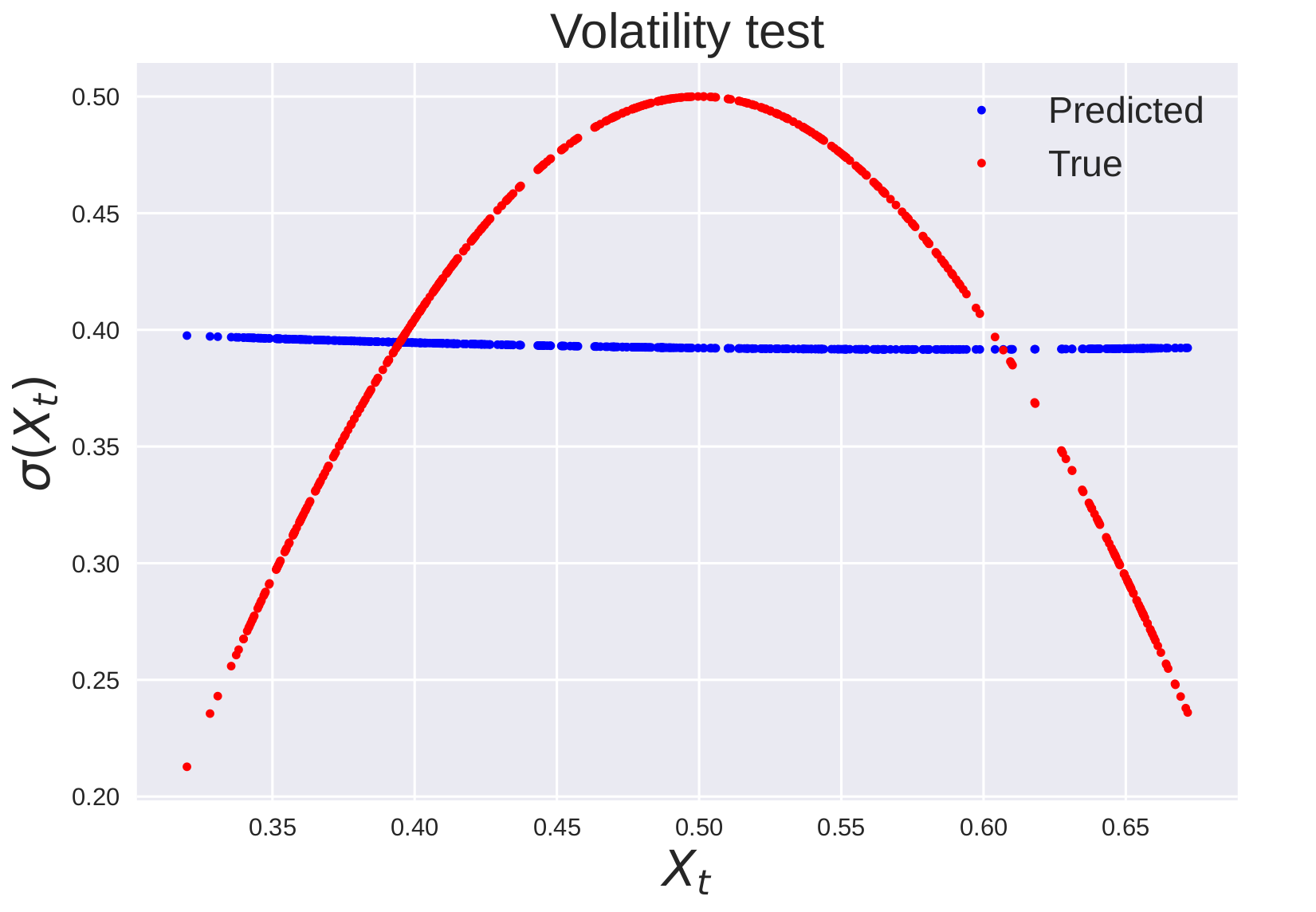}
    \caption{Trigonometric process benchmark prediction. From left to right: drift (trajectory 1), volatility (trajectory 1), drift (trajectory 2), volatility (trajectory 2).}
    \label{fig: benchmark, trig}
\end{figure}
\section{Experimental results: a comparison between perfectly specified and non perfectly specified kernels.} \label{sec: specified kernels}

We now illustrate how the choice of the kernel can have a significant impact on the recovery of the drift and volatility. The following examples illustrate how choosing from the correct parametric family of kernels can improve both the recovery and the prediction of the functions of interest.

\subsubsection{Overcoming non perfectly specified kernels}
We consider the Geometric Brownian Motion process defined as 
\begin{equation}
    dX_t = \mu X_t dt + \sigma X_tdW_t \quad \text{Geometric Brownian motion (GBM).}
\end{equation}
We generate one trajectory with parameters $\mu = 2.0, \sigma = 1.0$ and initial condition $X_0 = 1.0$, using the Euler-Maruyama discretization with timestep $\Delta t = 0.001$. We focus primarily on the recovery of the volatility (the recovery of the drift is very difficult at a fine time-scale). The generated data is illustrated in figure \ref{fig: data, GBM}.

\begin{figure}[h]
    \centering
    \includegraphics[width = 0.23\linewidth, height = 3cm]{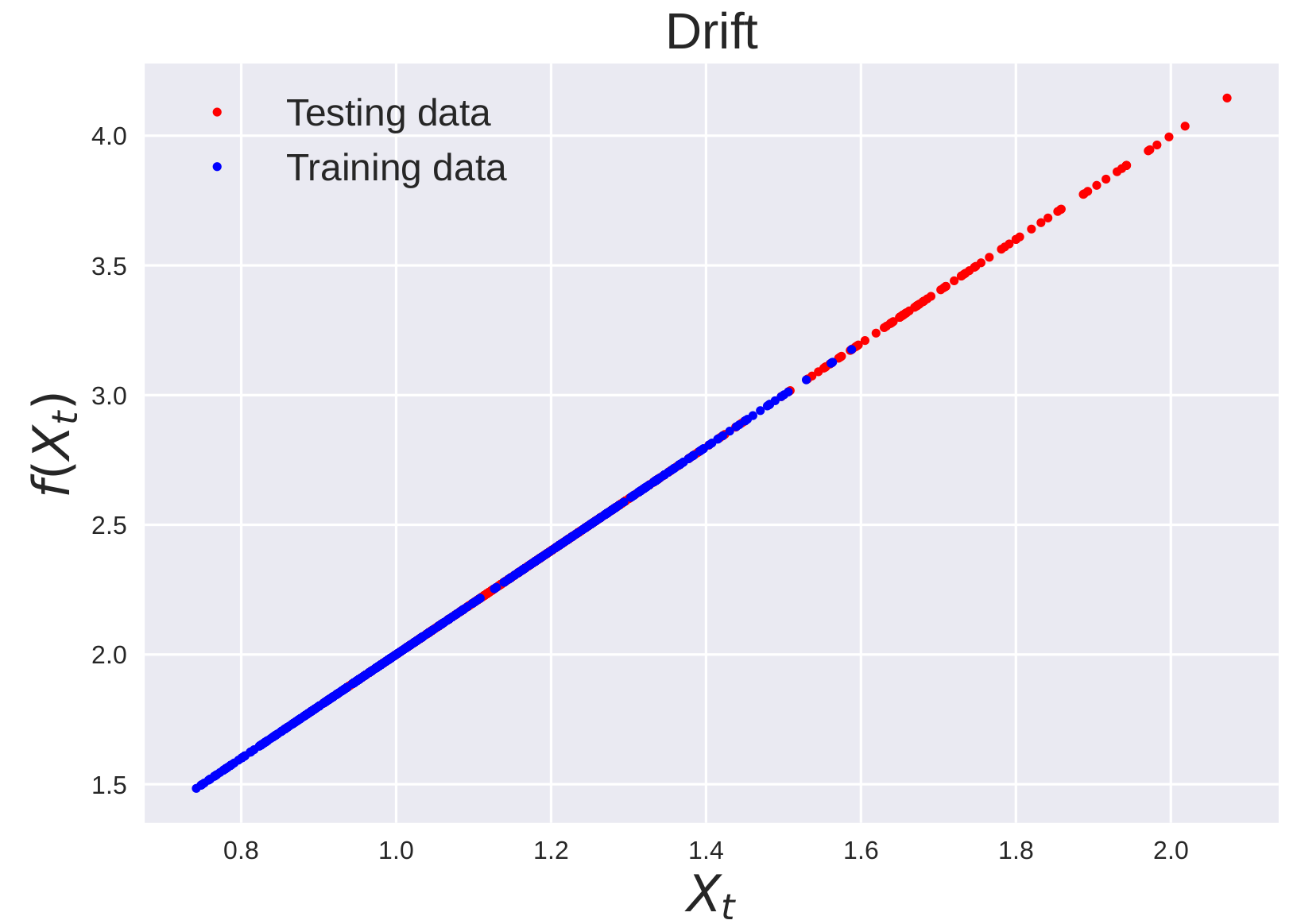}
    \includegraphics[width = 0.23\linewidth, height = 3cm]{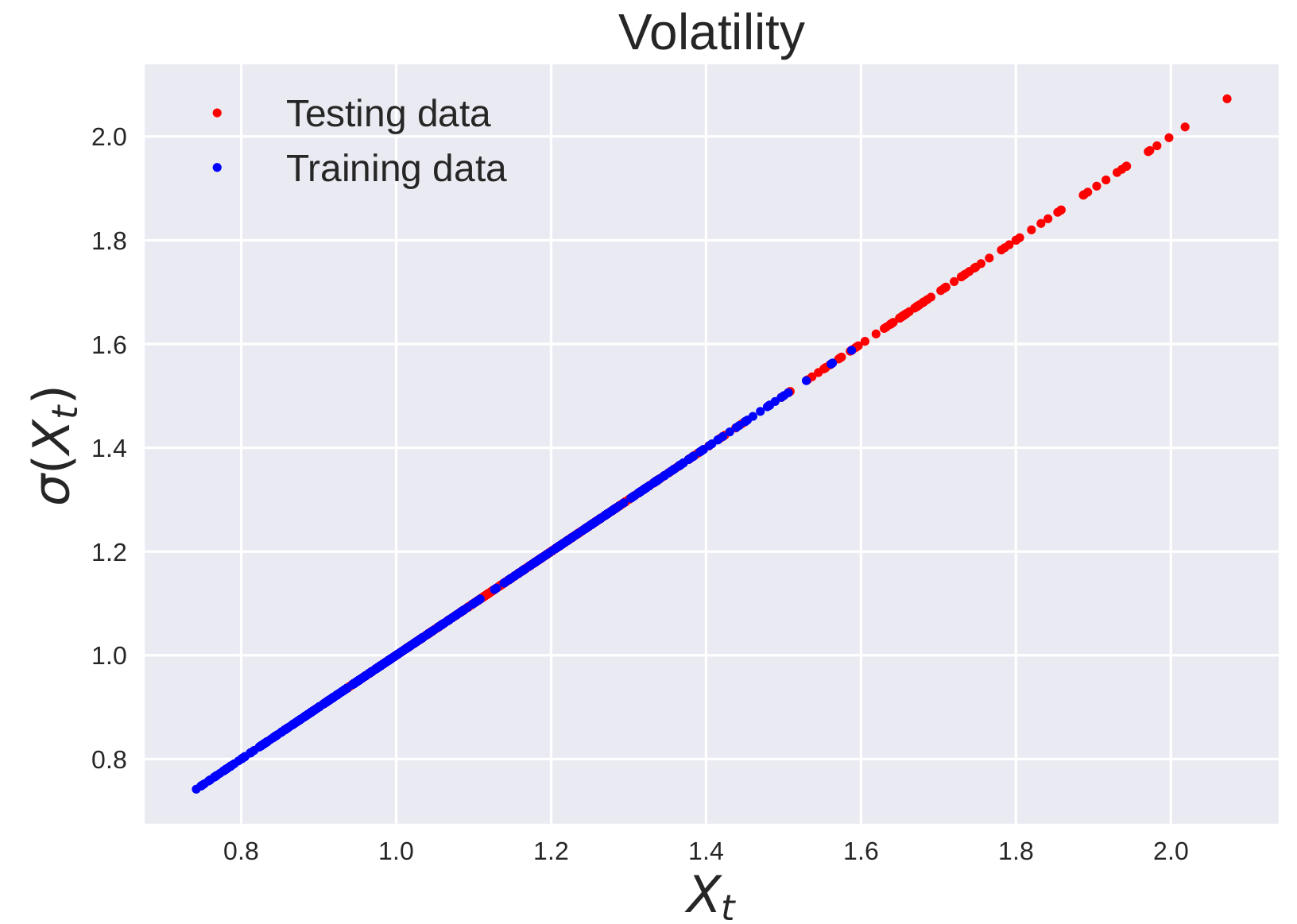}
    \includegraphics[width = 0.23\linewidth, height = 3cm]{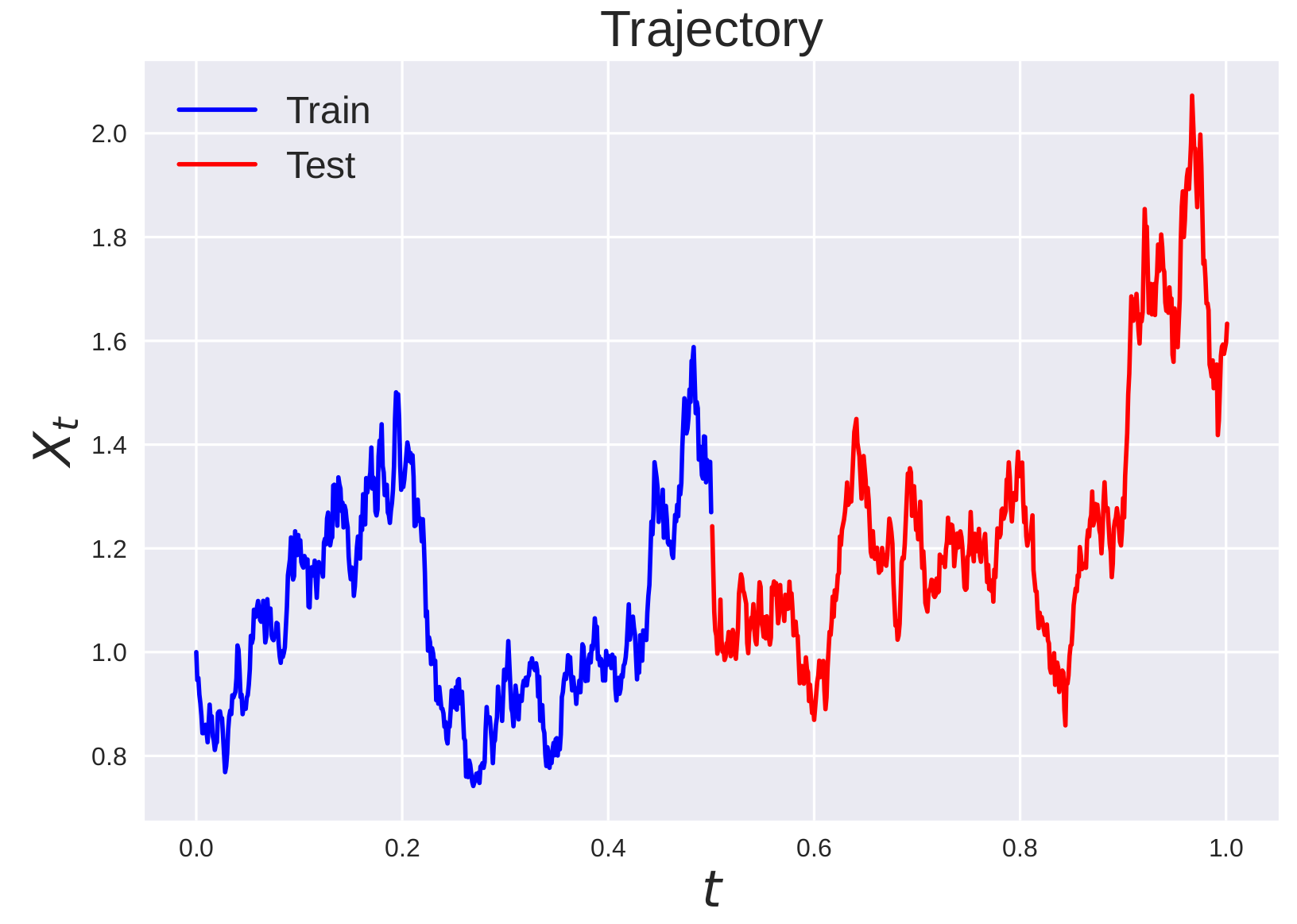}
    \includegraphics[width = 0.23\linewidth, height = 3cm]{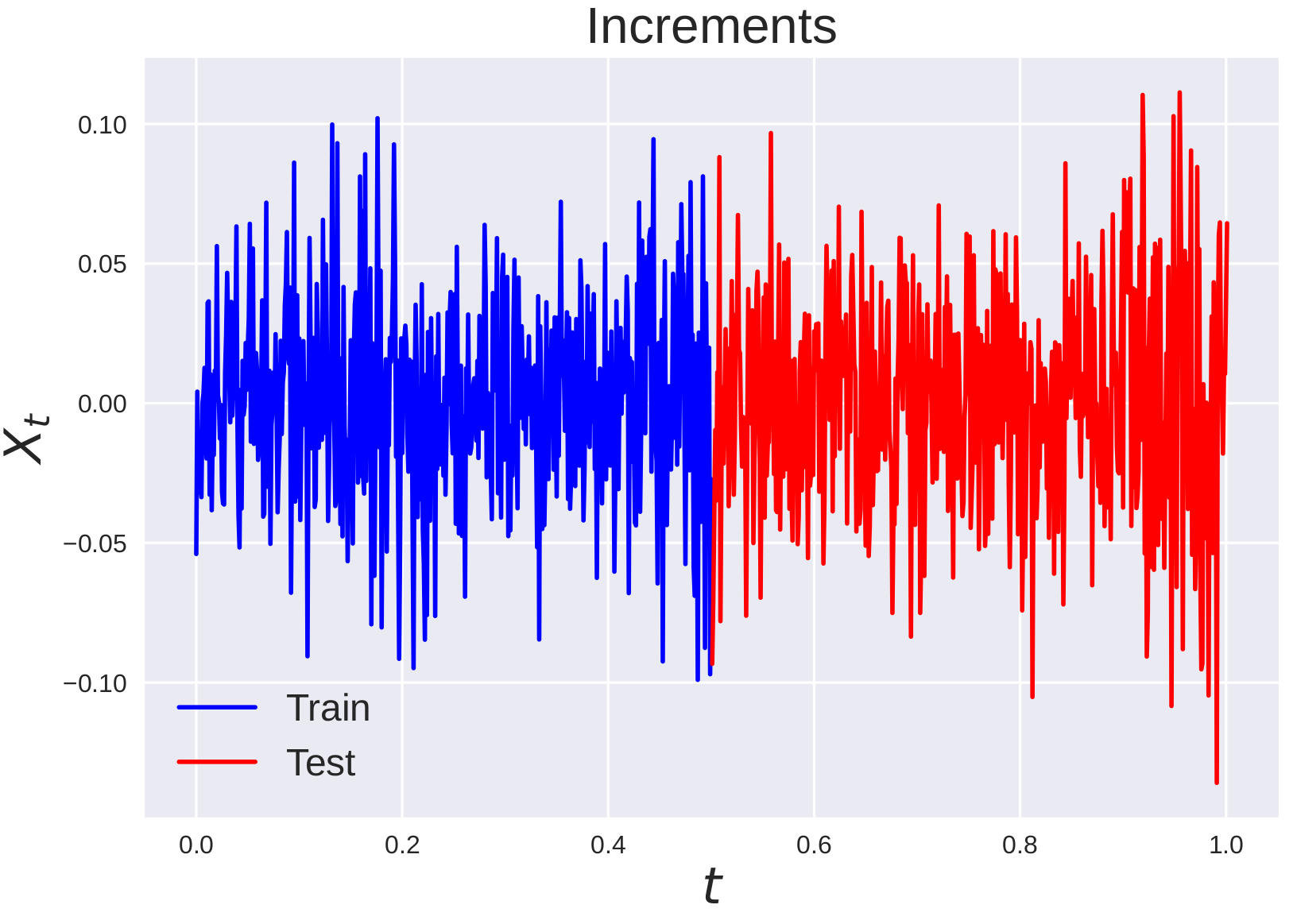}
    \caption{From left to right: drift function, volatility function, sample trajectory, and sample increments of GBM.}
    \label{fig: data, GBM}
\end{figure}

We compare the performance of the Matern kernel \eqref{eq: matern kernel} family with the family of linear kernels \cite{GPforML} defined as 
\begin{equation}\label{eq: linear kernel}
    K_{\text{linear}}(x,y) = \sigma^2(x^\intercal y +c)
\end{equation}
and parameterized by $(\sigma, c)$ which are learned. Note that this kernel induces the Reproducing Kernel Hilbert Space of linear functions and therefore is perfectly specified for GBM. The results are reported in table \ref{table: GBM}. The linear kernel is able to both recover and predict the volatility with or without learning the hyper-parameters, as it is perfectly specified for this problem.  In contrast, without learning the hyper-parameters, the Mat\'ern kernel is unable to accurately predict the future, reverting to the prior mean 0. Learning the hyper-parameters, however, enables the Mat\'ern kernel to correctly predict future values. These results are illustrated in figure \ref{fig: pred, GBM}. We also observe that in all cases, our proposed approach outperforms the Gaussian Process Regressor benchmark (see figure \ref{fig: benchmark, GBM}). 

\begin{table}[h]
\begin{center}
\begin{tabular}{|l|c|c|c|c|c|c|} 
 \hline
 & \multicolumn{3}{|c|}{Linear kernel} & \multicolumn{3}{|c|}{Matern kernel} 
  \\ 
 \hline
  &$\mathcal{L}(\bar{f}^*, \bar{\sigma}^* |X, Y)$ & $\delta_f$ & $\delta_\sigma$  &$\mathcal{L}(\bar{f}^*, \bar{\sigma}^* |X, Y)$ &$\delta_f$& $\delta_\sigma$\\ 
     \hline
Benchmark  & -2.755  & 2.077& 0.236& -2.755  &2.018& 0.237 \\
   \hline
Non-learned kernel   & \textbf{-2.800}  & 0.763 & 0.015 & -1.420  &0.962& 0.355 \\
 \hline
Learned kernel   &  \textbf{-2.800}   & \textbf{0.672} & \textbf{0.008}& \textbf{-2.800} & \textbf{0.500}& \textbf{0.010} \\
\hline
\end{tabular}
\caption{Comparison between the linear kernel and the Matern kernel on GBM.}
\label{table: GBM}
\end{center}
\end{table}

\begin{figure}[h]
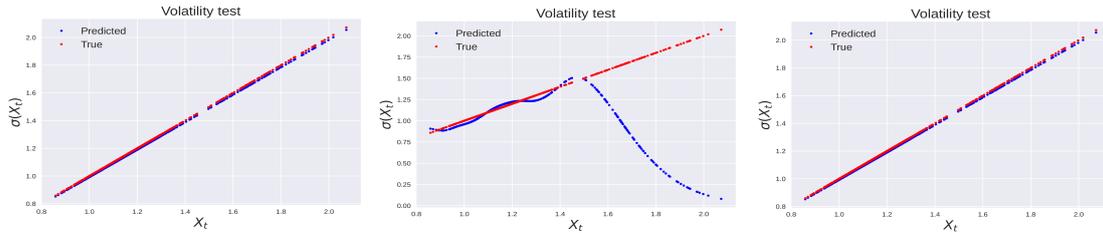

    \centering
        \includegraphics[width = 0.3\linewidth, height = 3cm]{images/GBM/New optimization/Linear/Pred1.png}
    \includegraphics[width = 0.3\linewidth, height = 3cm]{images/GBM/New optimization/Matern/Pred1.png}
    \includegraphics[width = 0.3\linewidth, height = 3cm]{images/GBM/New optimization/Matern/Pred3.png}
    \caption{Predicted volatility on the testing set for GBM. From left to right: linear kernel, Matern kernel (non-learned parameters) and Matern Kernel (learned parameters).}
    \label{fig: pred, GBM}
\end{figure}

\begin{figure}[h]
    \centering
    \includegraphics[width = 0.3\linewidth, height = 3cm]{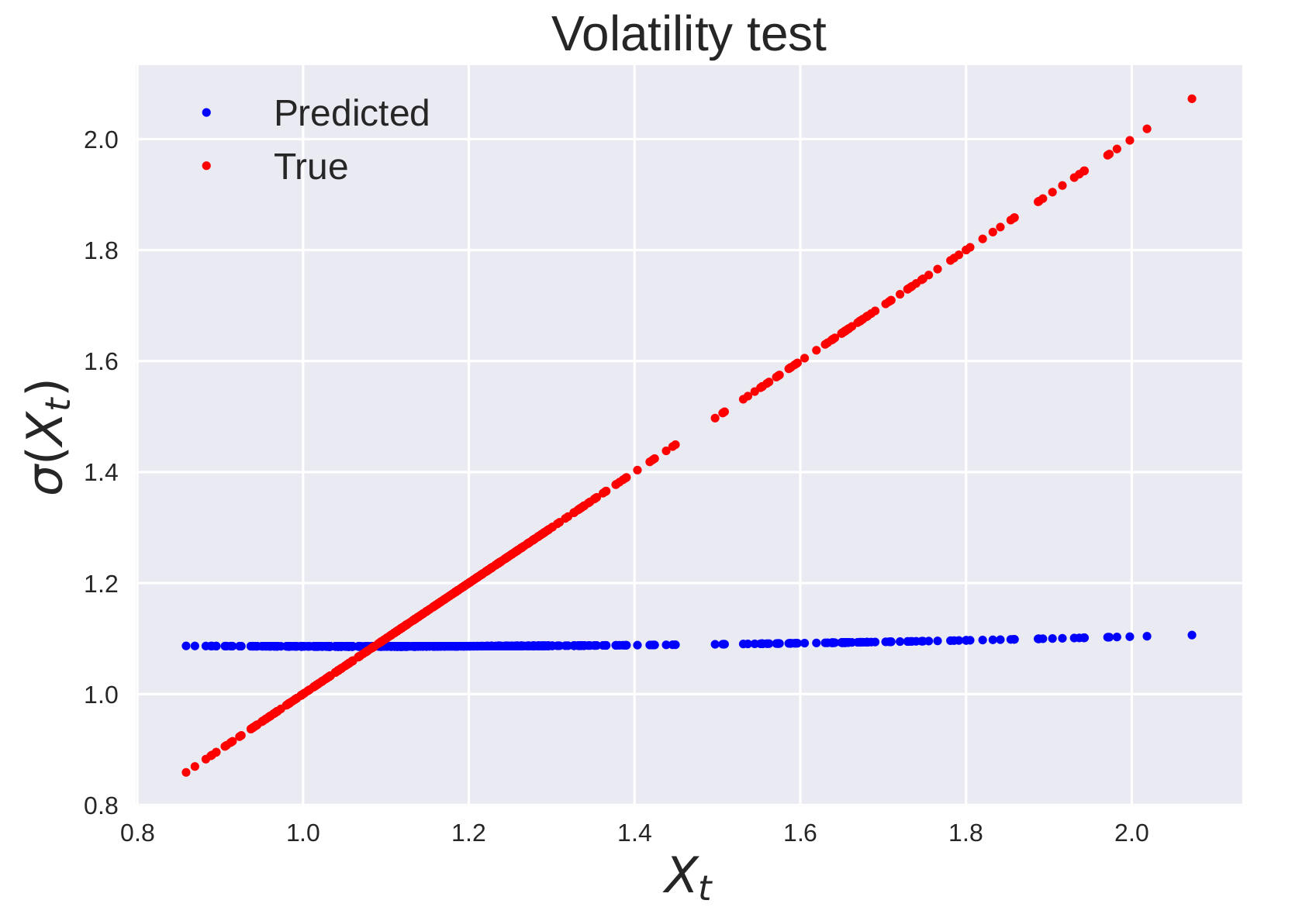}
    \includegraphics[width = 0.3\linewidth, height = 3cm]{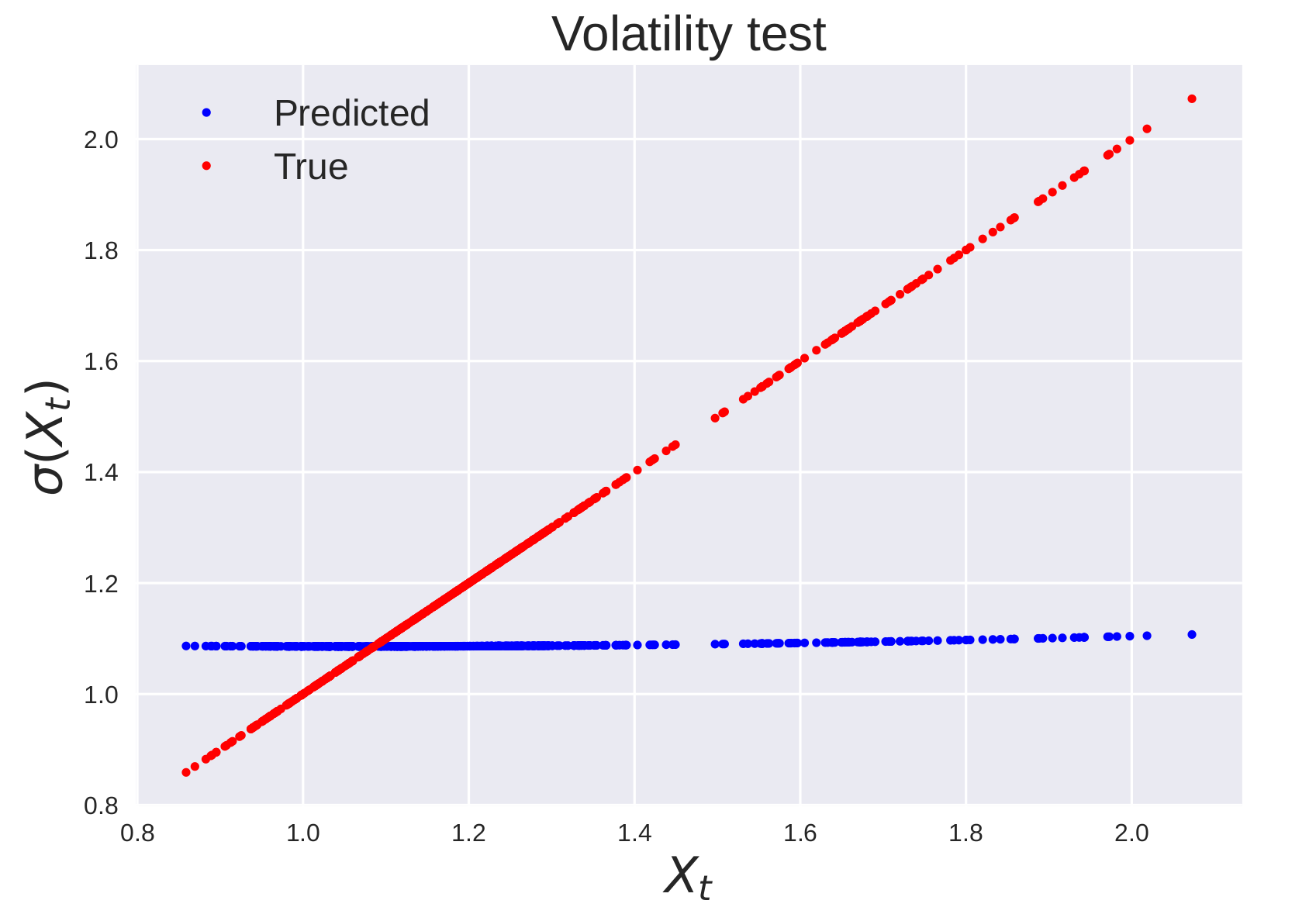}
    \caption{GBM benchmark prediction. From left to right: volatility (linear kernel) and volatility (Matern kernel).}
    \label{fig: benchmark, GBM}
\end{figure}

\subsubsection{A failure case}\label{sec: failure}

We now present a case where our methodology fails because of a poor choice of prior. We consider the Ornstein–Uhlenbeck process defined as 
\begin{equation*}
    dX_t = -\mu X_t dt + \sigma dWt, \quad \text{Ornstein–Uhlenbeck (OU).}
\end{equation*}
We discretize the process with parameters $\mu = -5$ and $\sigma = 1.0$ and initial condition $X_0 = 1.0$ using the Euler-Maruyama method with a time discretization of $\Delta t = 0.001$. One of the trajectories is illustrated in figure \ref{fig: data, OU}.

\begin{figure}[h]
    \centering
    \includegraphics[width = 0.23\linewidth, height = 3cm]{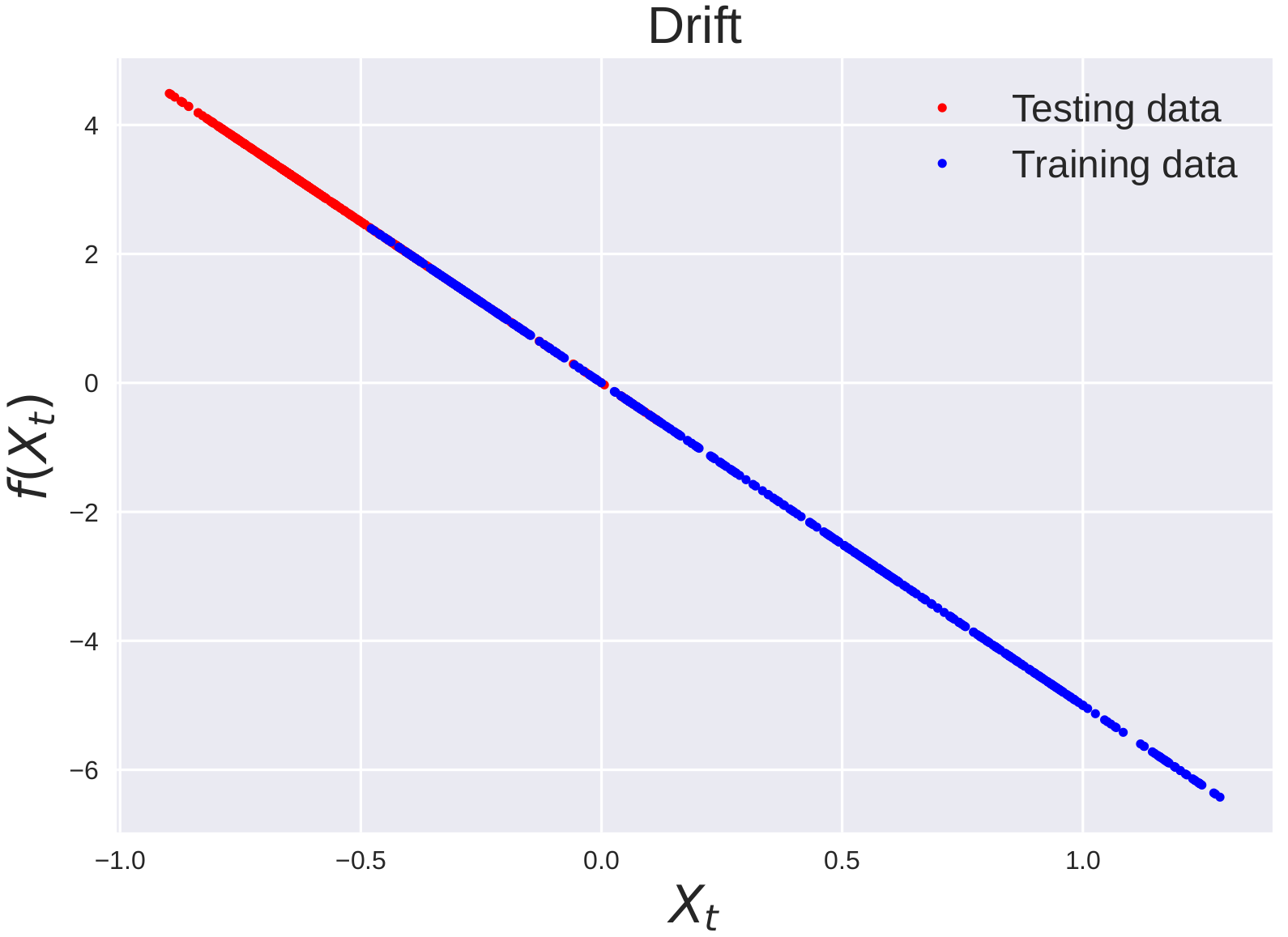}
    \includegraphics[width = 0.23\linewidth, height = 3cm]{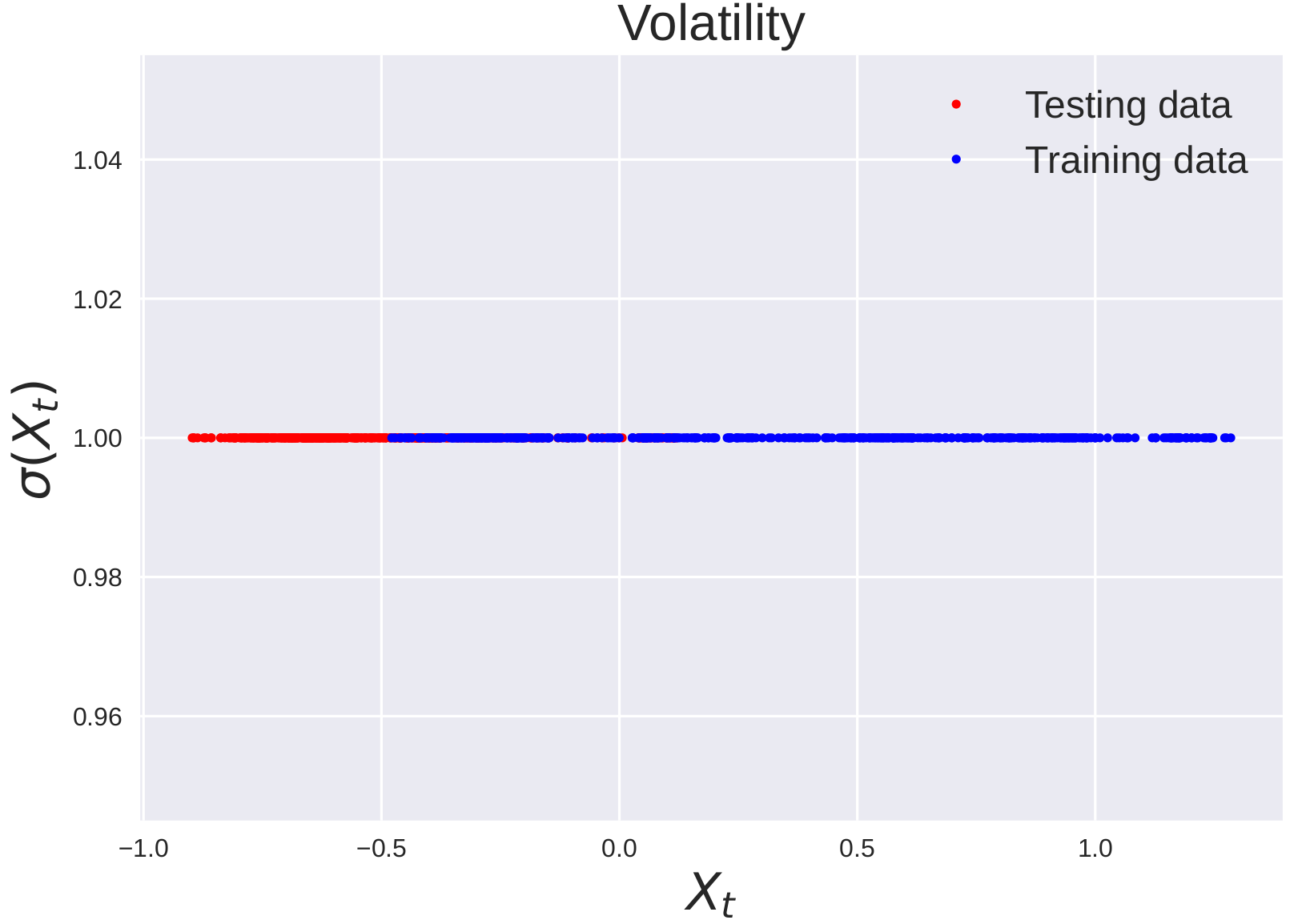}
    \includegraphics[width = 0.23\linewidth, height = 3cm]{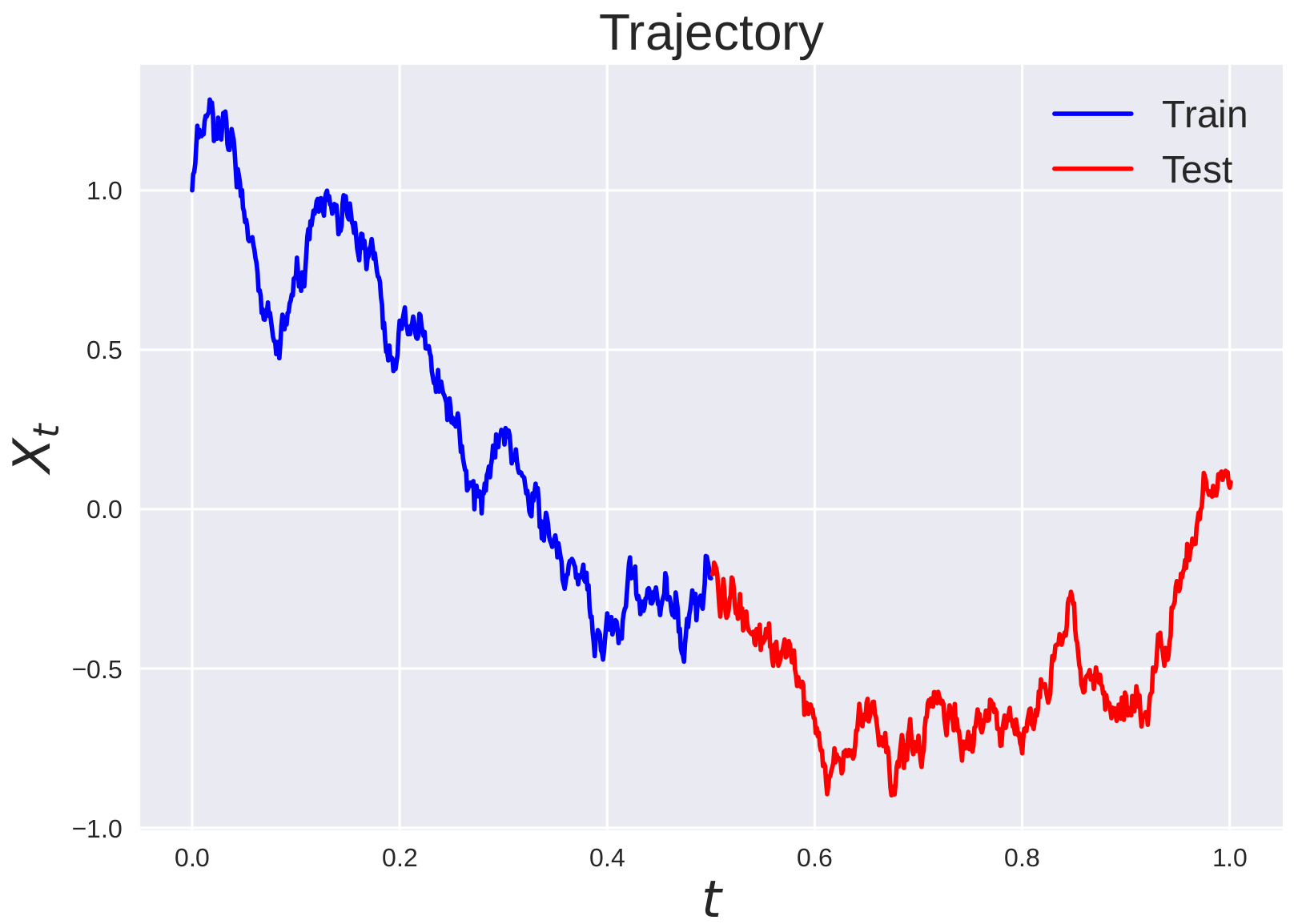}
    \includegraphics[width = 0.23\linewidth, height = 3cm]{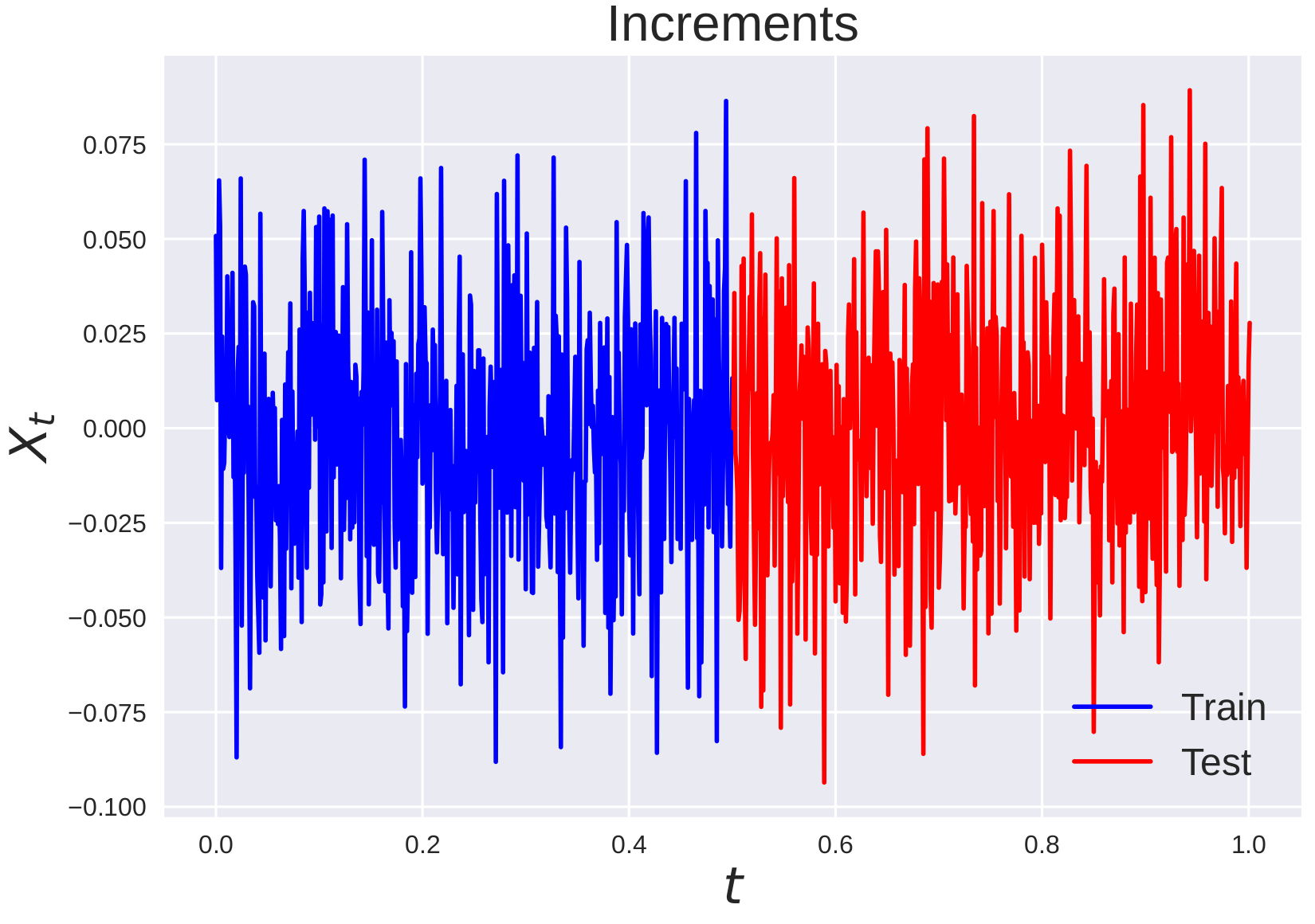}
    \caption{From left to right: drift function, volatility function, sample trajectory, and sample increments of the OU process.}
    \label{fig: data, OU}
\end{figure}

In this case, the drift is a linear function, and the volatility is a constant function. Hence a Matérn kernel prior is not perfectly specified. More precisely, the best prior for the volatility is a white noise kernel 
\begin{equation}
    K_{\text{wn}}(x,y) = \sigma^2 \delta(x-y).
\end{equation}
 Therefore, in this case, the benchmark method (Gaussian process regression) is not only better specified than our method, but it is also optimal. We again use both the Matérn \ref{eq: matern kernel} and linear kernels \eqref{eq: linear kernel}. The results are presented in table \ref{table: OU results}. In this case, the benchmark always outperforms our computational graph completion approach in the recovery of the volatility. With a Matérn kernel our method outperforms the benchmark, but with a perfectly specified linear kernel, the benchmark outperforms our method. Moreover, figure \ref{fig: pred, OU} shows that the Matérn kernel does not capture the overall shape of the drift and volatility even if the relative error is low. We do note, however our cross-validation method improves the performance.

\begin{table}[H]
\begin{center}
\begin{tabular}{|l|c|c|c|c|c|c|} 
 \hline
 & \multicolumn{3}{|c|}{Matérn kernel} & \multicolumn{3}{|c|}{Linear kernel} 
  \\ 
 \hline
  &$\mathcal{L}(\bar{f}^*, \bar{\sigma}^* |X, Y)$ & $\delta_f$ & $\delta_\sigma$  &$\mathcal{L}(\bar{f}^*, \bar{\sigma}^* |X, Y)$ &$\delta_f$& $\delta_\sigma$\\
  \hline
  Benchmark  & \textbf{ -2.973} & 1.799 & \textbf{0.006} & \textbf{-2.977} & \textbf{0.212}& \textbf{0.006}\\
   \hline
Non-learned kernel   & -2.971 &  1.038 & 0.065 &-2.957  & 0.510& 0.011\\
 \hline
Learned kernel   &  \textbf{ -2.973}  & \textbf{0.459} & 0.012& -2.978 & 1.066& 0.013 \\
\hline
\end{tabular}
\caption{Results for the OU process.}
\label{table: OU results}
\end{center}
\end{table}

\begin{figure}[H]
\centering
    \includegraphics[width = 0.22\linewidth, height = 3cm]{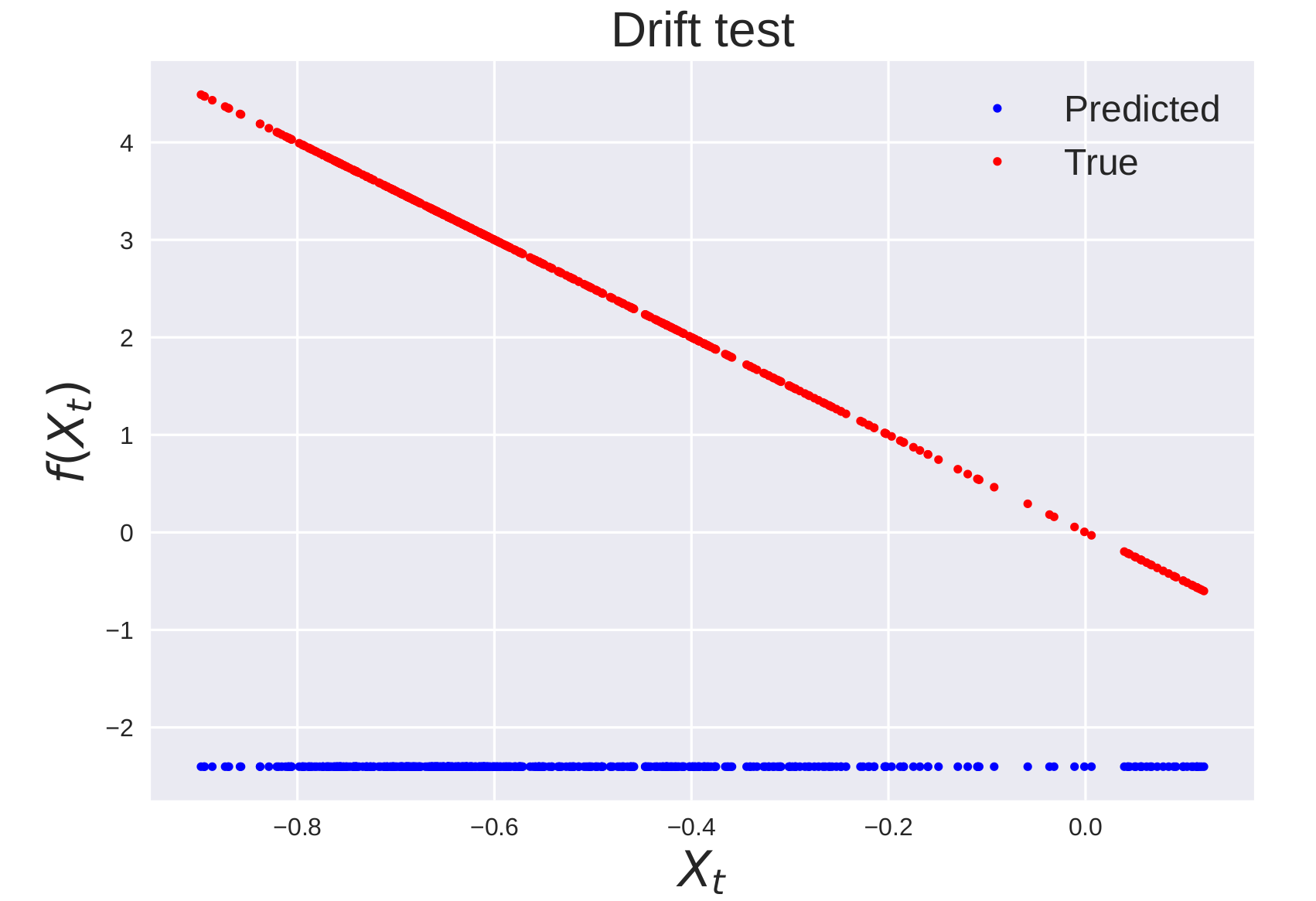}
    \includegraphics[width = 0.22\linewidth, height = 3cm]{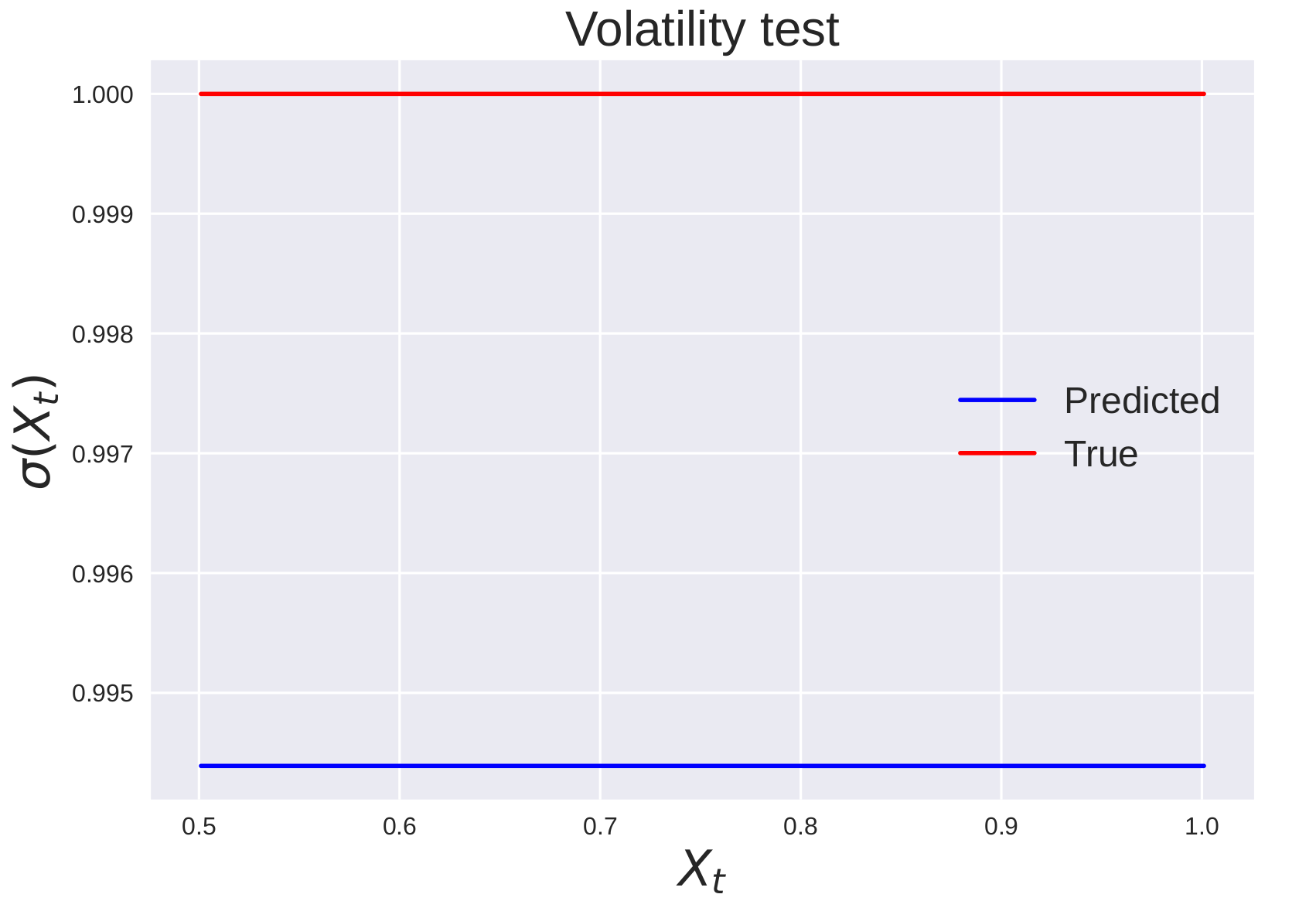}
    \includegraphics[width = 0.22\linewidth, height = 3cm]{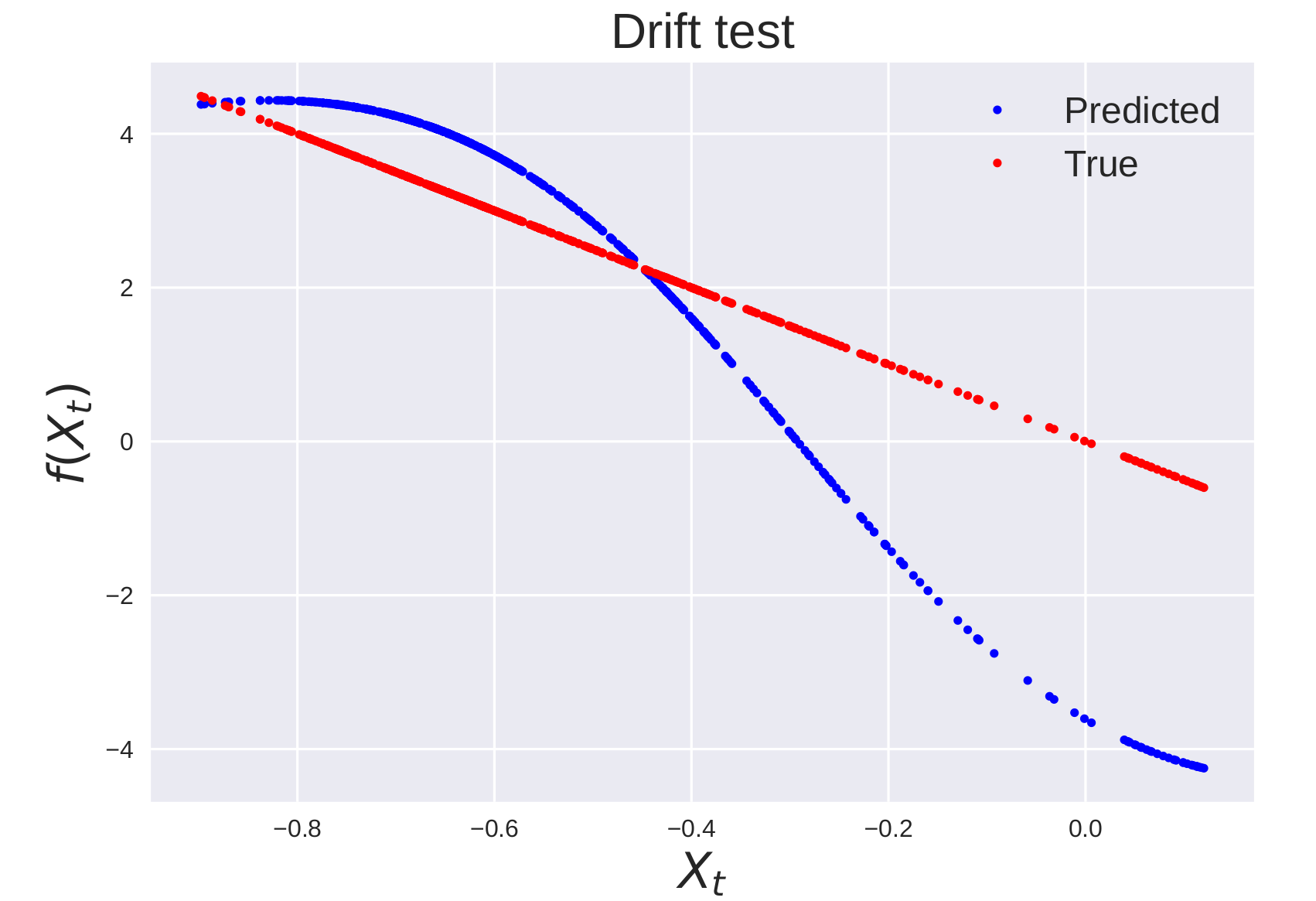}
    \includegraphics[width = 0.22\linewidth, height = 3cm]{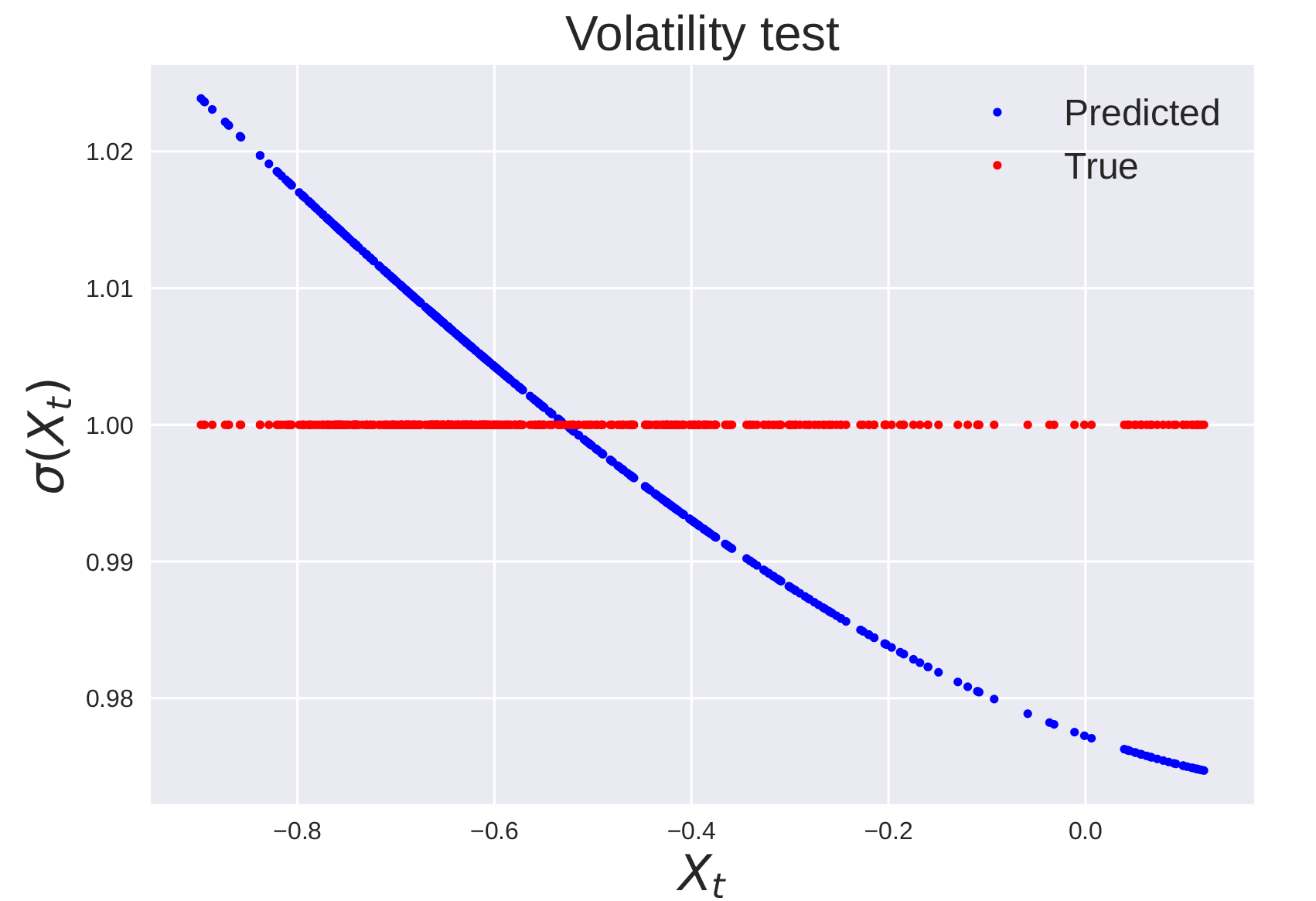}
    \caption{Prediction on the OU process with the Matern kernel. From left to right:  benchmark prediction (drift), benchmark prediction (volatility), CGC prediction (drift), CGC prediction (volatility).}
    \label{fig: pred, OU}
\end{figure}

\begin{figure}[H]
\centering
    \includegraphics[width = 0.22\linewidth, height = 3cm]{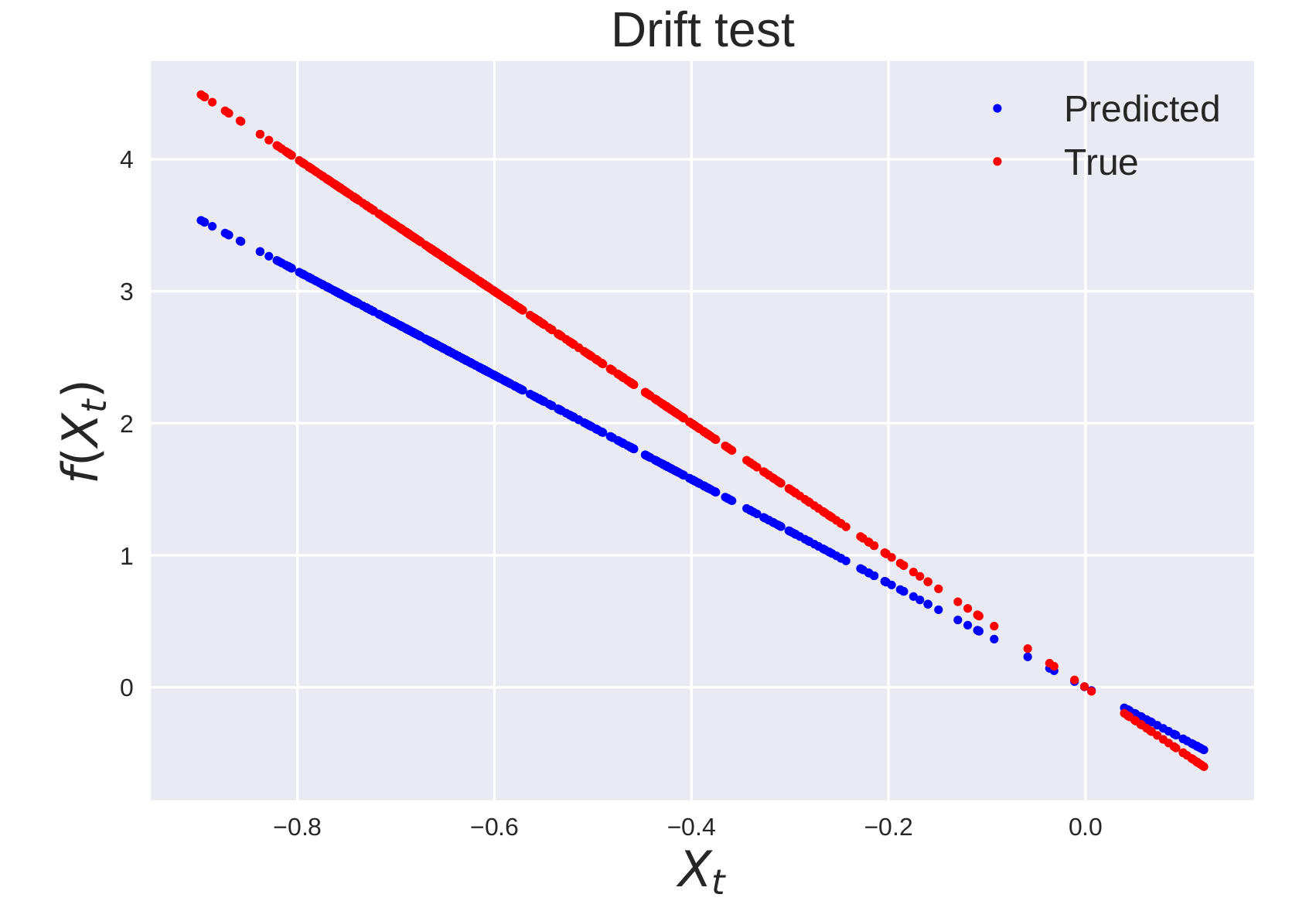}
    \includegraphics[width = 0.22\linewidth, height = 3cm]{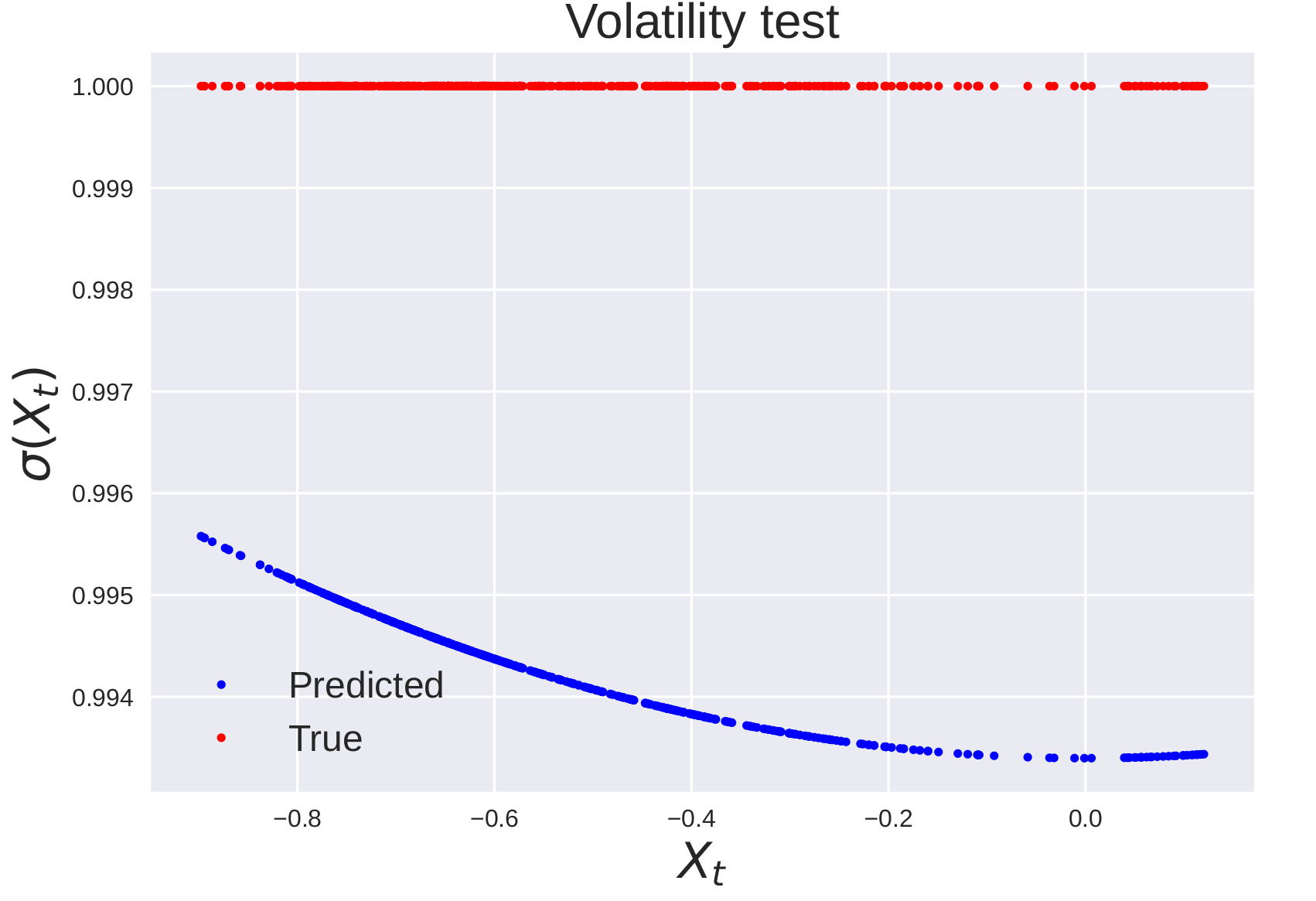}
    \includegraphics[width = 0.22\linewidth, height = 3cm]{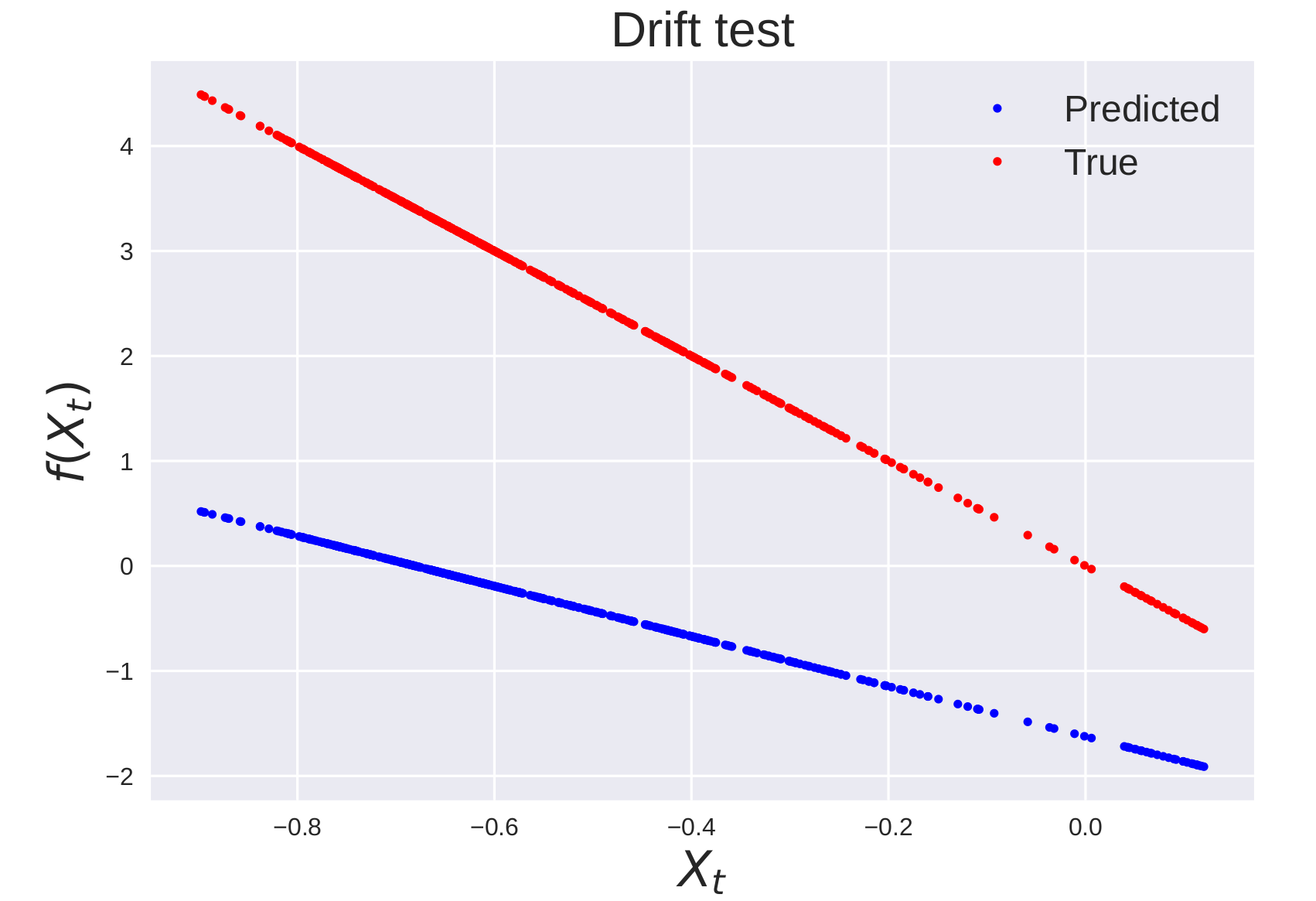}
    \includegraphics[width = 0.22\linewidth, height = 3cm]{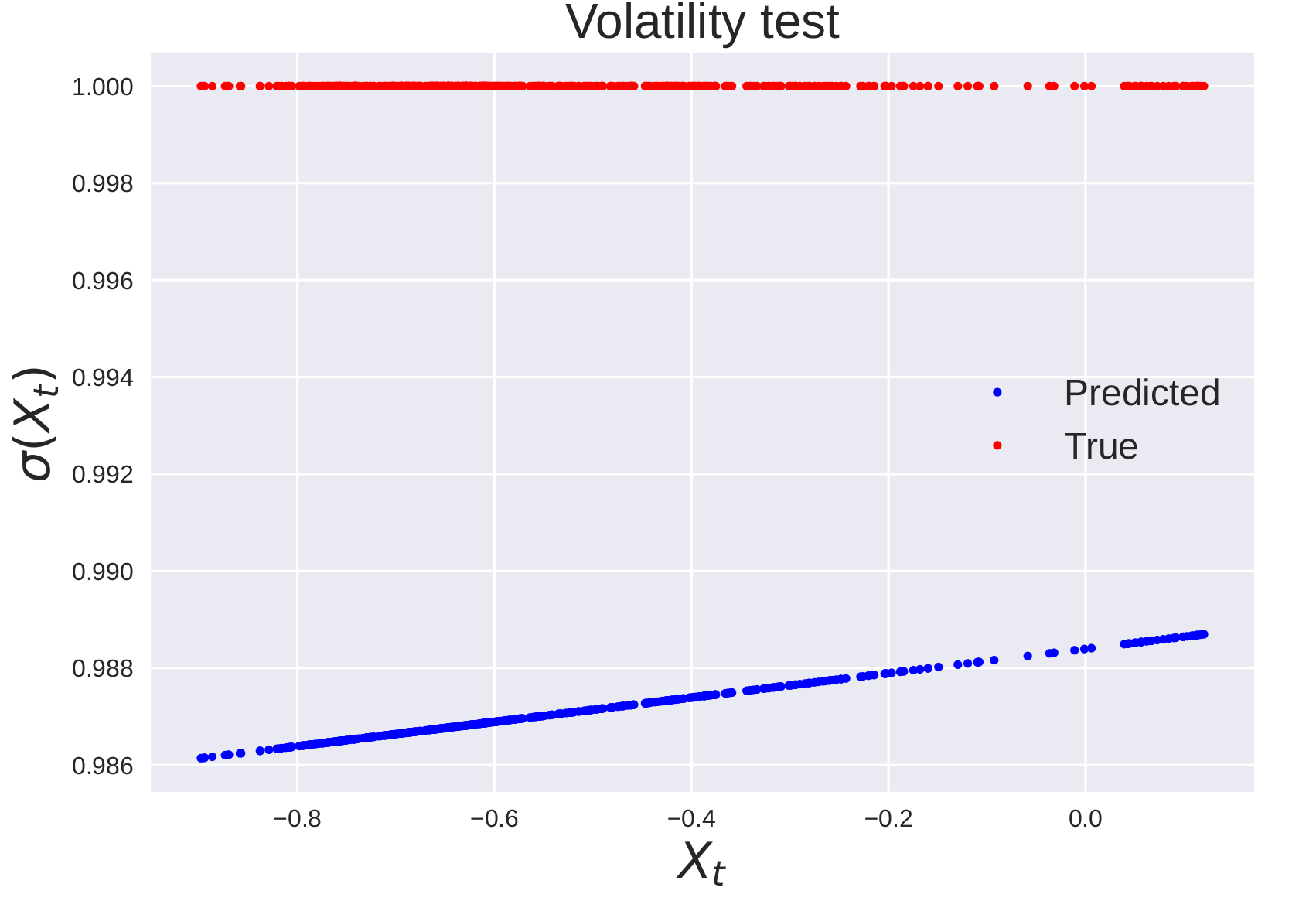}
    \caption{Prediction on the OU process with the linear kernel. From left to right:  benchmark prediction (drift), benchmark prediction (volatility), CGC prediction (drift), CGC prediction (volatility).}
    \label{fig: pred, OU linear}
\end{figure}

These results illustrate that choosing a perfectly specified kernel can improve the performance of the proposed method. Therefore, integrating prior knowledge such as the non-stationarity of the process into the choice of the prior can significantly improve performance. However, these results also illustrate how learning the parameters of a general kernel (such as the Mat\'{e}rn kernel) can yield similar performance to a well-specified kernel. A potential solution to this problem could be to learn a kernel which is the sum of a mix of stationary and non-stationary kernels such as 
\begin{equation}
    K_{\text{sum}}(x,y) = \sum_{i=1}^N \alpha_i K_i (x,y)
\end{equation}
where the weights $\alpha_i$ are learned. Learning the weights of a sum of kernels is a well-established problem in the area of Support Vector Machines (see, for example \cite{MKL}). Future avenues of research could therefore focus on efficient learning of hyper-parameters for this problem.  

\section{Experimental results: the effect of time discretization}\label{sec: time disretization}

We now examine the effect of time discretization. In all previous experiments, we observed the exact simulation of the trajectory. We now consider the cases where the observations do not exactly match the dynamics. More precisely, we generate a trajectory $(X_i)_{i=1}^M$ separated by regular time steps $\Delta t$. We then consider the subs-trajectories $(X_{ik})_{ik = 1}^N$ for $k = 1, 2, \dots, 10$ where the indexes $ik$ are chosen such that $(X_{ik})_{ik = 1}^N$ are separated by timesteps $k\times \Delta t$. This allows us to measure the impact of the discretization error on the effectiveness of our method. A priori, we would expect that our method would perform more poorly as $k$ grows larger, given that our modeling assumption is no longer fulfilled exactly.

We consider both our method with the initial guess of parameters (as detailed in the benchmark section \ref{sec: benchmarks details}) and with the optimized version of parameters. We also choose a base $\lambda$ which performs well for $k = 1$ and set $\lambda_k = k\lambda$ for other values of $k$.
\subsection{Exponential decay volatility.} 
\begin{figure}[H]
    \centering
    \includegraphics[width = 0.3\linewidth, height = 3cm]{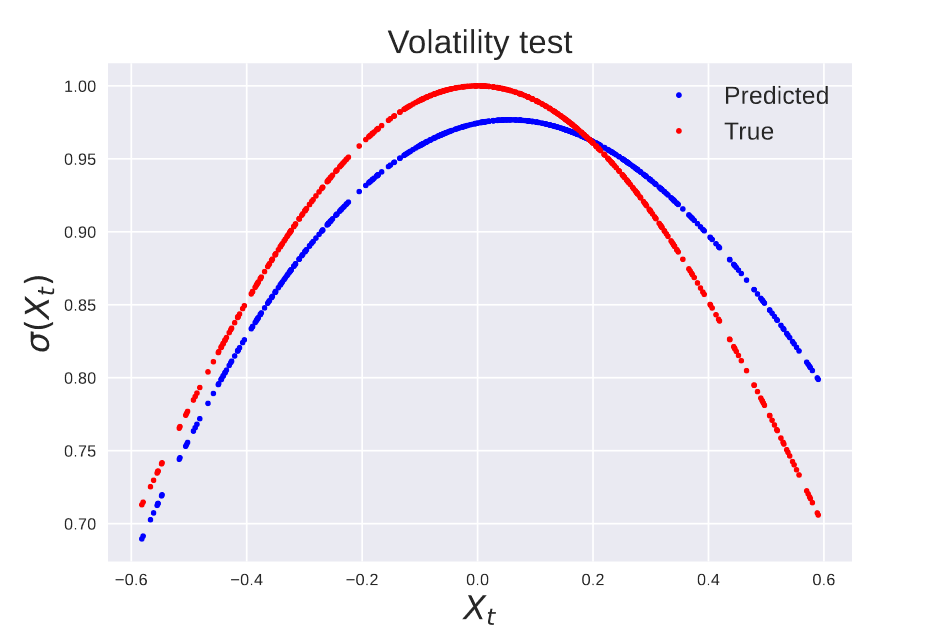}
    \includegraphics[width = 0.3\linewidth, height = 3cm]{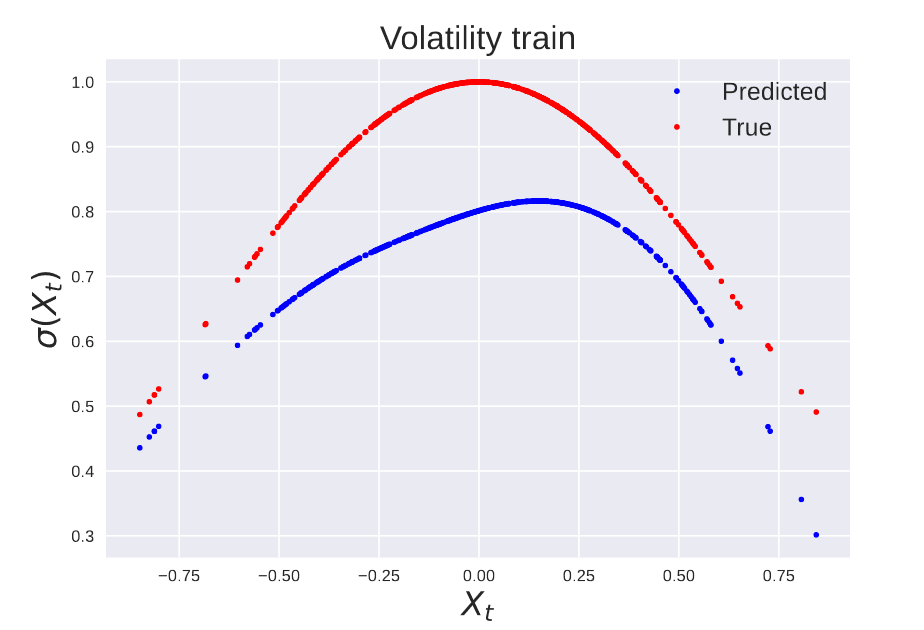}
    \includegraphics[width = 0.3\linewidth, height = 3cm]{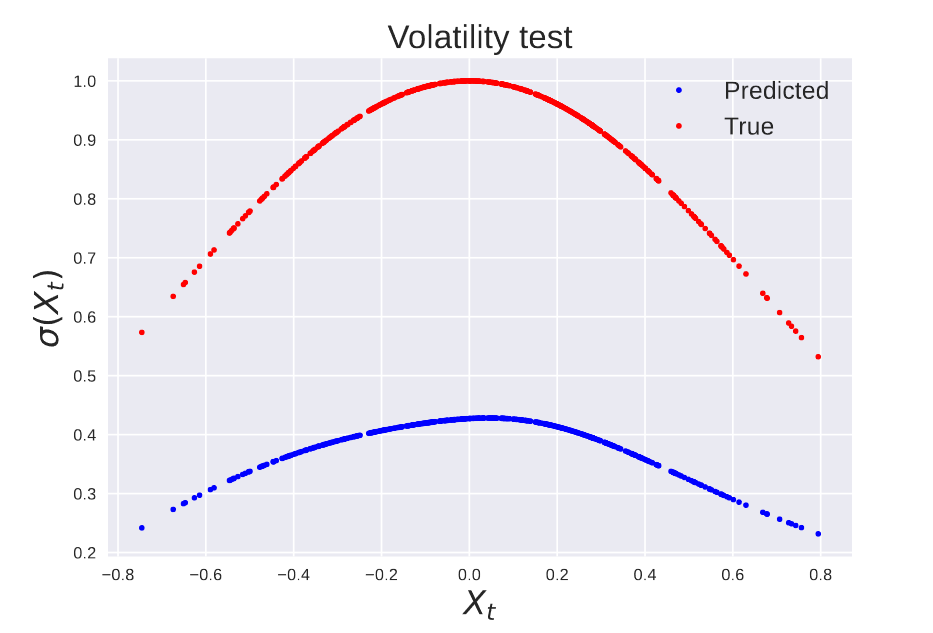}
    \caption{Exponential decay volatility process: prediction of the volatility for different values of $k$. From left to right: $k=1,2,5$.}
    \label{fig: exp multiscale vol}
\end{figure}

\begin{figure}[H]
    \centering
    \includegraphics[width = 0.3\linewidth, height = 3cm]{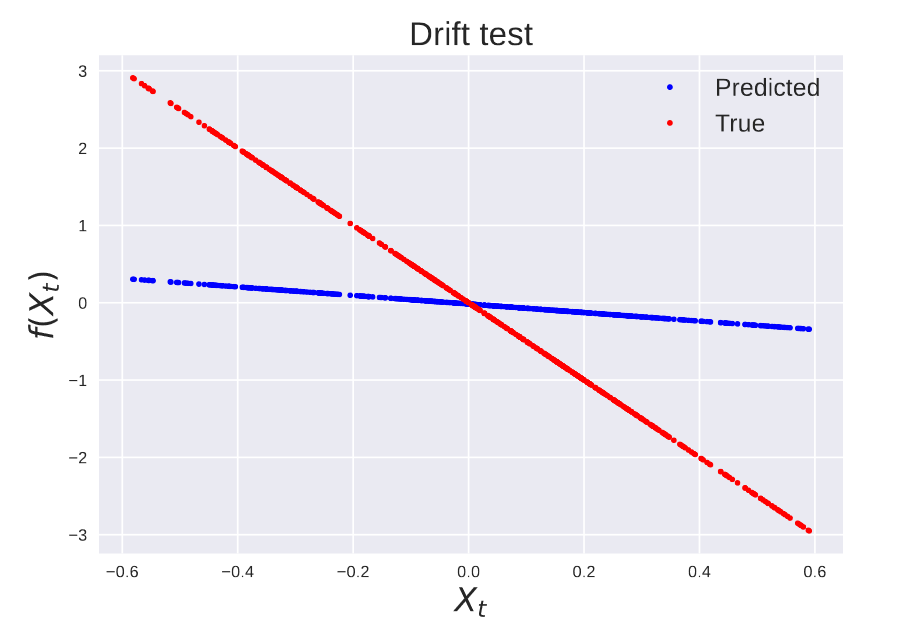}
    \includegraphics[width = 0.3\linewidth, height = 3cm]{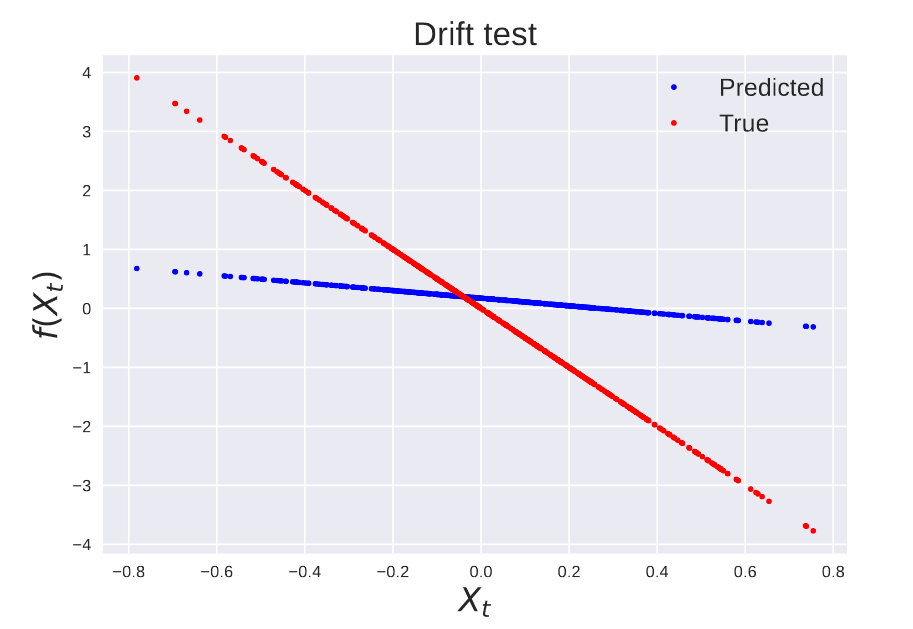}
    \includegraphics[width = 0.3\linewidth, height = 3cm]{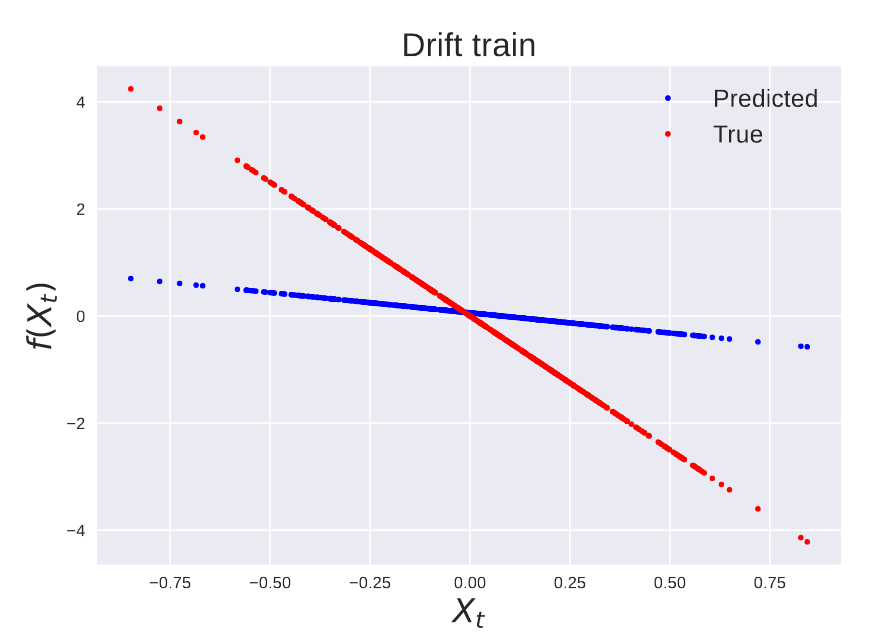}
    \caption{Exponential decay volatility process: prediction of the drift for different values of $k$. From left to right: $k=1,2,5$.}
    \label{fig: exp multiscale drift}
\end{figure}

We first consider the exponential decay volatility process described by \eqref{eq: exp decay discrete} with time discretization $\Delta t = 0.01$. For each $k$, we set $N = 500$ for both the training and test sets. The results for $k$ are presented in table \ref{table: exp multiscale} and illustrative examples are presented in figures \ref{fig: exp multiscale vol} and \ref{fig: exp multiscale drift}. We note that the choice of the Mat\'ern kernel for the drift function induces a large error. A better choice of the kernel (such as the linear kernel discussed in section \ref{sec: specified kernels}) would yield a better performance.

\begin{table}[H]
\begin{center}
\begin{tabular}{|l|c|c|c|c|c|c|} 
 \hline
 & \multicolumn{3}{|c|}{Non optimized parameters} & \multicolumn{3}{|c|}{Optimized parameters} 
  \\ 
 \hline
  &$\mathcal{L}(\bar{f}^*, \bar{\sigma}^* |X, Y)$ & $\delta_f$ & $\delta_\sigma$  &$\mathcal{L}(\bar{f}^*, \bar{\sigma}^* |X, Y)$ &$\delta_f$& $\delta_\sigma$\\
  \hline
  k = 1  & -1.887 &  0.472 & 0.062 & -1.887 & 0.889& 0.033\\
\hline
  k = 2  & -1.872 &  0.669 & 0.299 &-1.843 &0.897& 0.181\\
\hline
  k = 3  &-1.896 &  0.702 &0.419 & -1.888 &0.881&  0.415\\
\hline
  k = 4  & -1.871 &  0.886 &0.505 & -1.922 &0.902&  0.502\\
\hline
  k = 5  & 37.778 &  0.805 & 0.370 &-1.928 &0.855& 0.574\\
\hline
  k = 6  &-1.802 & 0.901 & 0.605 &-1.928 &0.982&  0.590\\
\hline
  k = 7  &-0.356 &  0.880 & 0.641 &-1.818 &0.865& 0.638\\
\hline
  k = 8  & -1.774 &  0.913 & 0.505 & -1.496 &0.913&  0.228\\
\hline
  k = 9  & -1.044 &  0.878 & 0.249 & -1.881 &0.919&  0.659\\
\hline
  k = 10  & -1.648 &  0.940 & 0.439 &  -1.883 &0.952&  0.695\\
\hline
\end{tabular}
\caption{Results for the exponential decay process for different time step $\Delta t$ observations.}
\label{table: exp multiscale}
\end{center}
\end{table}

\subsection{Trigonometric process}

\begin{figure}[H]
    \centering
    \includegraphics[width = 0.3\linewidth, height = 3cm]{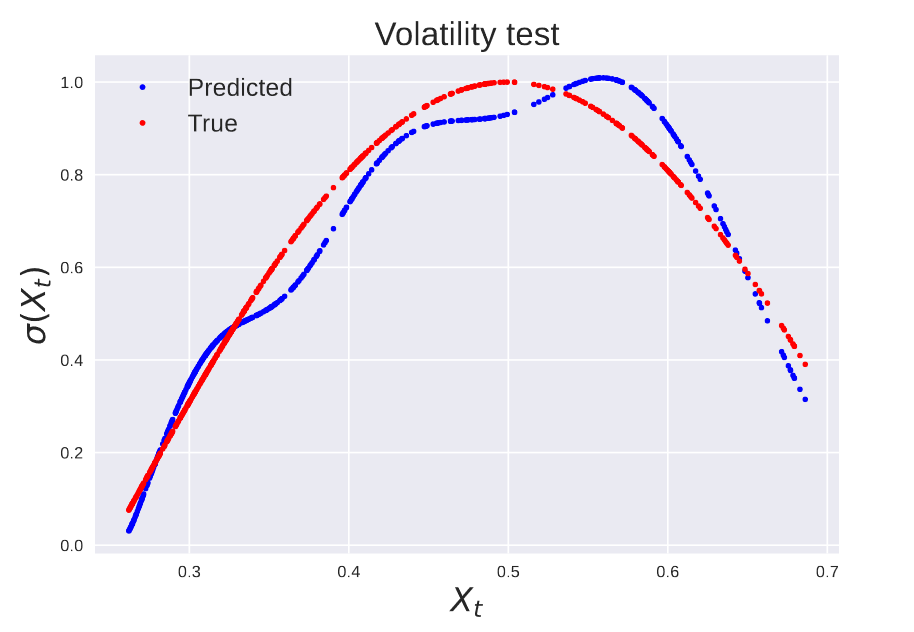}
    \includegraphics[width = 0.3\linewidth, height = 3cm]{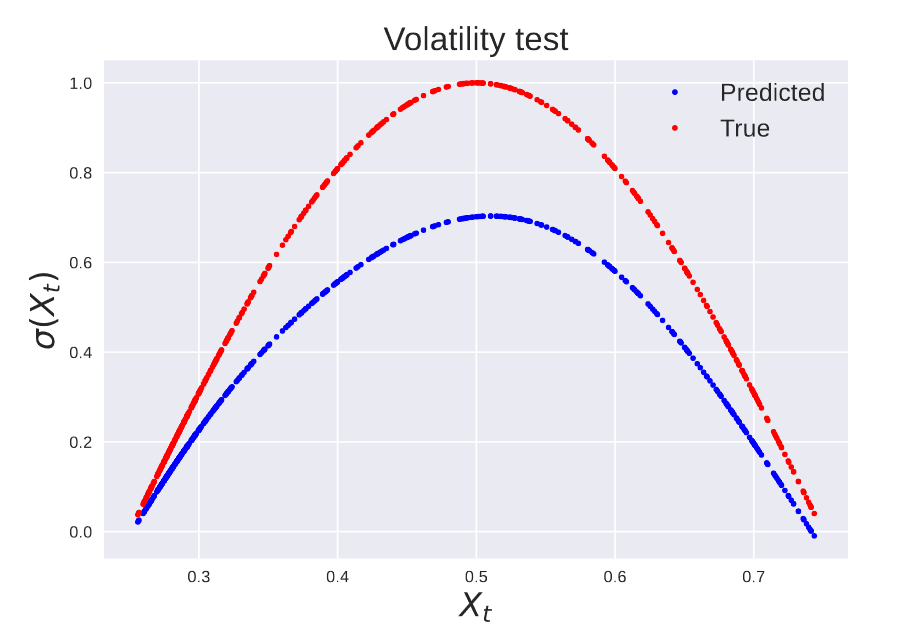}
    \includegraphics[width = 0.3\linewidth, height = 3cm]{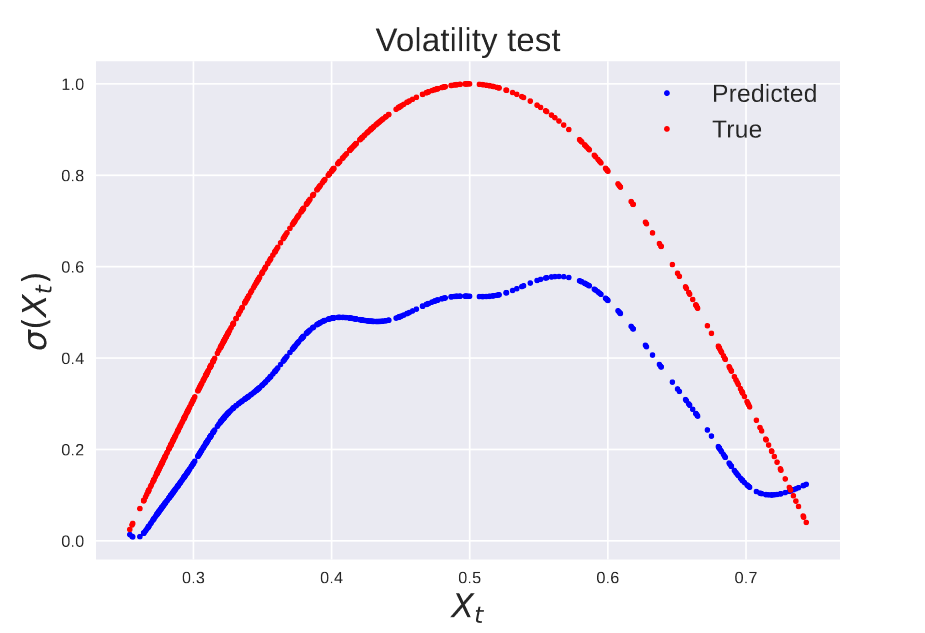}
    \caption{Trigonometric process: prediction of the volatility for different values of $k$. From left to right: $k=1,2,3$.}
    \label{fig: trig multiscale vol}
\end{figure}

\begin{figure}[H]
    \centering
    \includegraphics[width = 0.3\linewidth, height = 3cm]{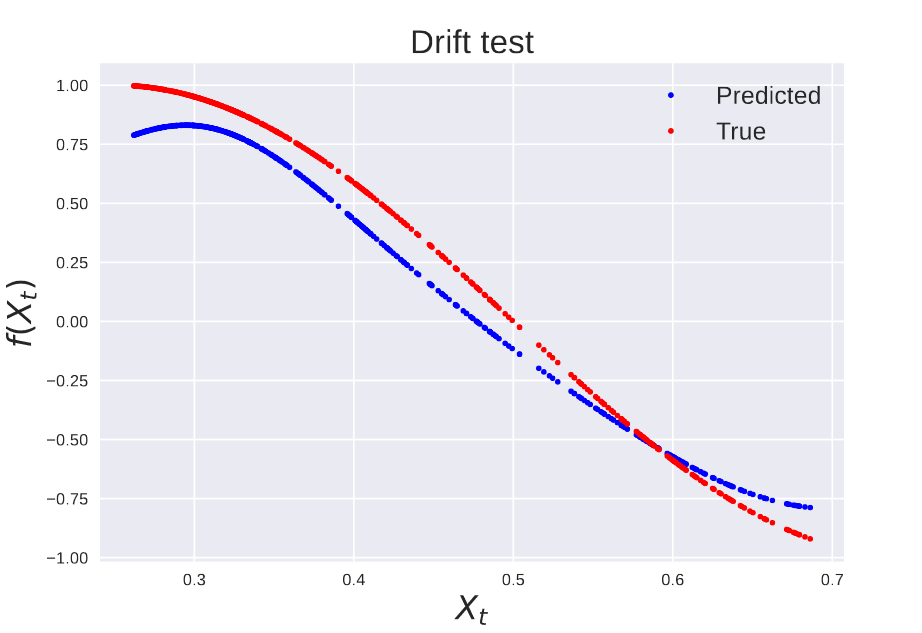}
    \includegraphics[width = 0.3\linewidth, height = 3cm]{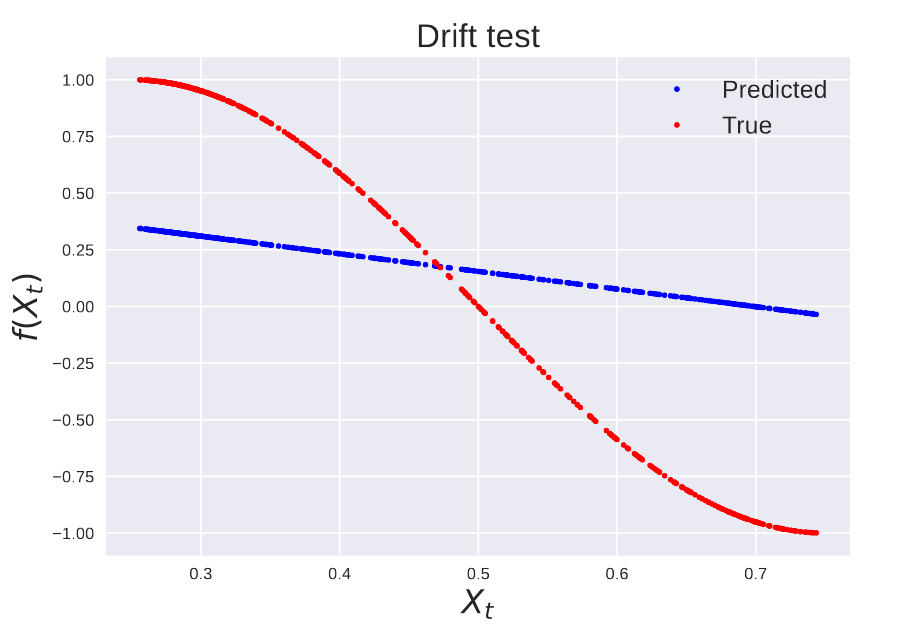}
    \includegraphics[width = 0.3\linewidth, height = 3cm]{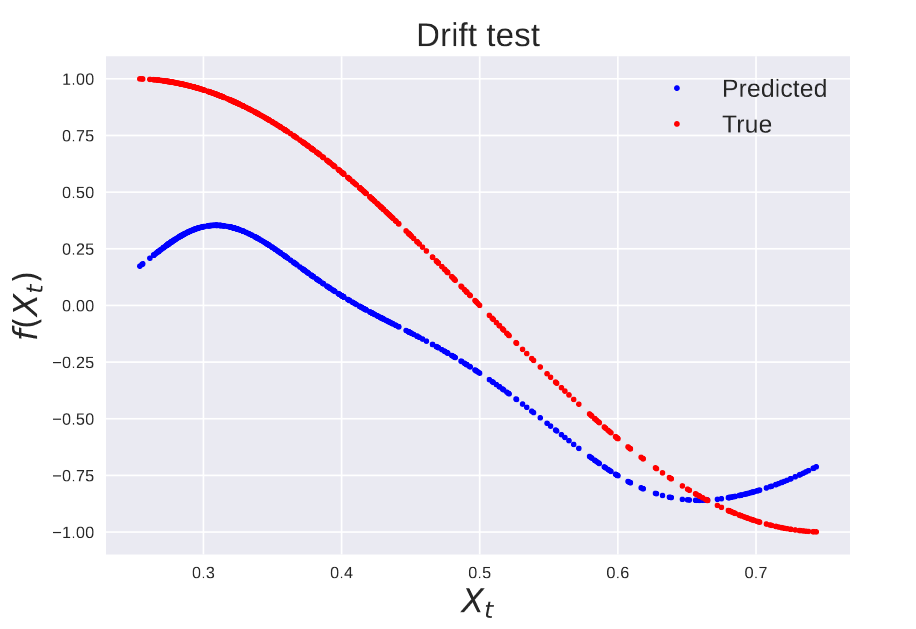}
    \caption{Trigonometric process: prediction of the drift for different values of $k$. From left to right: $k=1,2,5$.}
    \label{fig: trig multiscale drift}
\end{figure}

We now consider the trigonometric process described by \eqref{eq: trig process discrete} with time discretization $\Delta t = 0.001$. For each $k$, we set $N = 500$ for both the training and test sets. The results for $k$ are presented in table \ref{table: trig multiscale} and illustrative examples are presented in figures \ref{fig: trig multiscale vol} and \ref{fig: trig multiscale drift}.

\begin{table}[H]
\begin{center}
\begin{tabular}{|l|c|c|c|c|c|c|} 
 \hline
 & \multicolumn{3}{|c|}{Non optimized parameters} & \multicolumn{3}{|c|}{Optimized parameters} 
  \\ 
 \hline
  &$\mathcal{L}(\bar{f}^*, \bar{\sigma}^* |X, Y)$ & $\delta_f$ & $\delta_\sigma$  &$\mathcal{L}(\bar{f}^*, \bar{\sigma}^* |X, Y)$ &$\delta_f$& $\delta_\sigma$\\
  \hline
  k = 1  &-3.675 & 0.259 & 0.093 &  -3.674 & 0.163 & 0.092\\
\hline
  k = 2  &-3.883 &  0.717 & 0.300 &-3.807 &0.792 & 0.299\\
\hline
  k = 3  &-3.729 &  0.702 &0.419 & -3.725 &0.654& 0.434\\
\hline
  k = 4  &  -3.784 &  0.836 & 0.499 & -3.798 &0.848&  0.495 \\
\hline
  k = 5  & -4.027 &  0.904 & 0.587 &-4.019 &0.852& 0.593\\
\hline
  k = 6  &-3.910 & 0.911 & 0.594 &-3.942 &0.952&  0.597\\
\hline
  k = 7  & -3.909 &  0.954 &0.636 &-3.562 &0.925&0.211\\
\hline
  k = 8  &-3.652 &  0.772 & 0.668 &  -3.930 &0.767&  0.669\\
\hline
  k = 9  & -3.939 &  0.880 & 0.657 &-3.424&0.935&  0.145\\
\hline
  k = 10  & -3.505 &  0.888 & 0.254 & -3.926 &0.880& 0.705\\
\hline
\end{tabular}
\caption{Results for the trigonometric process for different time step $\Delta t$ observations.}
\label{table: trig multiscale}
\end{center}
\end{table}

 In general, we note that, as expected, as $k$ gets larger, our method has worse performance. Moreover, for values of $k \neq 1$, our cross-validation method does not perform as well, suggesting the need to adapt the cross-validation procedure to learning the parameter $\lambda$. There are, however some exceptions, such $k= 8, 9$ for the exponential volatility process and $k =7,9$ for the trigonometric process where one of the functions is better recovered. This suggests that a good choice of parameters via cross -validation leads to a good recovery even if the modeling assumption is incorrect, consistent with observations that cross-validation methods are somewhat robust to model misspecification \cite{chen2021consistency}.
\section{Experimental results: discussion}\label{sec: discussion}

We make the following observations regarding our results.

First, we observe that our method provides comparable or greater performance compared to simple kernel regression with hyper-parameters optimized through the minimization of the log marginal likelihood. Which method is preferable depends on the underlying SDE. We observe that for SDEs with constant volatility functions, such as the Ornstein–Uhlenbeck process, the simple kernel regression generally outperforms our method as measured by the likelihood. This is likely due to the better recovery of the volatility as the modeling assumption of the kernel regression (i.i.d. noise) better captures the true model (constant volatility). Nonetheless, our method does occasionally outperform kernel regression in predicting the drift of the SDE. Our method, however, notably outperforms kernel regression for SDEs with non-constant volatility, such as the trigonometric process and the exponential volatility process.

Second, we observe that the randomized cross-validation algorithm for hyper-parameter optimization reliably improves the performance of our method. In all cases, the selected parameters have better performance compared to the initial guess, as measured by the likelihood of the model. Moreover, as the Geometric Brownian Motion example illustrates, learning the parameters of the kernels can enable ill-specified kernels to perform comparably to well-specified kernels.

We also observe that a better likelihood generally implies a better capture of the drift and volatility. Hence, while these quantities are unobserved, better performance as measured by the likelihood generally implies that the model captures well both the drift and volatility. 

Finally, we note that as the observations increasingly depart from the true dynamics, our method performs more poorly. This suggests the need for a better model for the discretization error term, either through learning the parameter $\lambda$ or a better modeling assumption than the \emph{i.i.d} Gaussian noise presented in section \ref{sec:section2}.

\clearpage
\appendix

\section{Additional Plots.}\label{sec:section7}
In this section, we provide additional plots of the results of the numerical experiments.
\subsection{Exponential volatility process}
The following figures illustrate the results on the second trajectory of the exponential volatility process.

\begin{figure}[H]
    \centering
    \includegraphics[width = 0.23\linewidth, height = 3cm]{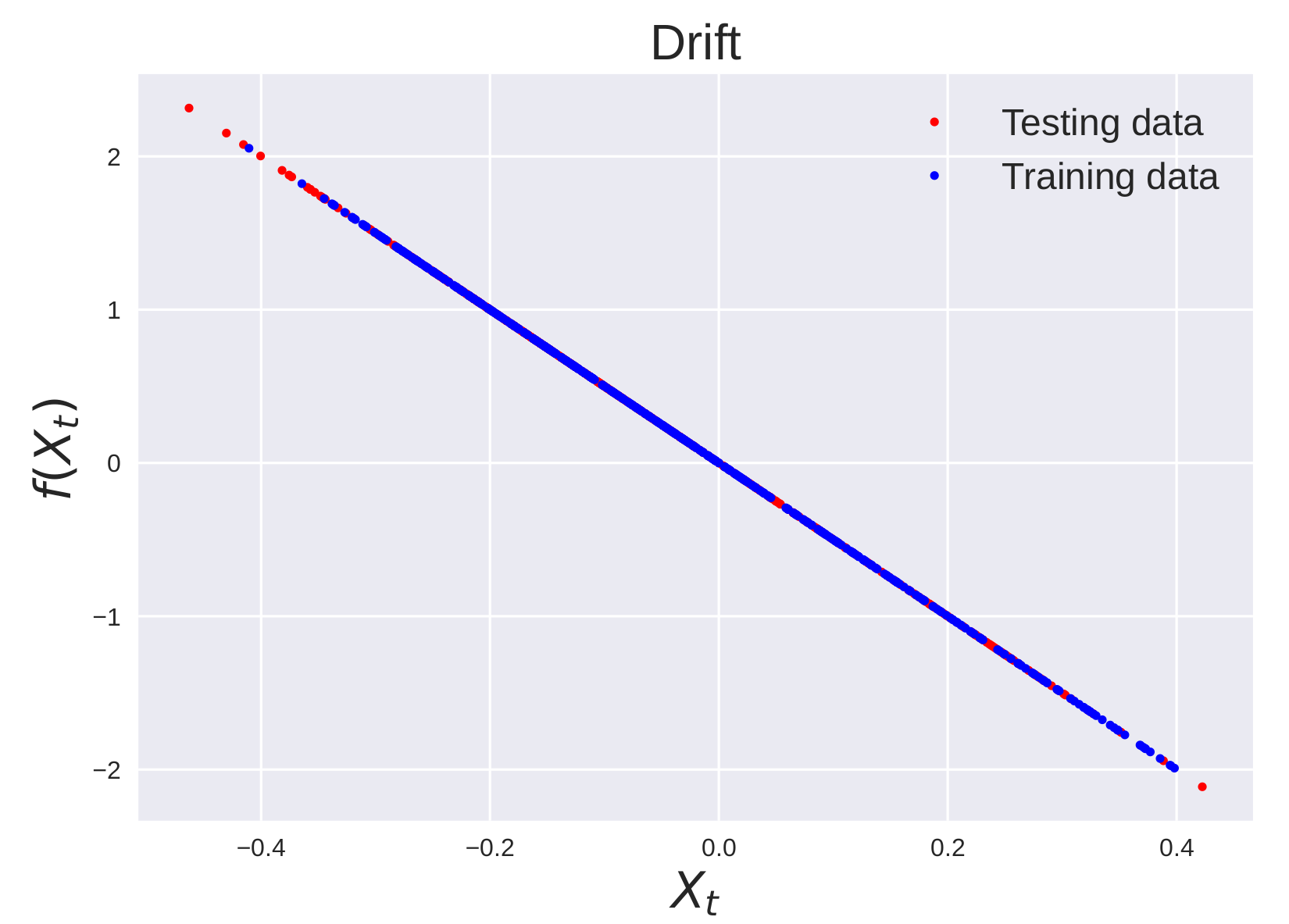}
    \includegraphics[width = 0.23\linewidth, height = 3cm]{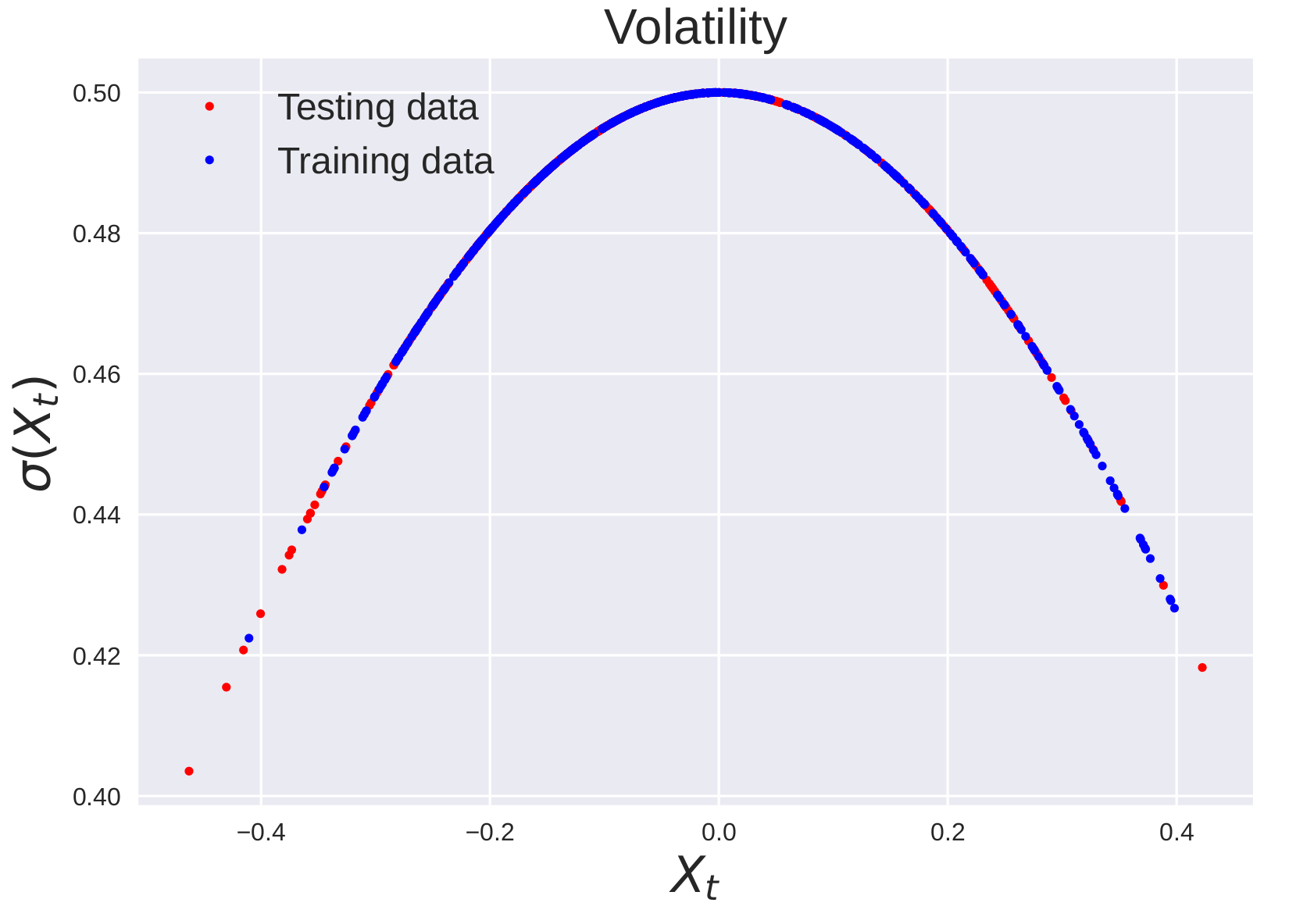}
    \includegraphics[width = 0.23\linewidth, height = 3cm]{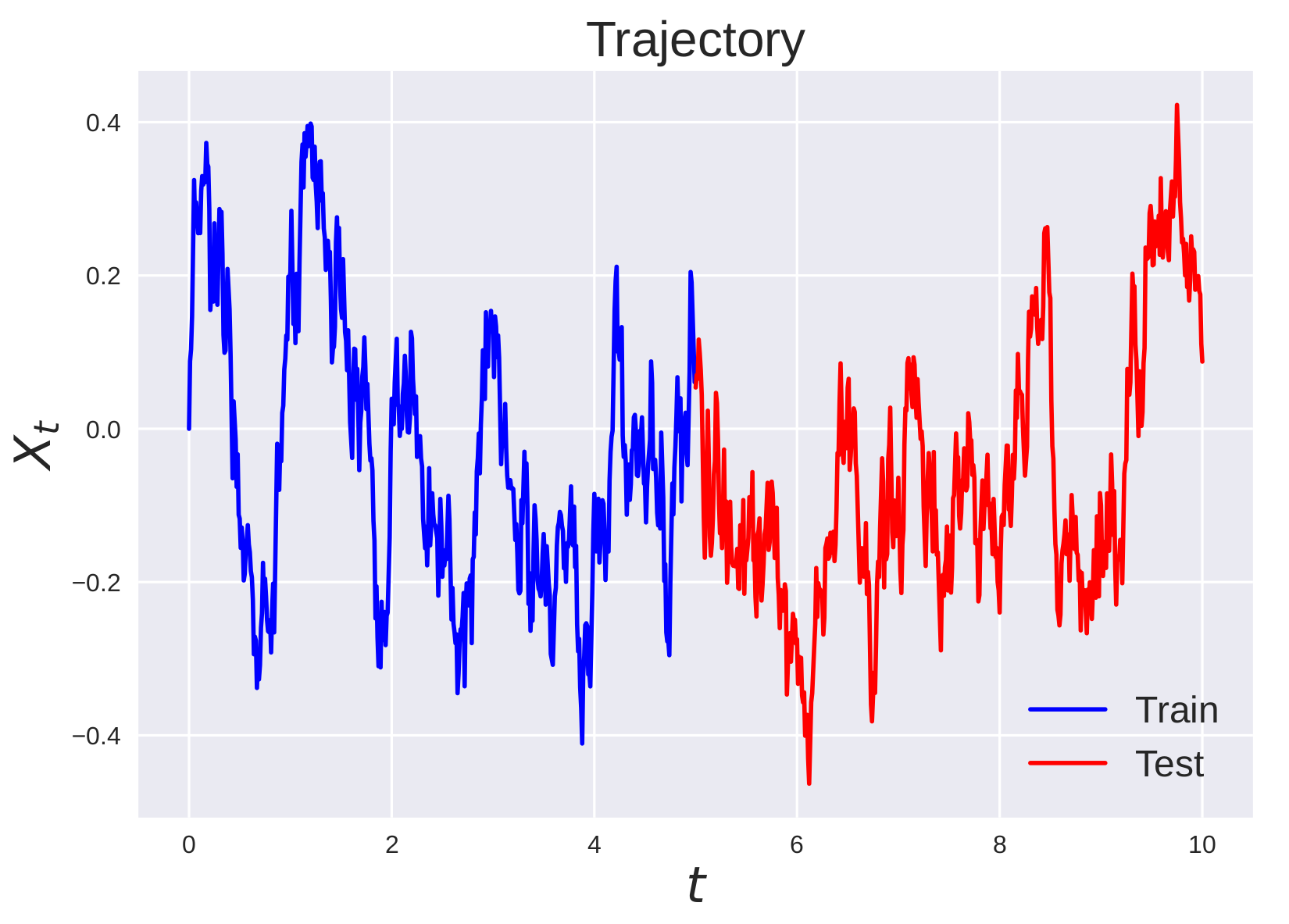}
    \includegraphics[width = 0.23\linewidth, height = 3cm]{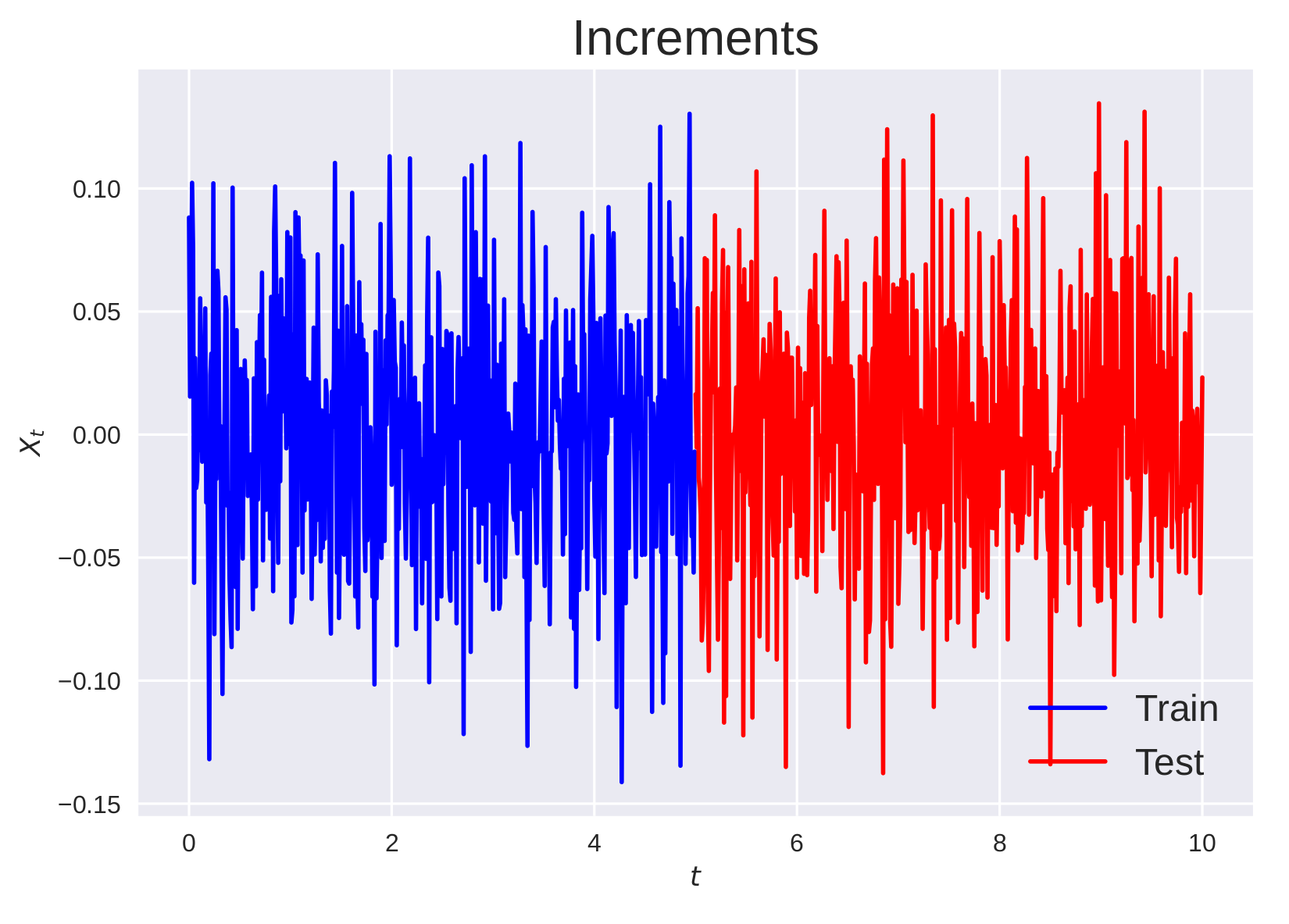}
    \caption{From left to right: drift function, volatility function, sample trajectory, and sample increments of the exponential volatility process.}
    \label{fig: data, exp vol 1}
\end{figure}

\begin{figure}[H]
    \centering
    \includegraphics[width = 0.23\linewidth, height = 3cm]{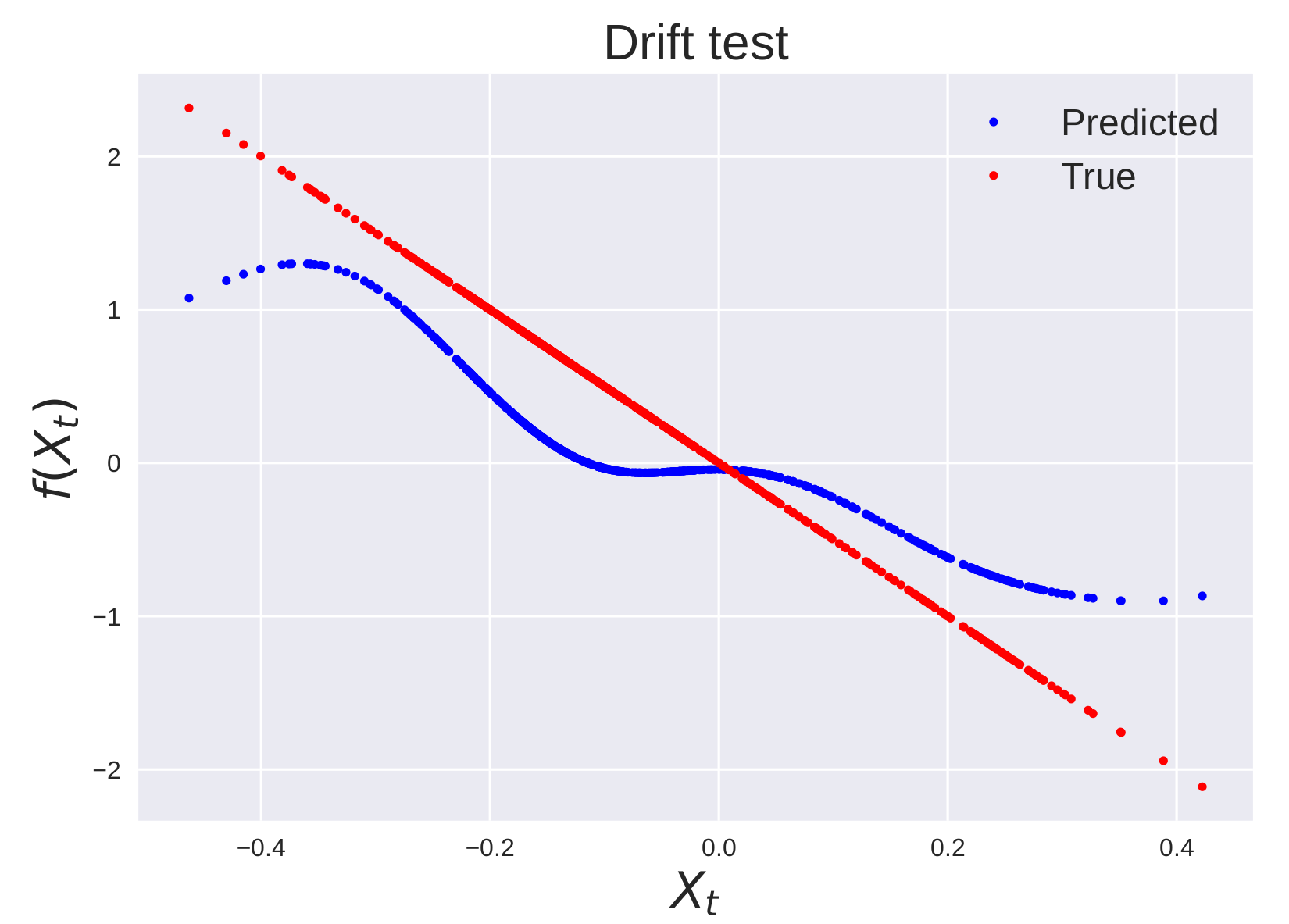}
    \includegraphics[width = 0.23\linewidth, height = 3cm]{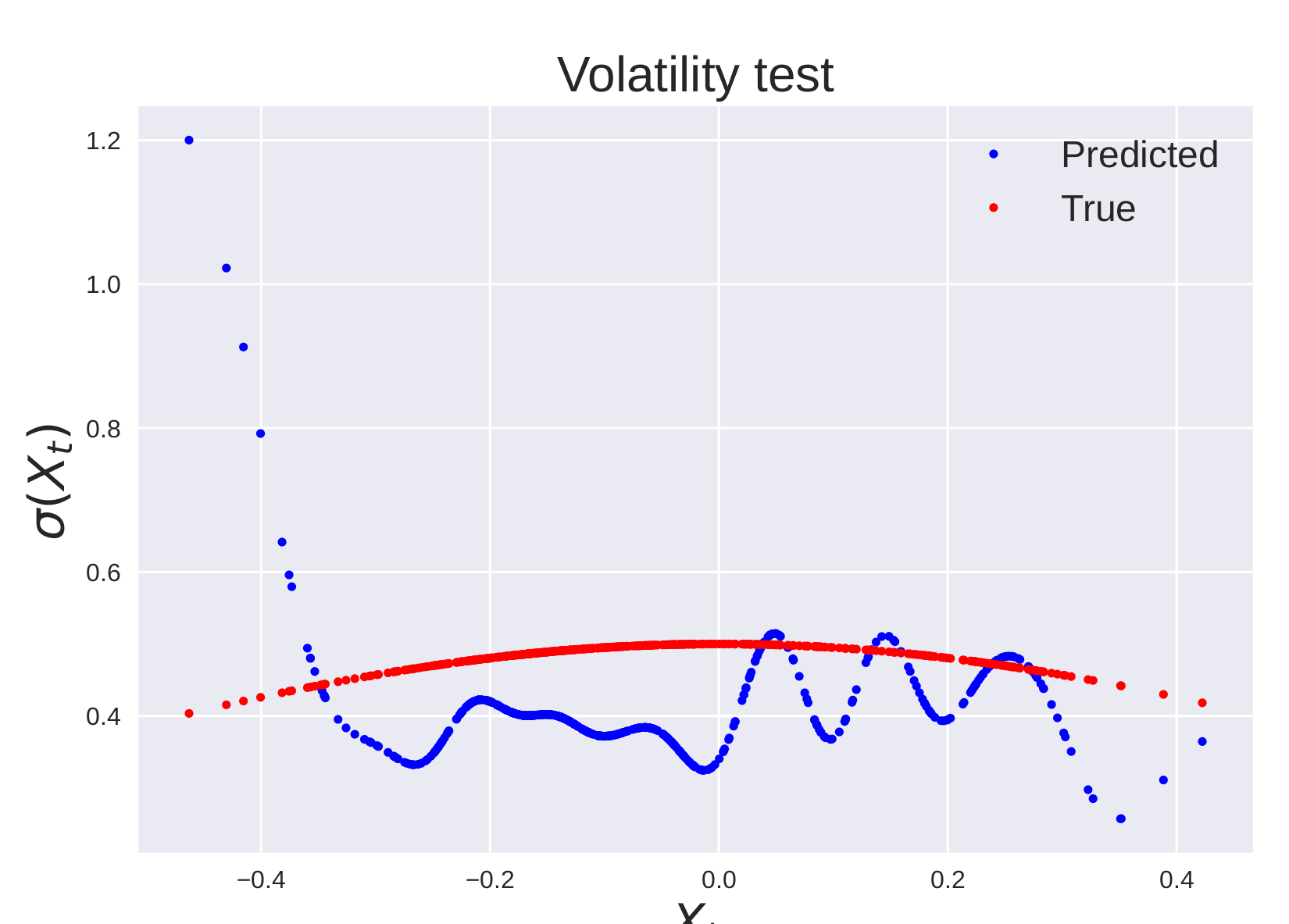}
    \includegraphics[width = 0.23\linewidth, height = 3cm]{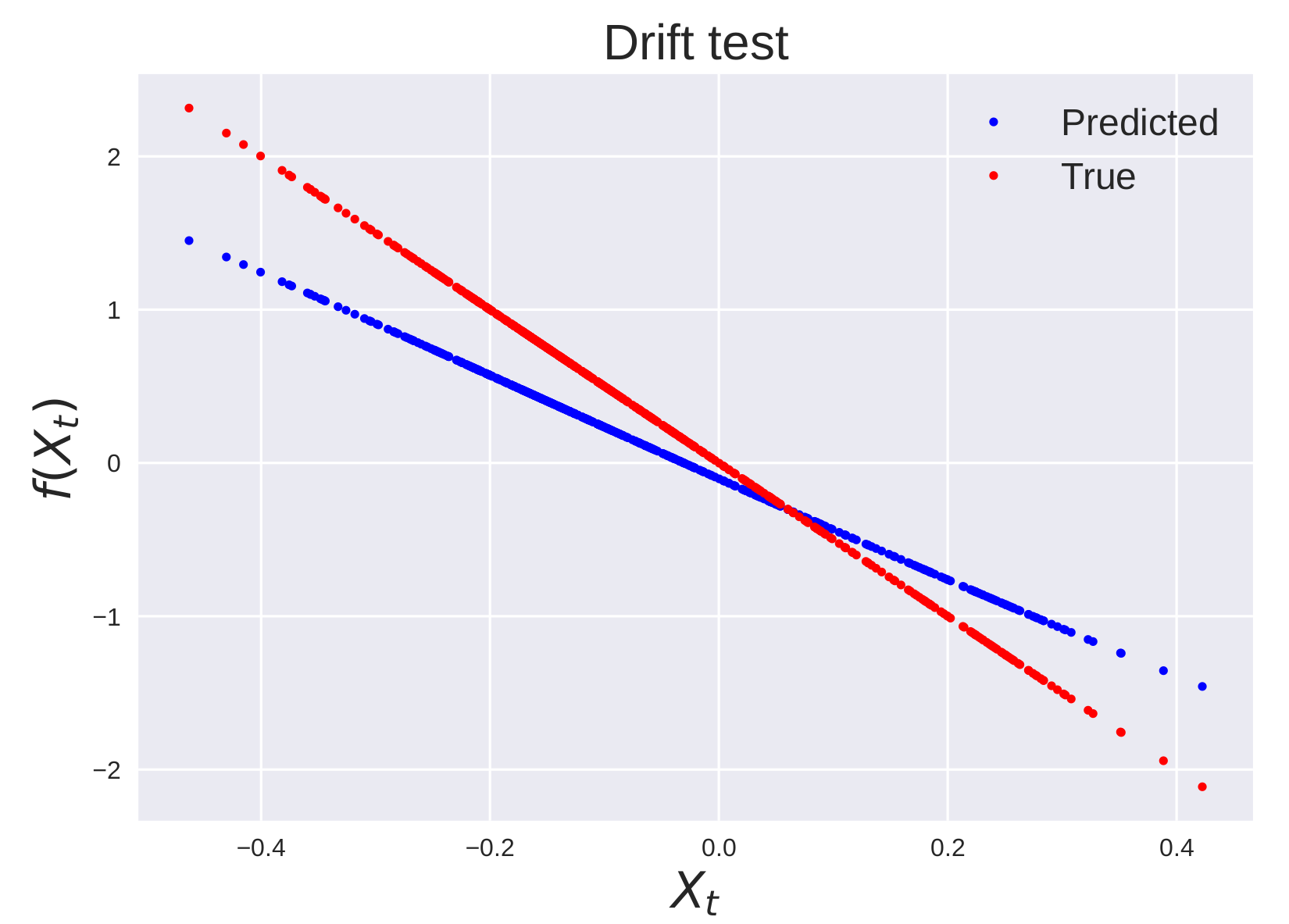}
    \includegraphics[width = 0.23\linewidth, height = 3cm]{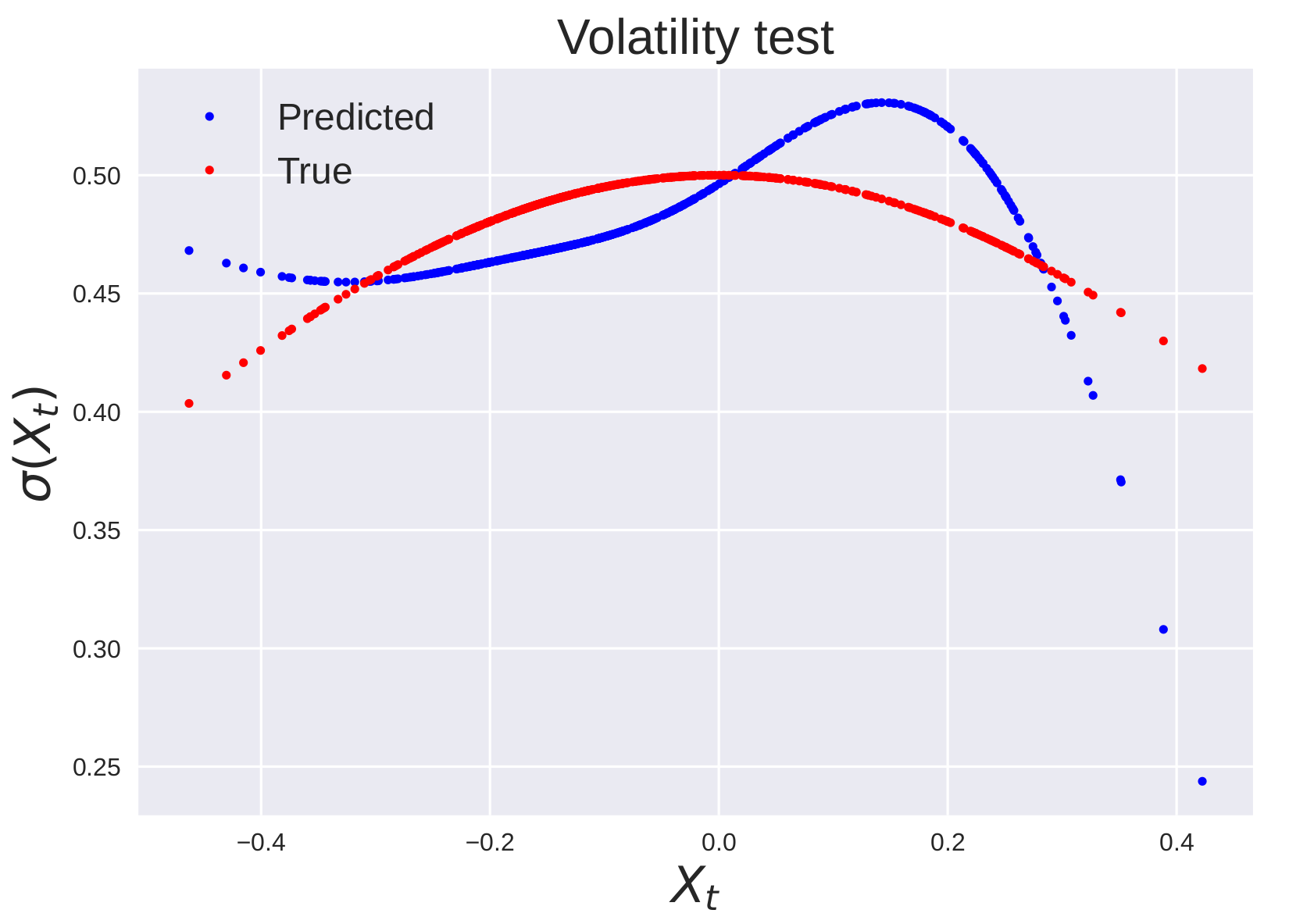}
    \caption{Predicted drift and volatility on the testing set for trajectory 1 of the exponential volatility process. From left to right: drift (non-learned kernel), volatility (non-learned kernel), drift (learned kernel), volatility (learned kernel) .}
    \label{fig: pred, exp vol 1}
\end{figure}

\subsection{Trigonometric process}

The following figures illustrate the results of the first trajectory of the trigonometric process.

\begin{figure}[H]
    \centering
    \includegraphics[width = 0.23\linewidth, height = 3cm]{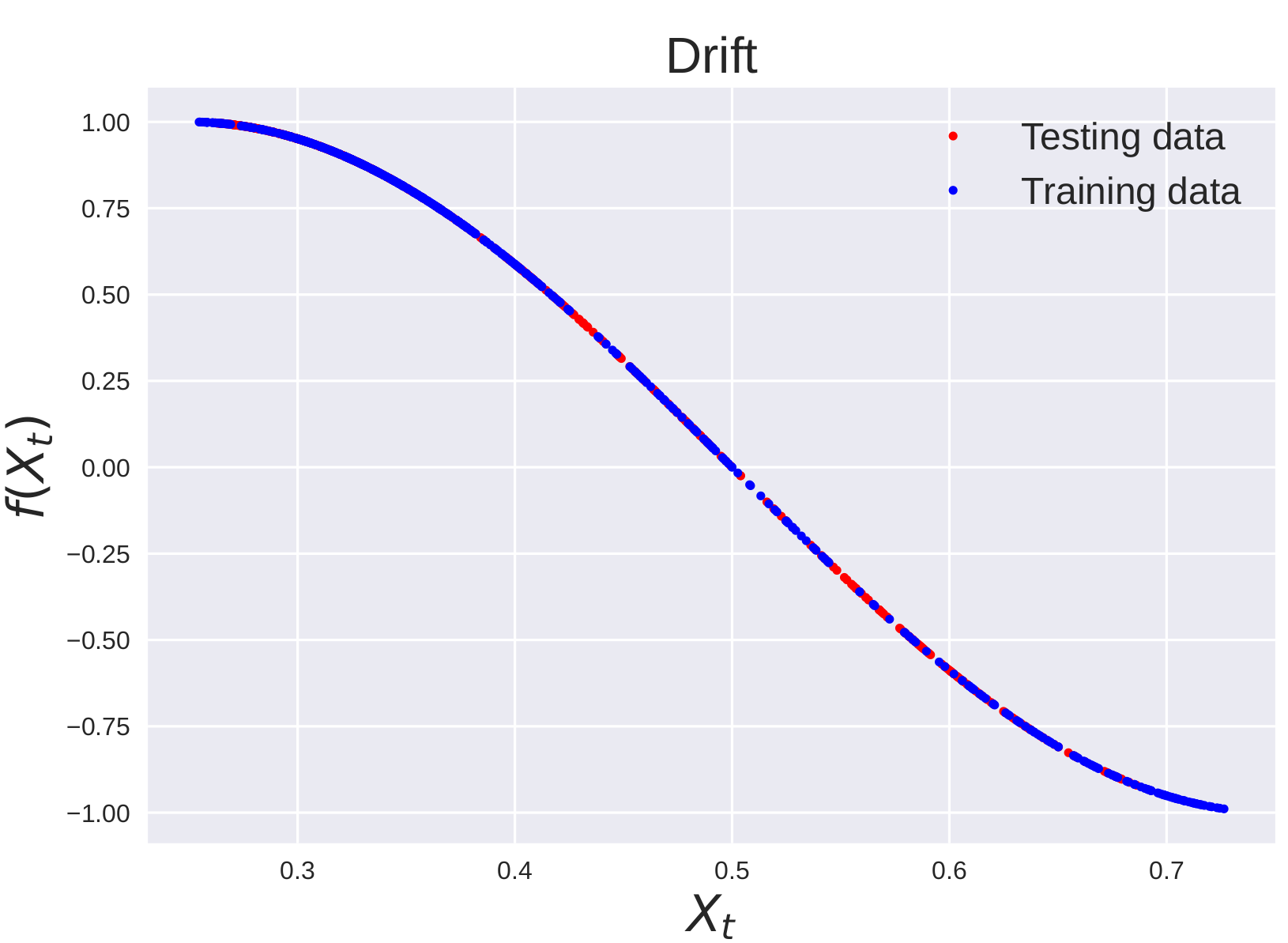}
    \includegraphics[width = 0.23\linewidth, height = 3cm]{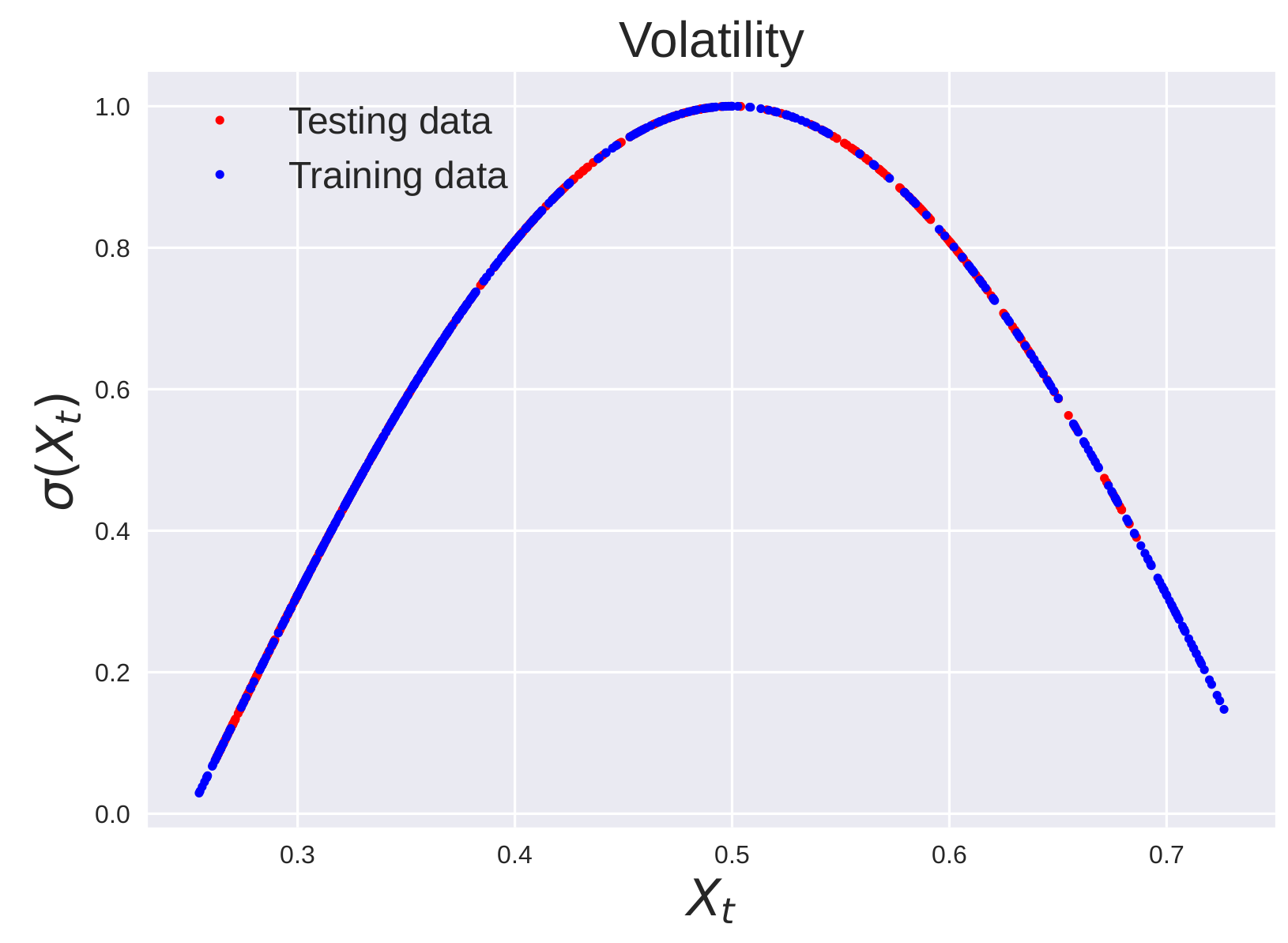}
    \includegraphics[width = 0.23\linewidth, height = 3cm]{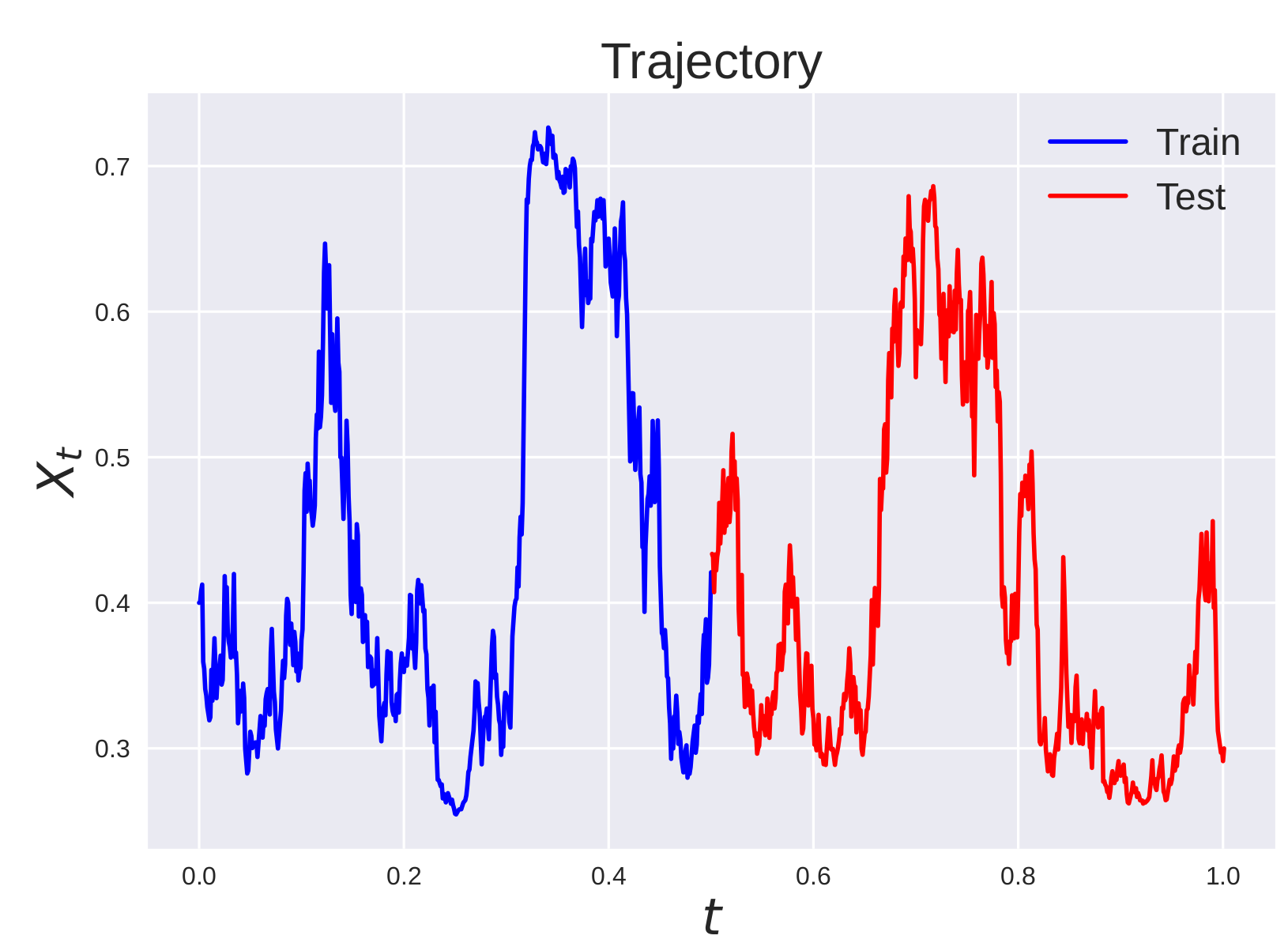}
    \includegraphics[width = 0.23\linewidth, height = 3cm]{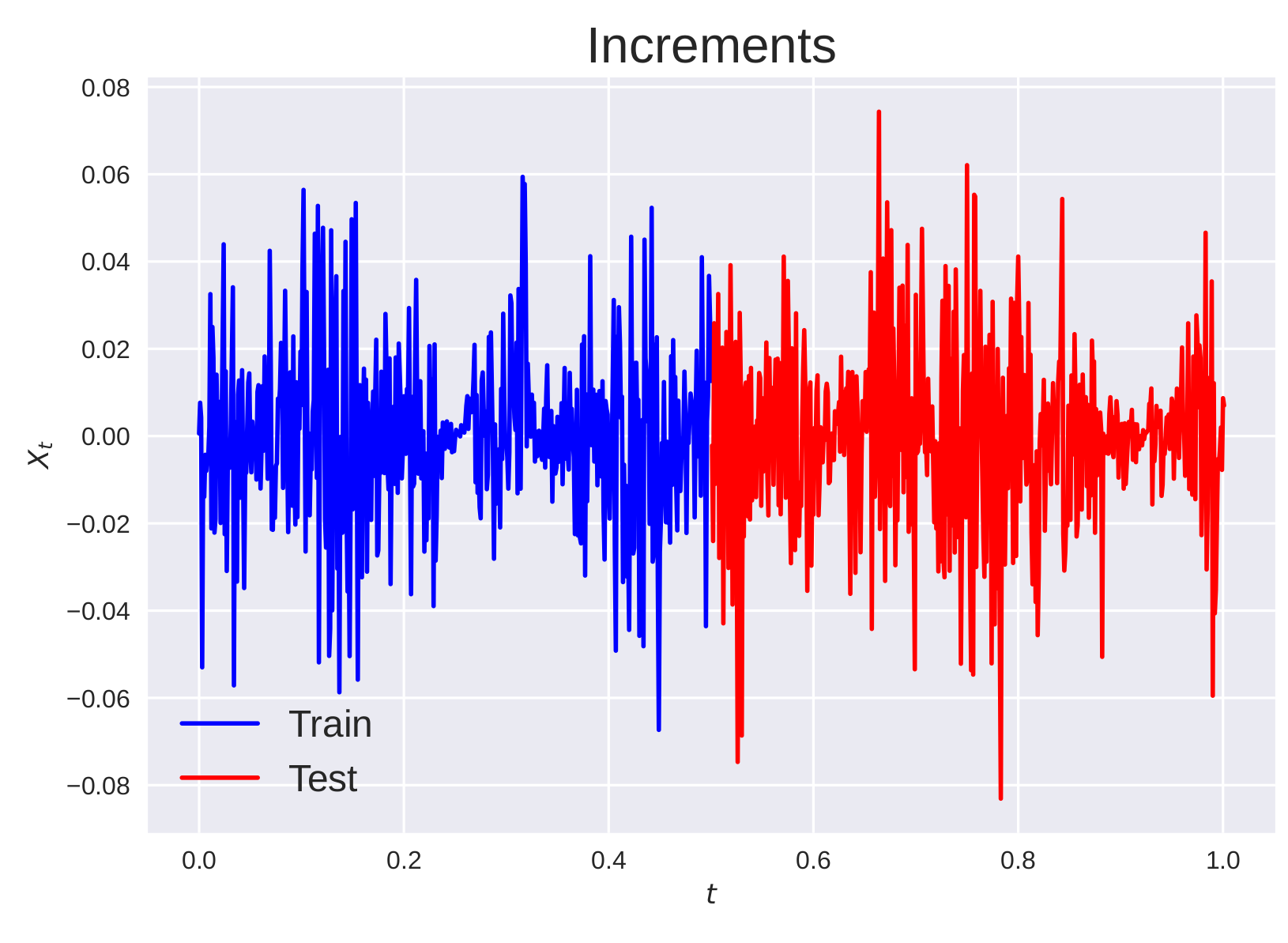}
    \caption{From left to right: drift function, volatility function, sample trajectory, and sample increments of the trigonometric process.}
    \label{fig: data, trig  1}
\end{figure}

\begin{figure}[H]
    \centering
    \includegraphics[width = 0.23\linewidth, height = 3cm]{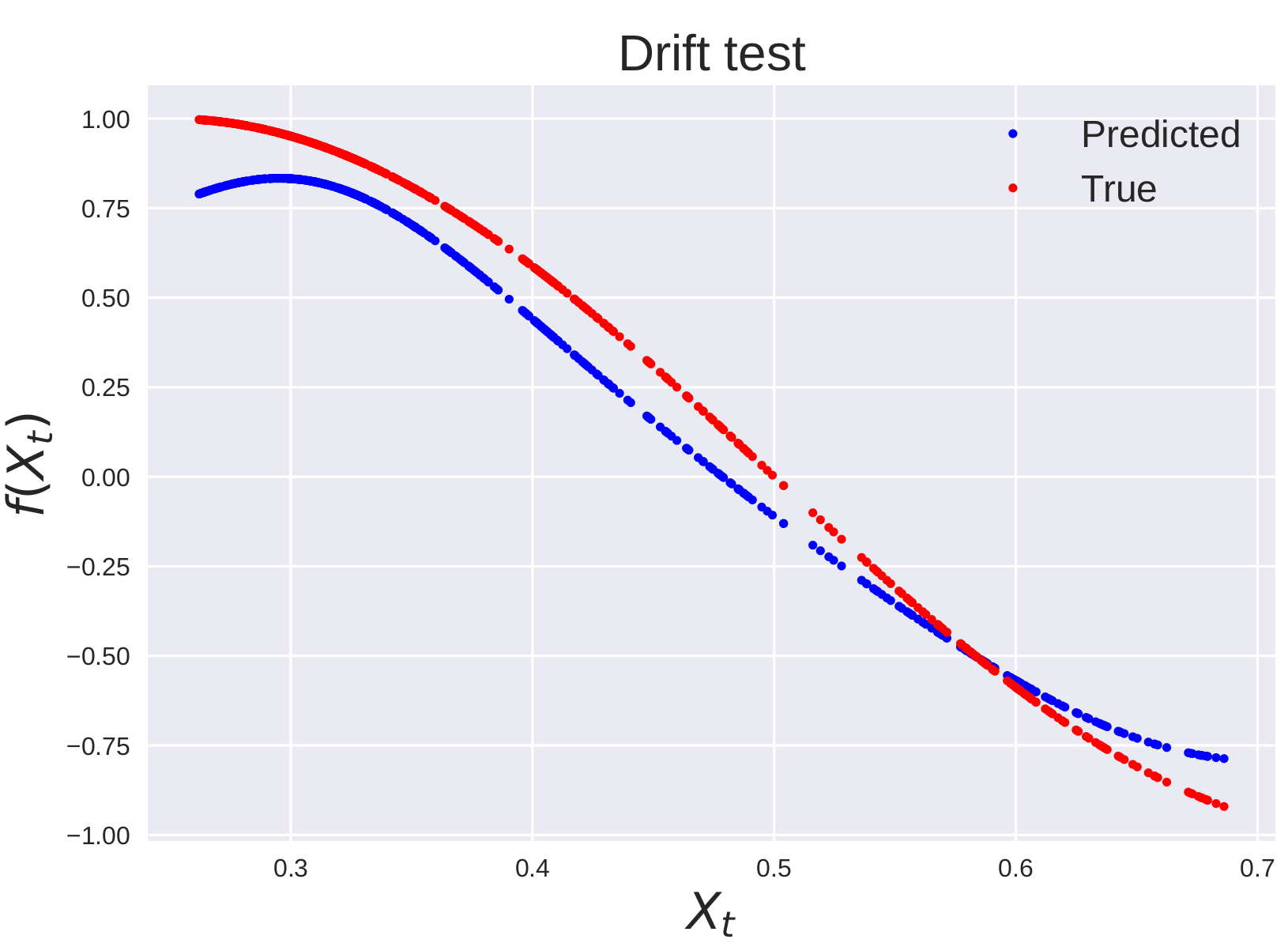}
    \includegraphics[width = 0.23\linewidth, height = 3cm]{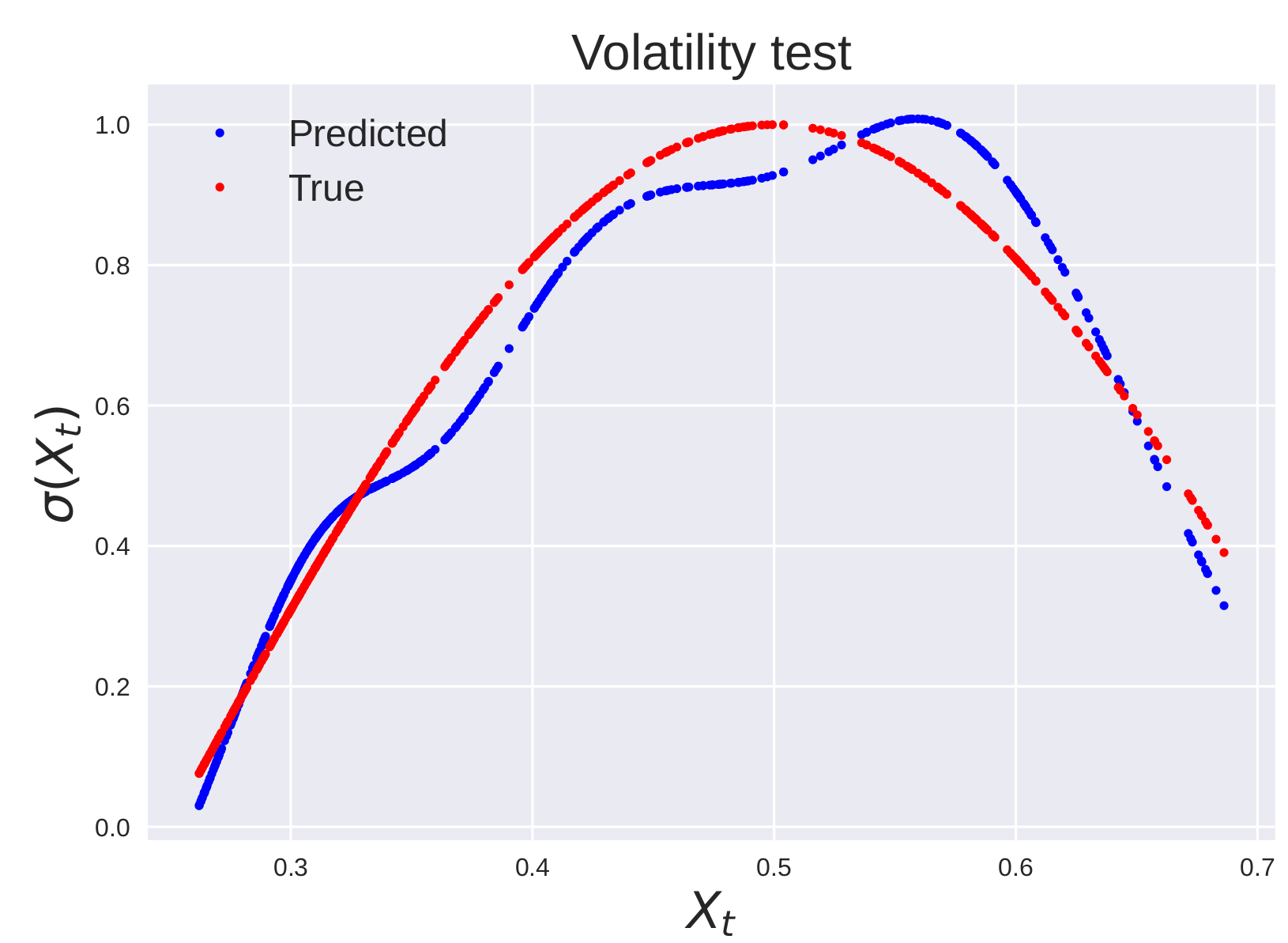}
    \includegraphics[width = 0.23\linewidth, height = 3cm]{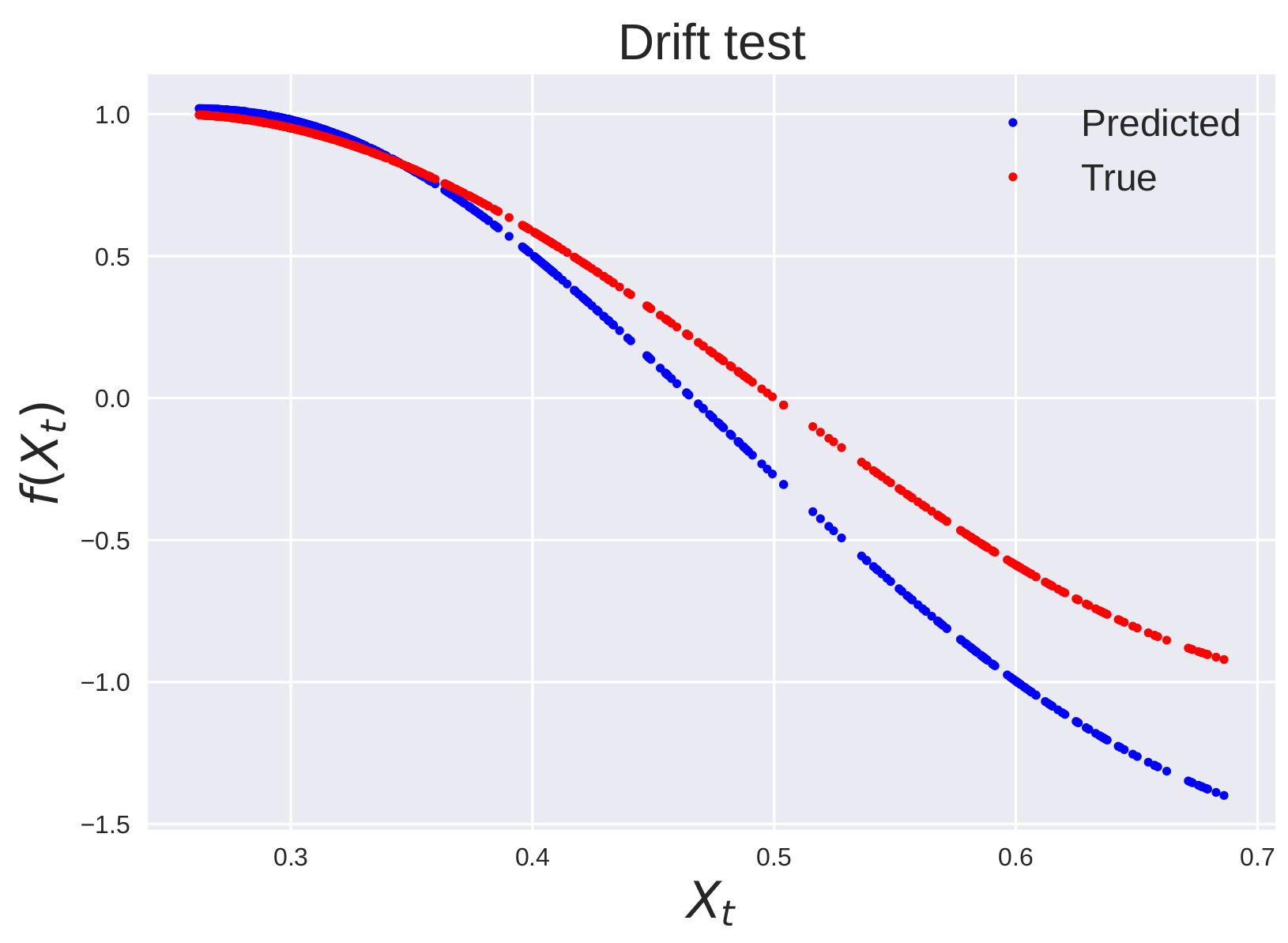}
        \includegraphics[width = 0.23\linewidth, height = 3cm]{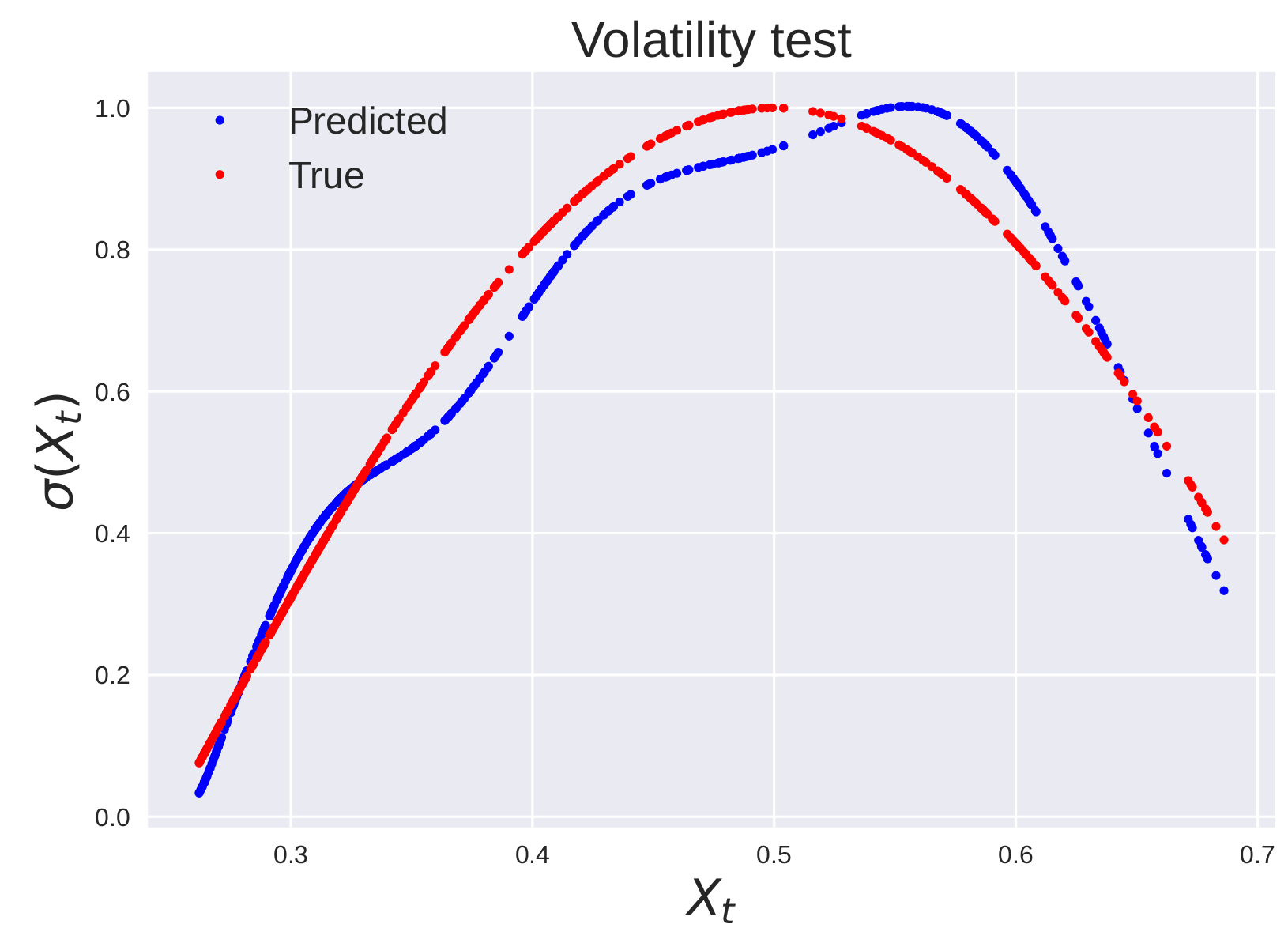}
    \caption{Predicted drift and volatility on the testing set for trajectory 1 of the trigonometric process. From left to right: drift (non-learned kernel), volatility (non-learned kernel), drift (learned kernel), volatility (learned kernel) .}
    \label{fig: pred, trig 1}
\end{figure}

\section{Gaussian Process Regression and Extension of \cite[Page 306]{bishop2006pattern}}
\label{app:gaussian_regression}
In this section, we give a very brief overview of Gaussian processes for regression. We suppose that the values of $Y(X)$ are distributed according to a Gaussian process, namely $Y(X) \overset{d}{\sim} \mathcal{GP}(\boldsymbol{O}, \boldsymbol{K})$. In particular, in the case of SDEs, given the data $(X, Y)$, the predicted drift and diffusion for a new point $x_*$ are given by

\begin{align*}
    &\bar{f}(x_*) = K(x_*, X)K(X,X)^{-1}Y \\
    &\bar{\sigma}(x_*) =  K(x_*, x_*) - K(x_*, X)K(X, X)^{-1}K(X, x_*)
\end{align*}

\noindent The above expressions are valid in noisy observation with independent and identical Gaussian noise. We derive this distribution in the case where the observations are noisy with independent, but not necessarily identical, Gaussian noise; the proof is generalized from the one presented in \cite[306]{bishop2006pattern}.\\
\indent Formally, suppose that we have at our disposal the noisy observations $(X_n, Y_n)_{1 \leq n \leq N}$, where $Y_i = f(X_i) + W_i$ and the $W_i$ are independent but not necessarily identically distributed $\mathcal{N}(0, \sigma^2(X_i))$ random variables. The problem is the identification of the unknown function $f$ given these noisy observations. Set $f_i = f(X_i)$, so that $\boldsymbol{f}:=(f_i)_{1 \leq i \leq N}$. In addition, set $\boldsymbol{Y} := (Y_i)_{1 \leq i \leq N}$. Observe that $\boldsymbol{Y} | \boldsymbol{f} \overset{d}{\sim} \mathcal{N}(\boldsymbol{f}, \boldsymbol{\Sigma})$ where $(\,\Sigma\,)_{ij} = \delta_{i j}\sigma^2(X_i)$, and assume that $\boldsymbol{f} \overset{d}{\sim} \mathcal{N} (\boldsymbol{0}, \boldsymbol{K})$. Then, \cite[Page 93]{bishop2006pattern}, 

\begin{equation*}
    p(\boldsymbol{Y}) = \int p(\boldsymbol{Y} | \boldsymbol{f})p(\boldsymbol{f})d\boldsymbol{f} = \mathcal{N}(\boldsymbol{0}, \boldsymbol{K + \Sigma}).
\end{equation*}
Now, let $\boldsymbol{C} = \boldsymbol{K + \Sigma}$ so that $\boldsymbol{C}_{ij} = k(X_i, X_j) + \sigma_i^2 \delta_{ij}$. Denote $\boldsymbol{Y}_{N+1} = (Y_1, ..., Y_{N+1})$, $\boldsymbol{Y}_{N} = (Y_1, ..., Y_{N})$. We wish to derive the conditional distribution $p(Y_{N+1} | \boldsymbol{Y}_{N})$. First, observe that 
\begin{equation*}
    p(\boldsymbol{Y}_{N+1}) = \mathcal{N}(\boldsymbol{0}, \boldsymbol{C}_{N+1}).
\end{equation*}
where $\boldsymbol{C}_{N+1}$ is the $(N + 1) \times (N+1)$ with entries defined as previously for the vector $\boldsymbol{Y}_ {N+1}$.
We may partition the covariance matrix as 
\begin{equation*}
   \boldsymbol{C}_{N+1} = \begin{pmatrix}
\boldsymbol{C}_{N} & \boldsymbol{K}\\
\boldsymbol{K}^T & c
\end{pmatrix}
\end{equation*}

\noindent where $c = K(X_{N+1}, X_{N+1}) + \sigma_{N+1}^2$ and $\overline{\boldsymbol{K}}$ is the vector with entries $\overline{\boldsymbol{K}}_{i} = K(X_{N+1}, X_i)$. Then $p(Y_{N+1} | \boldsymbol{Y}_N) = \mathcal{N}(m(X_N), \sigma^2(X_N))$ and the conditional mean and covariance are given by

\begin{align*}\label{eq: conditional }
    m(X_{N+1}) = \overline{\boldsymbol{K}}^T\boldsymbol{C}^{-1}\boldsymbol{y}_N = \overline{\boldsymbol{K}}^T(\boldsymbol{K + \Sigma})^{-1}\boldsymbol{y}_N \\
    \sigma^2(X_N) = c - \overline{\boldsymbol{K}}^T\boldsymbol{C}^{-1}\overline{\boldsymbol{K}} = c - \overline{\boldsymbol{K}}^T(\boldsymbol{K + \Sigma})^{-1}\overline{\boldsymbol{K}}.
\end{align*}

Note that in our case, the observations actually of the form $Y_i = \Delta t_i f(X_i) + W_i$. The same proof holds with the slight modification that the entries of the covariance matrix now become
\begin{align*}
    &\boldsymbol{C}_{ij} = \Delta t_i \Delta t_j k(X_i, X_j) + \sigma_i^2 \delta_{ij} \\
    &\overline{\boldsymbol{K}}_{i} = \Delta t_iK(X_{N+1}, X_i)
\end{align*}
which yields equations \eqref{eq:eq_2} and \eqref{eq:eq_3}.
\section{Log-marginal likelihood for hyper-parameter optimization}
\label{app:logmarginal}
In this section, we briefly present the log-marginal likelihood method for hyper-parameter optimization. The marginal log-likelihood over the kernel parameters can be expressed as:  
\begin{equation*}
    -\log (p(Y |\boldsymbol{\theta}, X)) \propto \frac{1}{2}Y^TK(X, X)^{-1}Y + \log(|K(X, X)|).
\end{equation*}

\noindent Therefore, the optimal parameters $\boldsymbol{\theta}$ are obtained via the minimization of the function in the previous equation (with respect to $\boldsymbol{\theta}$). In the case of noisy observations, the kernel $K$ can be defined as 
\begin{equation*}
    K(X_1, X_2) = K^{'}(X_1, X_2) + \delta(X_1, X_2)
\end{equation*}
where $K'$ is a standard kernel and $\delta$ is the white noise kernel defined as 
$$
\delta(X_1, X_2) := \begin{cases}
			c & \text{if} \; X_1 = X_2, \\
            0 & \text{otherwise},
		 \end{cases}
$$
where $c$ is the noise level, a kernel parameter that must be optimized. It is important to note that this kernel can only account for a constant level of noise, which is not the case for many SDEs.

\section{Randomized cross-validation for hyper-parameter learning}
\label{app:Randomized_CV_Learning}
In this section, we describe the general framework of \emph{Randomized cross-validation} we use in our optimization.\\
The general problem is the following: we have a set of training data $(\boldsymbol{X}, Y) := \{(X_i, Y_i)\}_{1 \leq i \leq N}$ and we are trying to recover a function $f:\boldsymbol{X} \rightarrow Y$. In particular, we assume that we have a class of functions $\mathcal{S}$ indexed by a set of parameters $\boldsymbol{w}_p \in \mathcal{W}$, i.e.~$\mathcal{S}:= \{ f(\,\cdot\,,\boldsymbol{w}_p)\,:\,\boldsymbol{w}_p \in \mathcal{W}\,\,\text{and}\,\,f(\,\cdot\,,\boldsymbol{w}_p) : \boldsymbol{X} \rightarrow Y\}$. Notice that in practice, this could be approximated via reproducing kernels as in this paper or via neural networks. We assume that we have a \emph{method} to find the optimal parameter $\boldsymbol{w}^{*}_{p} \in \mathcal{W}$ for a given training set of data $(\boldsymbol{X}, Y)$. In our case, we assume that such a method involves the minimization of some loss function $\mathcal{L}(f(\boldsymbol{X}, \boldsymbol{w_p}), Y)$ with respect to $\boldsymbol{w}_p$, where $\mathcal{L}\,:\,Y \times Y \rightarrow \mathbb{R}$. Moreover, like in many Machine Learning (ML) algorithms, we assume that we have a set of hyper-parameters $\boldsymbol{\theta}  \in \boldsymbol{\Theta}$ that affect the recovery of the optimal $\hat{f} \in \mathcal{S}$. Such hyper-parameters $\boldsymbol{\theta}$ can parametrize the function $\hat{f} := \hat{f}_{\boldsymbol{\theta}} := f(\,\cdot\,;\boldsymbol{w}_p,  \boldsymbol{\theta})$, regularize the loss function $\mathcal{L}$ (often through some prior distribution on the parameters $\boldsymbol{w}_p$) or affect the minimization of the loss $\mathcal{L}$ (such as the learning rate of a gradient descent). We summarize this point by saying that $\hat{f}$ is recovered by minimizing $\mathcal{L}_{\boldsymbol{\theta}}$.\\
In the present work, in order to recover the function $f:\boldsymbol{X} \rightarrow Y$, we apply the Robust Learning Algorithm, whose rationale is explained in the following paragraph.
\paragraph{\bf Robust Learning Algorithm via Randomized cross-validation.}
Many ML algorithms are built upon minimizing a loss function $\mathcal{L}:Y\times Y\rightarrow\mathbb{R}$ over the set of parameters $(\boldsymbol{w}_{p}, \boldsymbol{\theta})$ of a class of models $\hat{f}\left(\,\cdot\,,\boldsymbol{w}_{p};\boldsymbol{\theta}\right)$. In other words, \emph{ideally}, we can define the risk function $R$ as the expected value of the loss function $\mathcal{L}$ with respect to the data probability density function $p\left(\boldsymbol{x},y\right)$:
\begin{equation}
R\left(\boldsymbol{w}_{p}, \boldsymbol{\theta} \right):=\mathbb{E}_{(\boldsymbol{X}, Y) \sim p\left(\boldsymbol{x},y\right)}\left\{ \mathcal{L}\left(\hat{f}(\boldsymbol{X}, \boldsymbol{w}_{p} ;\boldsymbol{\theta}), Y \right) \right\}
\label{eq: Ideal risk}
\end{equation}
and then find the set of optimal parameters according to 
\begin{equation}
\left(\boldsymbol{w}_{p}^{*},\boldsymbol{\theta}^{*}\right) :=\underset{\boldsymbol{w}_{p},\boldsymbol{\theta}}{\arg\min}\,\,R\left(\boldsymbol{w}_{p},\boldsymbol{\theta}\right).\label{eq: Ideal argmin}
\end{equation}
Notice that, in this setup, by the assumption that one has access to $p\left(\boldsymbol{x},y\right)$,
there is no theoretical distinction between parameters and hyper-parameters.\\
\emph{However}, in practice, one only sees a realization of $\left(\boldsymbol{X},Y\right)$, namely $\mathcal{D}=\left\{ \left(\boldsymbol{X}_{i},Y_{i}\right)\right\} _{i\in\mathcal{I}}$. Therefore, it is impossible 
to use Equations~\eqref{eq: Ideal risk} and \eqref{eq: Ideal argmin}. Moreover, in order to achieve generalization, in practice, $\boldsymbol{w}_p$ is optimized by minimizing a loss function dependent on $\boldsymbol{\theta}$:
\begin{equation}
    \boldsymbol{w}_p^{*} = \underset{\boldsymbol{w}_{p}}{\arg\min} \, \mathcal{L}_{\boldsymbol{\theta}}\left(\hat{f}\left(\boldsymbol{X}; \boldsymbol{w}_{p}, \boldsymbol{\theta}\right),Y\right)
\end{equation}
At this point, the best parameters $\boldsymbol{\theta}$ can be chosen via a cross-validation approach where the function $\hat{f}_{\boldsymbol{\theta}}$ is evaluated on an unseen test set $(\boldsymbol{X_u}, Y_u)$: 

\begin{equation}
    R(\boldsymbol{\theta}) = \mathcal{L}\left(\hat{f}\left(\boldsymbol{X}_{u}; \boldsymbol{w}^{*}_{p}, \boldsymbol{\theta}\right), Y_u\right)
\end{equation}
\noindent Generally, the optimization of hyper-parameters \cite{bergstra2011algorithms,bergstra2012random, hutter2011sequential,snoek2012practical,maclaurin2015gradient,franceschi2017forward} is usually done on a prefixed number of cross validation sets. In this work, we propose an approach that is not bound to a fixed number of cross-validation sets. Our algorithm can be summarized as follows:

\begin{enumerate}
\item[(1)] Partition the available data $\mathcal{D}=\left\{ \left(\boldsymbol{X}_{i},Y_{i}\right)\right\} _{i\in\mathcal{I}}$ into a training subset $\mathcal{D}_{\mathcal{T}}=\left\{ \left(\boldsymbol{X}_{i},Y_{i}\right)\right\} _{i\in\mathcal{T}}$ and a test subsets $\mathcal{D}_{\mathcal{U}}=\left\{ \left(\boldsymbol{X}_{i},Y_{i}\right)\right\} _{i\in\mathcal{U}}$; the two subsets are mutually exclusive. 
\item[(2)] Randomly partition the training set $\mathcal{D}_{\mathcal{T}}$ into two mutually
exclusive subsets of almost equal size: $\mathcal{D}_{\prod}=\left\{ \left(\boldsymbol{X}_{i},Y_{i}\right)\right\} _{i\in\prod\left(\mathcal{T}\right)}$
and $\mathcal{D}_{\prod^{c}}=\left\{ \left(\boldsymbol{X}_{i},Y_{i}\right)\right\} _{i\in\prod^{c}\left(\mathcal{T}\right)}$.
Here, $\prod\left(\mathcal{T}\right)$ returns the first half of the
random permutation of indices in $\mathcal{T}$ and $\prod^{c}\left(\mathcal{T}\right)$
returns the second half of the same permutation. 
\item[(3)] Train a ML model on $\mathcal{D}_{\prod}$ and evaluate the random
loss on $\mathcal{D}_{\prod^{c}}$ representing a realization of the
generalization error. 
\item[(4)] Repeat steps (2) and (3) to optimize the expected loss over the random
sets $\mathcal{D}_{\prod^{c}}$ with respect to $\boldsymbol{w}_{p}$. 
\item[(5)] Check the goodness of fit by evaluating the loss over $\mathcal{D}_{\mathcal{U}}$.
\end{enumerate}
More precisely, the optimization problem is the following one: 
\begin{equation}
\begin{array}{ll}
& \boldsymbol{\theta}^{*}  = \underset{\boldsymbol{\theta}}{\arg\min}\,\,\mathbb{E}_{\varPi}\left\{ R_{\mathbb{I}-\prod}\left(\bar{\boldsymbol{w}}_{p},\boldsymbol{\theta}\right)\right\} \\
\text{\,\,\,\,\,\,\,\,\,\,\,\,\,\,\, s.t.} & \bar{\boldsymbol{w}}_{p}=\underset{\boldsymbol{w}_{p}}{\arg\min}\,\,R_{\prod}\left(\boldsymbol{w}_{p},\boldsymbol{\theta}\right).
\end{array}\label{eq: Method}
\end{equation}
Where, 
\begin{equation}
R_{\prod}\left(\boldsymbol{w}_{p},\boldsymbol{\theta} \right)=\mathbb{E}_{\boldsymbol{X},Y\sim\mathcal{D}_{\prod}}\left\{
\mathcal{L}_{\boldsymbol{\theta}}\left(\hat{f}\left(\boldsymbol{X};\boldsymbol{w}_{p}, \boldsymbol{\theta}\right), Y\right)\right\},\label{eq: Risk on Train}
\end{equation}
and
\begin{equation}
R_{\mathbb{I}-\prod}\left(\boldsymbol{w}_{p},\boldsymbol{\theta}\right)=\mathbb{E}_{\boldsymbol{X},Y\sim\mathcal{D}_{\prod^{c}}}\left\{
\mathcal{L}\left(\hat{f}\left(\boldsymbol{X};\boldsymbol{w}_{p}, \boldsymbol{\theta}\right), Y\right)\right\}.\label{eq: Risk on Eval}
\end{equation}
In the previous Equations, $\mathbb{E}_{\boldsymbol{X},Y\sim\mathcal{D}_{\prod}}\left\{\, .\,\right\} $
means that the expected value is taken over the empirical distribution
defined by $\mathcal{D}_{\prod}$. A similar notion applies to $\mathbb{E}_{\boldsymbol{X},Y\sim\mathcal{D}_{\prod^{c}}}\left\{\,.\,\right\} $.
Furthermore, $\mathbb{E}_{\varPi}\left\{\,.\,\right\}$ means that this
expected value is taken over all permutations of the train set. Note that the recovery of $\bar{\boldsymbol{w}}_p$ is done through the minimization of $\mathcal{L}_{\boldsymbol{\theta}}$ which we refer to as the train loss function, whereas the evaluation of $\hat{f}\left(\boldsymbol{X}; \boldsymbol{w}_{p}, \boldsymbol{\theta}\right)$ is done through $\mathcal{L}$ which we refer to as the test loss function. The Robust Learning Algorithm is presented here below in \eqref{alg: robust learning}:
\begin{algorithm}[!h]
\caption{\textbf{Robust Learning Algorithm}}
\label{alg: robust learning}
\begin{algorithmic}
\Require\quad 
\begin{enumerate}
    \item[(1)] Pick a model class $\hat{f}_{\boldsymbol{\theta}}$
    \item[(2)] Pick an initial guess $\boldsymbol{\theta}^{\left(0\right)}$.
    \item[(3)] Pick an active learning algorithm \emph{Al}.
    \item[(4)] Set $\mathcal{S}=\phi$, $n=0$ and $C=0$.
\end{enumerate}
\Ensure Partition the data $\mathcal{D}=\left\{ \left(\boldsymbol{x}_{i},y_{i}\right)\right\} _{i\in\mathcal{I}}$ into two mutually exclusive training $\mathcal{D}_{\mathcal{T}}=\left\{\left(\boldsymbol{x}_{i},y_{i}\right)\right\} _{i\in\mathcal{T}}$ and test subsets $\mathcal{D}_{\mathcal{U}}=\left\{ \left(\boldsymbol{x}_{i},y_{i}\right)\right\}_{i\in\mathcal{U}}$.
\While{$C = 0$}
    \State $n \gets n+1$
\While{$1 \leq j \leq K$}
    \State $j \gets j+1$
    \State Pick a random index permutation $\prod_{n}\left(\mathcal{T}\right)$ of $\mathcal{T}$.
    \State Divide the train set $\mathcal{D}_{\mathcal{T}}$ into $\mathcal{D}_{\prod_{n}}$
and $\mathcal{D}_{\prod_{n}^{c}}$. 
    \State Train $\hat{f}\left(\boldsymbol{X};\boldsymbol{w}_{p},\boldsymbol{\theta}^{\left(n-1\right)}\right)$ on $\mathcal{D}_{\prod_{n}}$ with respect to $R_{\prod_{n}}\left(\boldsymbol{w}_{p},\boldsymbol{\theta}^{\left(n-1\right)}\right)$ to obtain $\bar{\boldsymbol{w}}_{p}$.
    \State Evaluate $ e_j = R_{\mathbb{I}-\prod_{n}}\left(\bar{\boldsymbol{w}}_{p},,\boldsymbol{\theta}^{\left(n-1\right)}\right)$ on $\mathcal{D}_{\prod_{n}^{c}}$. 
\EndWhile
\State Set $R_{\mathbb{I}-\prod_{n}} = \frac{1}{K} \sum_{j = 1}^K e_j$.
\State Set $\mathcal{S}\leftarrow\mathcal{S}\cup\left\{ \left(\boldsymbol{\theta}^{\left(n-1\right)},R_{\mathbb{I}-\prod_{n}}\left(\bar{\boldsymbol{w}}_{p},,\boldsymbol{\theta}^{\left(n-1\right)}\right)\right)\right\} $.
\State \If{$Al\left(\mathcal{S}\right)$ has converged}
    \State $C=1$
    \State $\left(\boldsymbol{w}_{p}^{*},\boldsymbol{\theta}^{*}\right)=\left(\bar{\boldsymbol{w}}_{p},,\boldsymbol{\theta}^{\left(n-1\right)}\right)$.
    \ElsIf{$Al\left(\mathcal{S}\right)$ has not converged}
    \State $\boldsymbol{\theta}^{\left(n\right)}=Al\left(\mathcal{S}\right)$
    \EndIf
\EndWhile
\State Check the loss over $\mathcal{D}_{\mathcal{U}}$.
\end{algorithmic}
\end{algorithm}

In particular, notice that $R_{\mathbb{I}-\prod_{n}} = \frac{1}{K}\sum_{j=1}^{K} e_j$ is a (noisy) approximation of $$\mathbb{E}_{\boldsymbol{X},Y\sim\mathcal{D}_{\prod^{c}}}\left\{ \mathcal{L}\left(\hat{f}\left(\boldsymbol{X}; \boldsymbol{w}_{p}, \boldsymbol{\theta}\right),Y\right)\right\}.$$
%Generally, $K$ is a small value\footnote{In the original method proposed by Peyman it is set to $K = 1$.}.

\section{Extension to the multivariate case}\label{sec:multivariate}

We now examine the case where the SDE is multi-dimensional. In such a case 

\begin{equation}
    d\boldsymbol{X}_t = \boldsymbol{f(X_t)}dt + \boldsymbol{\sigma(X_t)}d\boldsymbol{W}_t 
\end{equation}

where $ \boldsymbol{X}_t \in \R^d$, $ \boldsymbol{f} : \R^d \mapsto \R^d, \boldsymbol{\sigma} : \R^d \mapsto \R^{d\times m} $ and $\boldsymbol{W}_t = (W^{i}_t)_{i \leq m}$ is a \textit{standard} multivariate Wiener process.

For simplicity, we place independent GP priors on each dimension, which is equivalent to using a matrix-valued kernel with diagonal entries:

\begin{equation}
    K(x,y) =  	
    \begin{pmatrix}
    K_1(x,y) & 0 \\
    0 & K_2(x,y)
    \end{pmatrix}
\end{equation}

More generally, we may consider matrix-valued kernels with non-zero off-diagonals, but the use of diagonal matrices is practical because it makes all optimization problems independent for each dimension. Then the problem leads to $d$ equations:

\begin{equation}
    dX_t^i = f(\boldsymbol{X_t})^idt + \sum_{j = 1}^{d} \sigma_{i,j}(\boldsymbol{X_t})dW_t^j
\end{equation}
which can all be optimized independently. The optimal value of $\bar{f}(\boldsymbol{X_t})$ conditioned on $\boldsymbol{\bar{\sigma}}$ is given by the posterior mean equation:

\begin{equation}\label{eq: mean drift}
    \boldsymbol{\bar{f}}(\boldsymbol{X}_n)^i = \E[\boldsymbol{f}(\boldsymbol{X}_n)^i] = K^i(\bs{X_n}, \bs{X})\Lambda( \Lambda K^i(\bs{X},\bs{X}) \Lambda + \Sigma^i + \lambda I)^{-1}\boldsymbol{Y}^i
\end{equation}

where $\Sigma^i = \sum_j \Sigma^j$ is the sum of the (diagonal) matrices with entries $\Sigma^j_{k,k} = (\boldsymbol{\bar{\sigma}}(\boldsymbol{X}_k)^j) ^2 \Delta t_k, 1 \leq k \leq N, 1 \leq j \leq d$. On the other hand, the MAP estimate for can be computed jointly for all entries of the matrix $\boldsymbol{ \sigma}^i$ with entries $ \sigma^i_k$ as 

\begin{align}
\begin{split}
        \underset{\boldsymbol{ \sigma}^i}{\arg \min}  &(\bs{Y^i} - \Lambda \bs{f(X)}^i)^T (\Sigma^i + \lambda I)^{-1}(\bs{Y^i} - \Lambda \bs{f(X)}^i) + \sum_{k= 1}^{N}\ln(\Delta t_k\sum_{i = 1}^d (\sigma^i_k)^2+\lambda) \\&+ \sum_{i = 1}^d  \bs{\sigma^i}^T G^i(\bs{X}, \bs{X})^{-1}\bs{\sigma^i}.
\end{split}
\end{align}

Additional assumptions can be made regarding the structure of the noise values, such as assuming that the noise is separate across all dimensions (i.e., $\bs{\sigma}$ is a diagonal matrix) or that the diffusion function $\bs{\sigma}$ is identical across all dimensions (i.e., $\bs{\sigma}$ is a diagonal matrix with identical values along the diagonals). 

\subsection*{Acknowledgments}
MD, BH, HO acknowledge partial support by the Air Force Office of Scientific Research under MURI award number FA9550-20-1-0358 (Machine Learning and Physics-Based Modeling and Simulation).
MD, PT and HO acknowledge  support from Beyond Limits (Learning Optimal Models)
through CAST (The Caltech Center for Autonomous Systems and Technologies).

%\newpage

\bibliographystyle{plain}
%\bibliography{bibliography,merged, kmd, kernel_flows_for_koopman, refs_irreg_kfs, RPS}

\bibliography{merged, RPS}
\end{document}